\documentclass[journal]{IEEEtran}

\usepackage{filecontents}
\usepackage[noadjust]{cite}

\ifCLASSINFOpdf
\else
\fi
\usepackage{amsmath}
\usepackage{amssymb}
\usepackage{mathtools}
\usepackage{enumitem}
\usepackage{ragged2e}
\usepackage{graphicx}
\usepackage{array}
\usepackage{bm}
\usepackage{circuitikz}
\usepackage{tikz}
\usepackage{siunitx} 
\usepackage[hypcap]{caption}
\usetikzlibrary{arrows}
\DeclareMathOperator*{\argmin}{\,min}
\usepackage{multibib}
\newcites{supp}{References}

\newcommand{\RNum}[1]{\uppercase\expandafter{\romannumeral #1\relax}}
\usepackage{adjustbox}

\usepackage{algorithmic}
\usepackage{array}
\usepackage{makecell}
\usepackage{multirow}
\newcolumntype{M}[1]{>{\centering\arraybackslash}m{#1}}
\usepackage{amsthm} % for proof
\newtheorem{theorem}{Theorem} % for theorem
\newtheorem{definition}{Definition}
\newtheorem{Lemma}{Lemma}
\usepackage[noadjust]{cite} % for references

\begin{document}

\title{Optimization Guarantees of Unfolded ISTA and ADMM Networks With Smooth Soft-Thresholding}

\author{
Shaik~Basheeruddin~Shah,~\IEEEmembership{Student~Member,~IEEE,} Pradyumna~Pradhan,~Wei~Pu,~\IEEEmembership{Member,~IEEE,} \\Ramunaidu~Randhi, Miguel~R.~D.~Rodrigues,~\IEEEmembership{Fellow,~IEEE}, Yonina~C.~Eldar,~\IEEEmembership{Fellow,~IEEE}
~
\thanks{Part of this work has been accepted for presentation at IEEE International Conference on Acoustics, Speech, and Signal Processing (ICASSP) 2022.\\
S. B. Shah and Y. C. Eldar are with the Weizmann Institute of Science, Rehovot, Israel. 
P. Pradyumna and R. Ramu Naidu are with the Department of Humanities and Sciences, Indian Institute of Petroleum and Energy, India.
W. Pu is with the University of Electronic Science and Technology, China.
M. Rodrigues is with the Department of Electronic and Electrical Engineering, University College
London, UK. 

This work was supported both by The
Alan Turing Institute and Weizmann – UK Making Connections Programme (Ref. 129589).}}
\maketitle

\begin{abstract}
Solving linear inverse problems plays a crucial role in numerous applications.
Algorithm unfolding based, model-aware data-driven approaches have gained significant attention for effectively addressing these problems.
Learned iterative soft-thresholding algorithm (LISTA) and alternating direction method of multipliers compressive sensing network (ADMM-CSNet) are two widely used such approaches, based on ISTA and ADMM algorithms, respectively.
In this work, we study optimization guarantees, i.e., achieving near-zero training loss with the increase in the number of learning epochs,  for finite-layer unfolded networks such as LISTA and  ADMM-CSNet with smooth soft-thresholding in an over-parameterized (OP) regime.
We achieve this by leveraging a modified version of the Polyak-Łojasiewicz, denoted PL$^*$, condition.
Satisfying the PL$^*$ condition within a specific region of the loss landscape ensures the existence of a global minimum and exponential convergence from initialization using gradient descent based methods. 
Hence, we provide conditions, in terms of the network width and the number of training samples, 
on these unfolded networks for the PL$^*$ condition to hold.
We achieve this by deriving the Hessian spectral norm of these networks.
Additionally, we show that the threshold on the number of training samples increases with the increase in the network width.
Furthermore, we compare the threshold on training samples of unfolded networks with that of a standard fully-connected feed-forward network (FFNN) with smooth soft-thresholding non-linearity.
We prove that unfolded networks have a higher threshold value than FFNN.
Consequently, one can expect a better expected error for unfolded networks than FFNN.
% To support the proposed theory, we present several numerical results.
% by deriving the minimum eigenvalue of the tangent kernel matrix at initialization.
% \textcolor{red}{Comparison of unfolded with the standard in terms of threshold on training samples.}
% Finally, we show that unfolded networks outperform standard FFNNs, in terms of parameter efficiency. Particularly, we demonstrate that, given a fixed number of training samples, the standard networks require a greater number of parameters to achieve near-zero training loss compared to LISTA/ADMM-CSNet.
% Next, we propose a threshold for the number of training samples to achieve zero training error for both unfolded networks. 
% We show that this threshold is high for the LISTA network when compared with the ADMM-CSNet.
% Through several simulations, we further justify the theoretical findings proposed in this work.
% We justify the theoretical findings in this work through several simulations.
% Specifically, we derive a constraint on the width of the finite-layer unfolded network to satisfy PL$^*$ by computing the corresponding Hessian spectral norm.
% Utilizing this constraint, we derive a threshold on the number of training samples to achieve near-zero empirical training loss.
% Specifically, by deriving the Hessian spectral norm \textcolor{blue}{of these networks}, we provide conditions on both the network width and the number of training samples to satisfy PL$^*$.
\end{abstract}

% Note that keywords are not normally used for peer review papers.

\IEEEpeerreviewmaketitle

\section{Introduction}
\IEEEPARstart{L}{inear} inverse problems are fundamental in many engineering and science applications \cite{EldarCS, DonohoCS}, where the aim is to recover a vector of interest or target vector from an observation vector.
Existing approaches to address these problems can be categorized into two types; model-based and data-driven.
Model-based approaches use mathematical formulations that represent knowledge of the underlying model, which connects observation and target information.  
These approaches are simple, computationally efficient, and require accurate model knowledge for good performance \cite{Nir1, Nir2}.
% Model-based approaches/algorithms use the mathematical formulation by knowing the domain knowledge (i.e. knowing the underlying model that relates observation and target information) which comes from the statistical model postulates.
% Model-based approaches typically use the mathematical formulation that represents the underlying model, and they require accurate model knowledge for good performance \cite{Nir1, Nir2}.
% Model-based approaches typically use the mathematical formulation representing the underlying model that maps the observation vector with the target vector. These approaches require accurate model knowledge for good performance \cite{Nir1, Nir2}.
In data-driven approaches, a machine learning (ML) model, e.g., a neural network, with a training dataset, i.e., a supervised setting, is generally considered. Initially, the model is trained by minimizing a certain loss function. Then, the trained model is used on unseen test data.
Unlike model-based methods, data-driven approaches do not require underlying model knowledge. 
However, they require a large amount of data and huge computational resources while training \cite{Nir1, Nir2}.
% used to recover the target vector on the new unseen observation vector.
% Usually, the data-driven approaches require a high amount of data and huge computational resources \cite{Nir1, Nir2}.
% Whereas, in data-driven approaches, an ML model is trained by minimizing a certain loss function using the available data; these approaches generally
% % Model-based approaches usually require accurate model knowledge for good performance.
% % Generally, data-driven approaches 
% require a huge amount of data and computational resources \cite{Nir1, Nir2}.
% % in signal processing. There are many applications where the given problem turns out to be addressing the linear inverse problem \cite{}.

By utilizing both domains' knowledge, i.e., the mathematical formulation of the model and ML ability, a new approach, called model-aware data-driven, has been introduced \cite{UnrollingPerformance, Unrolling}.
% Specifically, a neural network is designed based on an iterative algorithm that addresses the mathematical formulation (typically an optimization problem) corresponding to the given model.
This approach involves the construction of a neural network architecture based on an iterative algorithm, which solves the optimization problem associated with the given model.
% which solves the optimization problem corresponding to the given model.
% In specific, based on an iterative algorithm, which solves the optimization problem corresponding to the given model, a neural network is designed.
% by utilizing the given mathematical formulation of the model, 
This process is called algorithm unrolling or unfolding \cite{Unrolling}.
% An unfolded network is typically trained by using a limited amount of data. 
It has been observed that the performance, in terms of accurate recovery of the target vector, training data requirements, and computational complexity, of model-aware data-driven networks is better when compared with existing 
% model-based and data-driven
techniques \cite{UnrollingPerformance, ADMM-CSNet}.
Learned iterative soft-thresholding algorithm (LISTA) and alternating direction method of multipliers compressive sensing network (ADMM-CSNet) are two popular unfolded networks that have been used in many applications such as image compressive sensing \cite{ADMM-CSNet}, image deblurring \cite{Deblurring}, image super-resolution \cite{ImageSuperResolution}, super-resolution microscopy \cite{LearnedSparcom}, clutter suppression in ultrasound \cite{Ultrasound}, power system state estimation \cite{PowerSystem}, and many more.

Nevertheless, the theoretical studies supporting these unfolded networks remain to be established.
There exist a few theoretical studies that address the challenges of generalization \cite{Avner, Generalization1, Generalization2} and convergence rate \cite{liu2018alista, NEURIPS2018_cf8c9be2, LISTACOnvergence1} in unfolded networks.
% Having said that, there are a few theoretical works that address the generalization and optimization issues of unfolded networks.
For instance, in \cite{Avner}, the authors showed that unfolded networks exhibit higher generalization capability compared with standard ReLU networks by deriving an upper bound on the generalization and estimation errors.
In \cite{liu2018alista, NEURIPS2018_cf8c9be2, LISTACOnvergence1} the authors examined the LISTA network convergence to the ground truth as the number of layers increases i.e., layer-wise convergence (which is analogous to iteration-wise convergence in the ISTA algorithm).
Furthermore, in \cite{liu2018alista, NEURIPS2018_cf8c9be2, LISTACOnvergence1}, the network weights are not learned but are calculated in an analytical way (by solving a data-free optimization problem).
Thus, the network only learns a few parameters, like threshold, step size, etc., from the available data.
In this work, we study guarantees to achieve near-zero training loss with an increase in the number of learning epochs, i.e., \textit{optimization guarantees}, by using gradient descent (GD) for both LISTA and ADMM-CSNet with smooth activation in an over-parameterized regime. 
Note that, our work differs from \cite{liu2018alista, NEURIPS2018_cf8c9be2, LISTACOnvergence1}, as we focus on the convergence of training loss with the increase in the number of epochs by fixing the number of layers in the network.
% \textcolor{red}{We show that.... what is the conclusion?}
% analyzed the convergence of the output error after several layers, i.e., layer-wise convergence (which is analogous to iteration-wise convergence in the ISTA algorithm) in a finite-layer unfolded network, while we focus on the convergence of training loss.
% where the authors studied the convergence of the output error when the weights in the unfolded network are well learned. Whereas in this work, we focus on the convergence of the training loss.
% As recent literature has highlighted the importance of optimization \cite{Double_Descent1, Double_Descent2, Double_Descent3, Double_Descent4, Double_Descent5, Double_Descent6}, our work makes a contribution to the ongoing theoretical advancements in the field of unfolded networks.

\begin{figure}[t!]
    \centering
    \includegraphics[height = 4.8cm, width = 8.5cm]{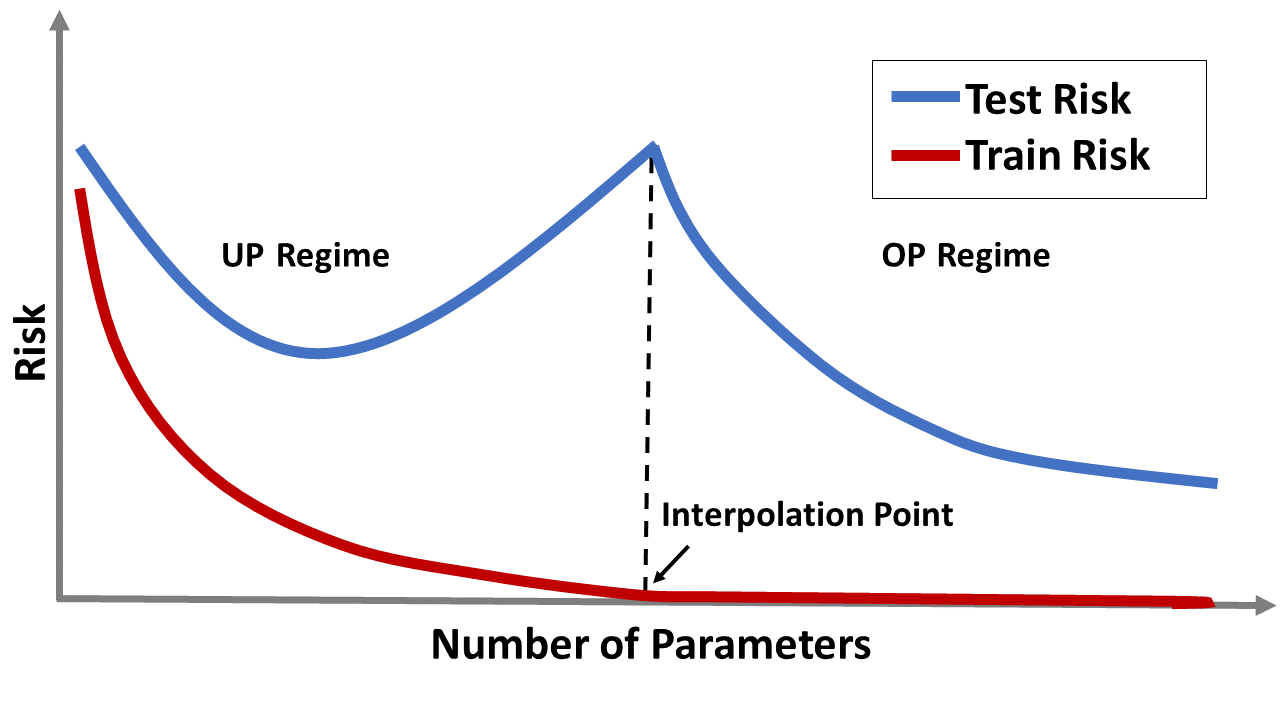}
    \caption{Double descent risk curve.}
    \label{DD_Curve}
\end{figure}

% \textbf{Motivation:} 
In classical ML theory, we aim to minimize the expected/test risk by finding a balance between under-fitting and over-fitting, i.e., achieving the bottom of the classical U-shaped test risk curve \cite{ClassicalML}.
However, modern ML results establish that large models that try to fit train data exactly, i.e., interpolate,
% implies getting zero empirical training loss, 
\textit{often} show high test accuracy even in the presence of noise \cite{Double_Descent1, Double_Descent2, Double_Descent3, Double_Descent4, Double_Descent5, Double_Descent6}. 
Recently, ML practitioners proposed a way to numerically justify the relationship between classical and modern ML practices. They achieved this by proposing a 
% Recently, the relation between classical and modern ML practices is justified numerically by a
performance curve called the double-descent test risk curve \cite{Double_Descent1, Double_Descent2, Double_Descent4, Double_Descent5}, which is depicted in Fig. \ref{DD_Curve}. 
This curve shows that increasing the model capacity (e.g., model parameters)
until interpolation results in the classical U-shaped risk curve; further increasing it beyond the interpolation point reduces the test risk. 
Thus, understanding the conditions -- as a function of the training data --  that allow perfect data fitting is crucial.

Neural networks can be generally categorized into under-parameterized (UP) and over-parameterized (OP), based on the number of trainable parameters and the number of training data samples.
If the number of trainable parameters is less than the number of training samples, then the network is referred to as an UP model, else, referred to as an OP model.
% Most of the modern neural networks fall in the OP model category. 
% As an unfolded network works on a limited amount of data, usually, such networks also fall under the OP model category.
The loss landscape of both UP and OP models is generally non-convex. However, OP networks satisfy \textit{essential non-convexity} \cite{LossLandscape}. 
Particularly, 
% the loss landscape of an UP model has isolated local minima with convexity around a small neighborhood of a local minimum. In contrast, 
the loss landscape of an OP model has a non-isolated manifold of global minima with non-convexity around any small neighborhood of a global minimum.
% Hence, one can't use the well-known convex theory to analyze the OP models.
Despite being highly non-convex, GD based methods work well for training OP networks \cite{GD1,GD2,GD3,GD4}. 
% In spite of satisfying essential non-convexity, the performance of simple gradient descent (GD) based approaches is good on the OP models while training \cite{GD1,GD2,GD3,GD4}.
Recently, in \cite{LossLandscape, Linearity}, the authors provided a theoretical justification for this.
% theoretically justified the reason behind this.
Specifically, they proved that the loss landscape, corresponding to the squared loss function, of a typical smooth OP model holds the modified version of the Polyak-Łojasiewicz condition, denoted PL$^*$, on most of the parameter space.
Indeed, a necessary (but not sufficient) condition to satisfy the PL$^*$ is that the model should be in OP regime.
% They proved that the loss landscape of most of the OP models holds the modified version of the Polyak-Łojasiewicz condition, denoted PL$^*$, on most of the parameter space.
Satisfying PL$^*$ on a region in the parameter space guarantees the existence of a global minimum in that region, and exponential convergence to the global minimum from the Gaussian initialization using simple GD.

% Unfolded networks usually work on a limited amount of data, hence, such networks generally fall under the OP regime. 
Motivated by the aforementioned PL$^*$-based mathematical framework of OP networks, in this paper, we analyze optimization guarantees of finite-layer OP based unfolded ISTA and ADMM networks.
%with a smooth version of the soft-thresholding activation function.
Moreover, as the analysis of PL$^*$ depends on the double derivative of the model \cite{LossLandscape}, we consider a smooth version of the soft-thresholding as an activation function.
The major contributions of the paper are summarized as follows:
\begin{itemize}
    \item As the linear inverse problem aims to recover a vector, we initially extend the gradient-based optimization analysis
    % (leveraging PL$^*$ condition) 
    of the OP model with a scalar output, proposed in \cite{LossLandscape}, to a vector output. 
    In the process, we prove that a necessary condition to satisfy PL$^*$ is $P\gg mT$, where $P$ denotes the number of parameters, $m$ is the dimension of the model output vector, and $T$ denotes the number of training samples.
    \item In \cite{Linearity, LossLandscape}, the authors provided a condition on the width of a fully-connected feed-forward neural network (FFNN) with scalar output to satisfy the PL$^*$ condition by utilizing the Hessian spectral norm of the network.
    Motivated by this work, we derive the Hessian spectral norm of finite-layer LISTA and ADMM-CSNet with smoothed soft-thresholding non-linearity. We show that the norm is on the order of $\Tilde{\Omega}\left(1/\sqrt{m}\right)$, where $m$ denotes the width of the network which is equal to the target vector dimension. 
    % Based on this, we provide a constraint on $m$ to satisfy the PL$^*$ condition. 
     % we first compute the Hessian spectral norm of both LISTA and ADMM-CSNet; and we show that it is of the order of $O(\sqrt{m})$, where $m$ denotes the width of the network which is also equal to the dimension of the vector of interest.
    % Then, using this Hessian spectral norm, we provide a constraint on the finite-layer unfolded network width to satisfy the PL$^*$ condition.
    \item By employing the Hessian spectral norm, we derive necessary conditions on both $m$ and $T$ to satisfy the PL$^*$ condition for both LISTA and ADMM-CSNet.  
    Moreover, we demonstrate that the threshold on $T$, which denotes the maximum number of training samples that a network can memorize, increases as the network width increases. 
    % Using the Hessian spectral norm, we also derive a threshold on the number of training data samples to obtain near-zero empirical training loss for both LISTA and ADMM-CSNet. We demonstrate that this threshold increases with the increase in the network width. 
    % such that the empirical training error of both LISTA and ADMM-CSNet networks converges to zero.
    % We show that this threshold is higher for LISTA in comparison with ADMM-CSNet. 
    % This implies that there are certain regimes where the ADMM-CSNet training error does not converge to zero but the LISTA network training error does.
    \item We compare the threshold on the number of training samples of LISTA and ADMM-CSNet with that of FFNN, solving a given linear inverse problem. Our findings show that LISTA/ADMM-CSNet exhibits a higher threshold value than FFNN. Specifically, we demonstrate this by proving that the upper bound on the minimum eigenvalue of the tangent kernel matrix at initialization is high for LISTA/ADMM-CSNet compared to FFNN.
    This implies that, with fixed network parameters, the unfolded network is capable of memorizing a larger number of training samples compared to FFNN.
    % the threshold on T — guaranteeing memorisation — can be higher for unfolded networks than conventional ones, 
    Therefore, we should expect to obtain a better expected error (which is upper bounded by the sum of generalization and training error \cite{Expected_error}) for unfolded networks than FFNN.
    % because training error is zero whereas generalization error is smaller \cite{Avner}.}
    % \textcolor{red}{As a result, LISTA/ADMM-CSNet gets a better generalization performance compared to FFNN (refer to \cite{Avner}).}
    % show that this threshold is high for LISTA and ADMM-CSNet in comparison with the threshold of FFNN that addresses the linear inverse problem.

    \item Additionally, we numerically evaluate the parameter efficiency of unfolded networks in comparison to FFNNs. In particular, we demonstrate that FFNNs require a higher number of parameters to achieve near-zero empirical training loss compared to LISTA/ADMM-CSNet for a given fixed $T$ value.
    % \item We compare unfolded networks with non-unfolded networks, such as fully-connected feed-forward neural networks (FFNNs), in terms of training loss convergence. Specifically, for a fixed number of trainable parameters, we show that  LISTA/ADMM-CSNet gives better training error convergence compared with fully-connected FFNNs.
\end{itemize}

\textbf{Outline:}
The paper is organized as follows: Section \RNum{2} presents a comprehensive discussion on LISTA and ADMM-CSNet, and also formulates the problem.
Section \RNum{3} extends the PL$^*$-based optimization guarantees of an OP model with scalar output to a model with multiple outputs. 
% introduces some basic preliminaries related to OP networks with multiple outputs. 
Section \RNum{4} begins by deriving the Hessian spectral norm of the unfolded networks.
Then, it provides conditions on the network width and on the number of training samples to satisfy the PL$^*$ condition.
Further, it also establishes a comparative analysis of the threshold for the number of training samples among LISTA, ADMM-CSNet, and FFNN.
% Further, it derives the bound on the number of training samples to get zero training error.
Section \RNum{5} discusses the experimental results and
Section \RNum{6} draws conclusions.
% Note that our work also differs from \cite{LossLandscape, Linearity}, where the authors provided the mathematical framework for getting zero empirical training error in an OP network by leveraging the PL$^*$ condition.
% % Further, they showed that sufficiently wide neural networks with scalar output satisfy the PL$^*$ condition. Indeed, the theoretical analysis performed in this work is motivated from \cite{LossLandscape,Linearity}.
% Inspired by this mathematical framework, we deal with the unfolded networks in this paper.

\textbf{Notations:}
The following notations are used throughout the paper. 
The set of real numbers is denoted by
$\mathbb{R}$.
% denotes the set of real numbers. 
We use bold lowercase letters, e.g., $\mathbf{y}$, for vectors, capital letters, e.g., ${W}$, for matrices, and bold capital letters, e.g., $\mathbf{H}$, for tensors.
Symbols $||\mathbf{z}||_1$,  $||\mathbf{z}||$, and $||\mathbf{z}||_\infty$ denote the $l_1$-norm, $l_2$-norm, and $l_\infty$-norm of $\mathbf{z}$, respectively.
The spectral norm and Frobenius norm of a matrix ${W}$ are written as $||W||$ and $||W||_F$, respectively.
We use $[L]$ to denote the set $\{1,2,\dots,L\}$, where $L$ is a natural number.
The first-order derivative or gradient of a function $L(\mathbf{w})$ w.r.t. $\mathbf{w}$ is denoted as $\nabla_\mathbf{w}L(\mathbf{w})$. 
The asymptotic upper bound and lower bound on a quantity are described using $O(\cdot)$ and $\Omega(\cdot)$, respectively. Notations 
$\tilde{O}(\cdot)$ and $\tilde{\Omega}(\cdot)$ are used to suppress the logarithmic terms in $O(\cdot)$ and $\Omega(\cdot)$, respectively. For example, $O\left(\frac{1}{m}\ln(m)\right)$ is written as $\tilde{O}\left(\frac{1}{m}\right)$.
Symbols $\gg$ and $\ll$ mean ``much greater than'' and ``much lesser than'', respectively.
Consider a matrix $G$ with $G_{i,j} = \sum_{k}A_{i,j,k}v_k$, where $A_{i,j,k}$ is a component in tensor $\mathbf{A}\in\mathbb{R}^{m_1\times m_2\times m_3}$. The spectral norm of $G$ can be bounded as 
\begin{equation}
   \|G\|\leq \|\mathbf{A}\|_{2,2,1}\|\mathbf{v}\|_\infty.
   \label{Holders_inequality}
\end{equation}
Here $\|\mathbf{A}\|_{2,2,1}$ denotes the $(2,2,1)$-norm of the tensor $\mathbf{A}$, which is defined as 
\begin{equation}
    \|\mathbf{A}\|_{2,2,1} = {\sup\limits_{\|\mathbf{r}\| = \|\mathbf{s}\| = 1}}\sum\limits_{k=1}^{m_3}\left|\sum\limits_{i=1}^{m_1}\sum\limits_{j=1}^{m_2}A_{i,j,k}r_i s_j\right|,
    \label{2_2_2_Norm}
\end{equation}
where $\mathbf{r}\in \mathbb{R}^{m_1\times 1}$ and $\mathbf{s}\in \mathbb{R}^{m_2\times 1}$.
% Consider a function $\mathcal{F}(\cdot)$ with $P\times 1$ as its domain and $m\times T$ as its co-domain, i.e., $\mathcal{F}(\cdot):\mathbb{R}^{P}\xrightarrow{}\mathbb{R}^{m\times T}$. 
% We define the following for $\mathcal{F}(\cdot)$.
% \textit{Lipschitz continuity}:  $\mathcal{F}(\cdot)$ is $L_\mathcal{F}$-Lipschitz continuous if $\|\mathcal{F}(\mathbf{w}_1)-\mathcal{F}(\mathbf{w}_2)\|_F\leq L_\mathcal{F}\|\mathbf{w}_1-\mathbf{w}_2\|,\ \forall\  \mathbf{w}_1,\ \mathbf{w}_2\in \mathbb{R}^P$.
% \textit{Smoothness}:  $\mathcal{F}(\cdot)$ is $\beta_\mathcal{F}$-smooth if the gradient of the function is $\beta_\mathcal{F}$-Lipschitz, i.e.,
% $\|\nabla_\mathbf{w_1}\mathcal{F}(\mathbf{w_1})-\nabla_\mathbf{w_2}\mathcal{F}(\mathbf{w_2})\|_F\leq \beta_\mathcal{F}\|\mathbf{w_1}-\mathbf{w_2}\|$, $\forall\ \mathbf{w}_1,\ \mathbf{w}_2\in \mathbb{R}^P$.
% \textit{Hessian Spectral Norm:} The Hessian spectral norm of $\mathcal{F}(\cdot)$ is defined as $\|\mathbf{H}_\mathcal{F}(\mathbf{w})\| = \text{max}_{i\in[T]}\|\mathbf{H}_{\mathcal{F}_i}(\mathbf{w})\|$, where $\mathbf{H}_\mathcal{F}\in\mathbb{R}^{T\times m\times P\times P}$ is a tensor with $(\mathbf{H}_{\mathcal{F}})_{i,j,k,l} = \frac{\partial^2\left(\mathcal{F}(\mathbf{w})\right)_{j,i}}{\partial w_k \partial w_l}$, $\mathbf{H}_{\mathcal{F}_i} = \frac{\partial^2(\mathcal{F}(\mathbf{w}))_i}{\partial \mathbf{w}^2}$, and $(\mathcal{F}(\mathbf{w}))_i$ is the $i^\text{th}$ column in  $\mathcal{F}(\mathbf{w})$.
\section{Problem Formulation}
\subsection{LISTA and ADMM-CSNet}
\label{Problem Formulation}
Consider the following linear inverse problem
\begin{equation}
    \mathbf{y} = {A}\mathbf{x} + \mathbf{e}.
\end{equation}
Here $\mathbf{y}\in\mathbb{R}^{n\times 1}$ is the observation vector, $\mathbf{x}\in\mathbb{R}^{m\times 1}$ is the target vector, ${A}\in\mathbb{R}^{n\times m}$ is the forward linear operator matrix with $m>n$, and $\mathbf{e}$ is noise with $\|\mathbf{e}\|_2<\epsilon$, where the constant $\epsilon>0$.
Our aim is to recover $\mathbf{x}$ from a given $\mathbf{y}$.
% this problem is typically solved using
% As stated earlier, we typically solve this problem using model-based or data-driven, or model-aware data-driven approaches.
% In model-based approaches, the mathematical composition that represents the underlying domain knowledge is utilized. 

In model-based approaches, an optimization problem is formulated using some prior knowledge about the target vector and is usually solved using an iterative algorithm.
For instance, by assuming $\mathbf{x}$  is a $k$-sparse vector \cite{LASSO}, the least absolute shrinkage and selection operator (LASSO) problem is formulated as
\begin{equation}
    {\argmin_\mathbf{x}}\  \frac{1}{2}\|\mathbf{y} - A\mathbf{x}\|^2 + \gamma\|\mathbf{x}\|_1,
    \label{LASSO}
\end{equation}
where $\gamma$ is a regularization parameter. 
Iterative algorithms, such as ISTA and ADMM \cite{ProximalAlg}, are generally used to solve the LASSO problem.
The update of $\mathbf{x}$ at the $l^\text{th}$ iteration in ISTA is \cite{ISTA}
\begin{equation}
    \mathbf{x}^{l} = S_{\gamma\tau}\left\{\left(\mathbf{I}-\tau{A}^T{A}\right)\mathbf{x}^{l-1} + {\tau}{A}^T\mathbf{y}\right\},
    \label{ISTA}
\end{equation}
where $\mathbf{x}^0$ is a bounded input initialization, $\tau$ controls the iteration step size, and $S_{\lambda}(\cdot)$ is the soft-thresholding operator applied element-wise on a vector argument
% For any $x\in\mathbb{R}$, $S_{\lambda}(x)$ is defined as,
% . For $x\in\mathbb{R}$, $S_{\gamma\tau}(x)$ is defined as
$ S_{\lambda}(x) = \text{sign}(x)\text{max}\left(|x|-\lambda,0\right).$
The $l^\text{th}$ iteration in ADMM is \cite{ADMMBook}
\begin{equation}
\begin{aligned}
       \mathbf{x}^{l} &= \left(A^TA+\rho\mathbf{I}\right)^{-1}\left({A}^T\mathbf{y}+\rho\left(\mathbf{z}^{l-1}-\mathbf{u}^{l-1}\right)\right), \\
     \mathbf{z}^l &= S_{\frac{\gamma}{\rho}}\left(\mathbf{x}^l+\mathbf{u}^{l-1}\right), \\
     \mathbf{u}^l &= \mathbf{u}^{l-1}+\left(\mathbf{x}^l-\mathbf{z}^l\right),  
     \label{ADMM}
\end{aligned}
\end{equation}
where $\mathbf{x}^0$, $\mathbf{z}^0$, and $\mathbf{u}^0$, are bounded input initializations to the network and $\rho >0$ is a penalty parameter.
Model-based approaches are in general sensitive to inaccurate knowledge of the underlying model \cite{Nir1, Nir2}.
In turn, data-driven approaches use an ML model to recover the target vector. These approaches generally require a large amount of training data and computational resources \cite{Nir1, Nir2}.

% In data-driven approaches, an ML model, e.g., a neural network, with a training dataset, i.e., a supervised setting, is generally considered. Initially, the model is trained by minimizing a certain loss function, and then the trained model is used to recover the target vector on the new unseen observation vector.
% Usually, the data-driven approaches require a high amount of data and huge computational resources \cite{Nir1}. 

\begin{figure}[t!]
    \centering
    \includegraphics[height = 2.5cm, width = 6cm]{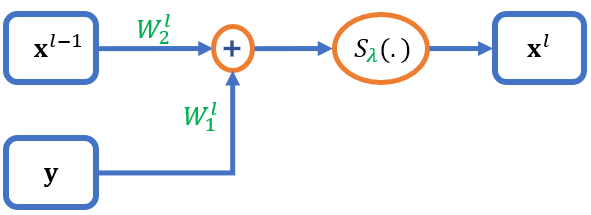}
    \caption{$l^\text{th}$ layer of the unfolded ISTA network.}
    \label{fig_1_ISTA_lth_layer}
\end{figure}

\begin{figure}[t!]
    \centering
    \includegraphics[height = 4.25cm, width = 8cm]{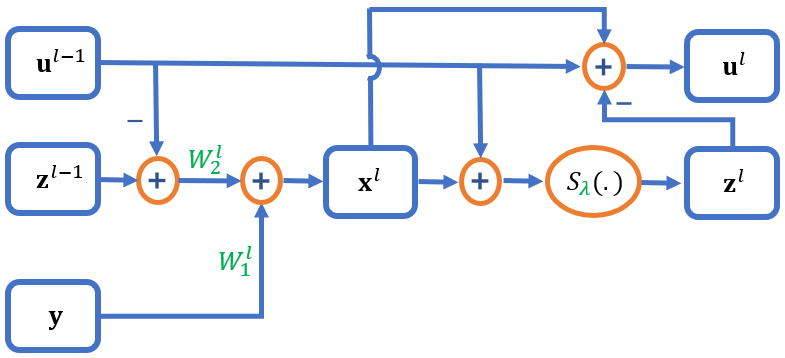}
    \caption{$l^\text{th}$ layer of the unfolded ADMM network.}
    \label{fig_2_ADMM_lth_layer}
\end{figure}

A model-aware data-driven approach 
% is proposed by utilizing the domain knowledge of both model-based and data-driven approaches. 
% Specifically, we use the mathematical formulation from the model-based and data learning from data-driven approaches.
is generally developed using algorithm unfolding or unrolling \cite{Unrolling}.
% In unfolding, for the formulated optimization problem and its corresponding iterative solver such as \eqref{LASSO} and \eqref{ISTA} or \eqref{ADMM}, a neural network is constructed by mapping each iteration in the iterative algorithm to a network layer.
In unfolding, a neural network is constructed by mapping each iteration in the iterative algorithm (such as \eqref{ISTA} or \eqref{ADMM}) to a network layer.
Hence, an iterative algorithm with $L$-iterations leads to an $L$-layer cascaded deep neural network.
The network is then trained by using the available dataset containing a series of pairs $\{\mathbf{y}_i,\mathbf{x}_i\}, i \in [T]$.
For example, the update of $\mathbf{x}$ at the $l^\text{th}$ iteration in ISTA, given in \eqref{ISTA}, is rewritten as 
\begin{equation}
    \mathbf{x}^{l} = S_{\lambda}\left\{{W}_2^l\mathbf{x}^{l-1} + W_1^l\mathbf{y}\right\},
    \label{ISTA_unfolded}
\end{equation}
where $\lambda = \gamma\tau$, $W_1^l = \tau{A}^T$, and $W_2^l = \mathbf{I}-\tau{A}^T{A}$.
% Fig. \ref{fig_1_ISTA_lth_layer} depicts the $l^\text{th}$ layer of the unfolded ISTA network obtained by considering $W_1^l$, $W_2^l$, and $\lambda$ as learnable parameters. 
By considering $W_1^l$, $W_2^l$, and $\lambda$ as network learnable parameters, one can map the above $l^\text{th}$ iteration to an $l^\text{th}$ layer in the network as shown in Fig. \ref{fig_1_ISTA_lth_layer}.
The corresponding unfolded network is called learned ISTA (LISTA) \cite{UnrollingPerformance}.
Similarly, by considering $W_1^l = \left(A^TA+\rho\mathbf{I}\right)^{-1}{A}^T$, $W_2^l = \left(A^TA+\rho\mathbf{I}\right)^{-1}\rho$, and $\lambda = \frac{\gamma}{\rho}$ as learnable parameters, \eqref{ADMM} is rewritten as 
\begin{equation}
\begin{aligned}
       \mathbf{x}^{l} &= W_1^l\mathbf{y}+ W_2^l\left(\mathbf{z}^{l-1}-\mathbf{u}^{l-1}\right), \\
     \mathbf{z}^l &= S_{\lambda}\left(\mathbf{x}^l+\mathbf{u}^{l-1}\right), \\
     \mathbf{u}^l &= \mathbf{u}^{l-1}+\left(\mathbf{x}^l-\mathbf{z}^l\right).  
     \label{LearnedADMM}
\end{aligned}
\end{equation}
The above $l^\text{th}$ iteration in ADMM can be mapped to an $l^\text{th}$ layer in a network as shown in Fig. \ref{fig_2_ADMM_lth_layer}, leading to ADMM-CSNet \cite{ADMM-CSNet}. 
Note that from a network point of view, the inputs of $l^\text{th}$ layer are $\mathbf{x}^{l-1}$ and $\mathbf{y}$ for LISTA, and $\mathbf{z}^{l-1}$, $\mathbf{u}^{l-1}$ and $\mathbf{y}$ for ADMM-CSNet. 
It has been observed that the performance of LISTA and ADMM-CSNet is better in comparison with ISTA, ADMM, and traditional networks, in many applications \cite{UnrollingPerformance, ADMM-CSNet}.
For instance, to achieve good performance the number of layers required in an unrolled network is generally much smaller than the number of iterations required by the iterative solver \cite{UnrollingPerformance}.
In addition, an unrolled network works effectively even if the linear operator matrix, ${A}$, is not known exactly.
% Unlike classical deep neural networks,
An unrolled network typically requires less data for training compared to standard deep neural networks \cite{Nir1} to achieve a certain level of performance on unseen data.
% as it is designed based on the mathematical formulation of the underlying model.
Due to these advantages, LISTA and ADMM-CSNet have been used in many applications \cite{ADMM-CSNet, Deblurring, ImageSuperResolution, LearnedSparcom, Ultrasound, PowerSystem}.
That said, the theoretical foundations supporting these networks remain to be established.
While there have been some studies focusing on the generalization \cite{Avner, Generalization1, Generalization2} and convergence rate \cite{liu2018alista, NEURIPS2018_cf8c9be2, LISTACOnvergence1} of unfolded networks, a comprehensive study of the optimization guarantees 
% specifically for finite $L$-layer LISTA and ADMM-CSNet 
is lacking.
Here, we analyze the conditions on finite $L$-layer LISTA and ADMM-CSNet to achieve near-zero training loss with the increase in the number of epochs.

\subsection{Problem Formulation}
\label{Problem_Formulation}
% The problem here is to analyze the optimization guarantees of a finite $L$-layer unfolded ISTA and ADMM networks. 
We consider the following questions: Under what conditions does the training loss in LISTA and ADMM-CSNet converge to zero as the number of epochs tends to infinity using GD?
Additionally, how do these conditions differ for FFNNs?

For the analysis, we consider the following training setting: Let $\mathbf{x} = F(\mathbf{w},\lambda;\mathbf{y})$ be an $L$-layer unfolded model, where $\mathbf{y}\in \mathbb{R}^{n\times 1}$ is the model input vector, $\mathbf{x}\in \mathbb{R}^{m\times 1}$ is the model output, and $\mathbf{w}\in\mathbb{R}^{P\times 1}$ and $\lambda$ are the learnable parameters. 
To simplify the analysis, $\lambda$ is assumed to be constant, henceforth, we write $F(\mathbf{w},\lambda;\mathbf{y})$ as $F(\mathbf{w};\mathbf{y})$.
This implies that $\mathbf{w}_{P\times 1} = \text{Vec}\left([\mathbf{W}]_{L\times m \times (m+n)}\right)$ is the only learnable (untied) parameter vector, where
\begin{equation}
    \mathbf{W} = \left[W^1\ W^2\ \dots\ W^L\right], 
\end{equation}
and $\left[W^l\right]_{m\times (m+n)} = \left[W_1^l\ \ W_2^l\right]$ is the parameter matrix corresponding to the $l^\text{th}$-layer. Alternatively, we can write 
\begin{equation}
    \mathbf{W} = \left[\left[\mathbf{W}_1\right]_{L\times m\times n}\ \ \left[\mathbf{W}_2\right]_{L\times m\times m}\right],
\end{equation}
$\mathbf{W}_1 = \left[W_1^1\ \dots\ W_1^L\right]$ and
$\mathbf{W}_2 = \left[W_2^1\  \dots\ W_2^L\right]$.
% Given the model $F(\mathbf{w};\mathbf{y})$, we first 
Consider the training dataset $\left\{ \mathbf{y}_i, \mathbf{x}_i\right\}_{i=1}^{T}$.
An optimal parameter vector $\mathbf{w}^*$, such that $F(\mathbf{w}^*; \mathbf{y}_i) \approx \mathbf{x}_i,\ \forall i\in[T]$, is found by minimizing  an empirical loss function $L(\mathbf{w})$,
% i.e., the loss over an entire dataset, 
defined as
\begin{equation}
    L(\mathbf{w}) =\sum_{i=1}^{T}l(\mathbf{f}_i,\mathbf{x}_i),
\end{equation}
where $l(\cdot)$ is the loss function,  $\mathbf{f}_i = (\mathcal{F}(\mathbf{w}))_i = F(\mathbf{w}, \mathbf{y}_i)$, $\mathcal{F}(\cdot):\mathbb{R}^{P\times 1}\xrightarrow{}\mathbb{R}^{m\times T}$, and $(\mathcal{F}(\mathbf{w}))_i$ is the $i^\text{th}$ column in  $\mathcal{F}(\mathbf{w})$.
% As the unfolded network falls under the multi-output regression model, 
We consider the squared loss, hence
\begin{equation}    
    L(\mathbf{w}) = \frac{1}{2}\sum_{i=1}^{T}\|\mathbf{f}_i-\mathbf{x}_i\|^2 = \frac{1}{2}\|\mathcal{F}(\mathbf{w})-X\|_F^2,
    \label{Loss_function}
\end{equation}
where ${X} = \left[\mathbf{x}_1,\dots,\mathbf{x}_T\right]$.
We choose GD as the optimization algorithm for minimizing $L(\mathbf{w})$, hence, the updating rule is $$\mathbf{w}_{t+1}=\mathbf{w}_t-\eta \nabla_{\mathbf{w}}L(\mathbf{w}) $$
where $\eta$ is the learning rate.

Our aim is to derive conditions on LISTA and ADMM-CSNet such that $L(\mathbf{w})$ converges to zero with an increase in the number of epochs using GD, i.e., $\lim_{t \rightarrow \infty} L(\mathbf{w}_t) = 0.$
In addition, we compare these conditions with those of FFNN, where we obtain the conditions for FFNN by extending the analysis given in \cite{LossLandscape}.
Specifically, in Section \ref{Conditions_for_Unfolded_Networks_to_Satisfy_PL$^*$}, we derive a bound on the number of training samples to achieve near zero training loss for unfolded networks. Further, we show that this threshold is lower for FFNN compared to unfolded networks.
% As a result, one can expect a better expected error for unfolded networks than that of FFNN.

% To begin with, in the following section, we discuss the optimization theory for a multi-output model based on the PL$^*$ condition, which paves the way to analyze the optimization guarantees of unfolded networks.
% of considered unfolded networks.
% Precisely, under what conditions, do we obtain the global minimum in LISTA and ADMM-CSNet networks using a simple GD approach while training?

\section{Revisiting PL$^*$-Based Optimization Guarantees}
\label{Preliminaries}
In \cite{LossLandscape} the authors proposed PL$^*$-based optimization theory for a model with a scalar output. Motivated by this, in this section, we extend this theory to a multi-output model, as we aim to recover a vector in a linear inverse problem.

Consider an ML model, not necessarily an unfolded network, $\mathbf{x} = F(\mathbf{w};\mathbf{y})$, with the training setup mentioned in Section \ref{Problem_Formulation}, where $\mathbf{y}\in \mathbb{R}^{n\times 1}$, $\mathbf{x}\in \mathbb{R}^{m\times 1}$, and $\mathbf{w}\in\mathbb{R}^{P\times 1}$.
Further, assume that the model is $L_\mathcal{F}$-Lipschitz continuous and $\beta_\mathcal{F}$-smooth.
A function $\mathcal{F}(\cdot):\mathbb{R}^{P}\xrightarrow{}\mathbb{R}^{m\times T}$ is $L_\mathcal{F}$-Lipschitz continuous if 
\begin{equation*}
    \|\mathcal{F}(\mathbf{w}_1)-\mathcal{F}(\mathbf{w}_2)\|_F\leq L_\mathcal{F}\|\mathbf{w}_1-\mathbf{w}_2\|,\ \forall\mathbf{w}_1, \mathbf{w}_2\in \mathbb{R}^P,
\end{equation*}
and is $\beta_\mathcal{F}$-smooth if the gradient of the function is $\beta_\mathcal{F}$-Lipschitz, i.e.,
\begin{equation*}
    \|\nabla_\mathbf{w}\mathcal{F}(\mathbf{w_1})-\nabla_\mathbf{w}\mathcal{F}(\mathbf{w_2})\|_F\leq \beta_\mathcal{F}\|\mathbf{w_1}-\mathbf{w_2}\|,
\end{equation*}
$\forall\mathbf{w}_1,\ \mathbf{w}_2\in \mathbb{R}^P$. 
The Hessian spectral norm of $\mathcal{F}(\cdot)$ is defined as 
\begin{equation*}
    \|\mathbf{H}_\mathcal{F}(\mathbf{w})\| = \underset{i\in[T]}{\text{max}}\|\mathbf{H}_{\mathcal{F}_i}(\mathbf{w})\|,
\end{equation*}
where $\mathbf{H}_\mathcal{F}\in\mathbb{R}^{T\times m\times P\times P}$ is a tensor with $(\mathbf{H}_{\mathcal{F}})_{i,j,k,l} = \frac{\partial^2\left(\mathcal{F}(\mathbf{w})\right)_{j,i}}{\partial w_k \partial w_l}$ and $\mathbf{H}_{\mathcal{F}_i} = \frac{\partial^2(\mathcal{F}(\mathbf{w}))_i}{\partial \mathbf{w}^2}$.
As stated earlier, the loss landscape of the OP model typically satisfies PL$^*$ on most of the parameter space.
Formally, the PL$^*$ condition is defined as follows \cite{PL,PL1}:
\begin{definition}
Consider a set $C\subset\mathbb{R}^{P\times 1}$ and $\mu>0$. Then, a non-negative function $L(\mathbf{w})$ satisfies $\mu$-PL$^*$ condition on $C$ if $\|\nabla_\mathbf{w} L(\mathbf{w})\|^2\geq \mu L(\mathbf{w}),\ \forall\mathbf{w}\in C$.
\end{definition}
\begin{definition}
The tangent kernel matrix, $[K(\mathbf{w})]_{mT\times mT}$, of the function $\mathcal{F}(\mathbf{w})$, 
% given in \eqref{Loss_function},
is a block matrix with $(i,j)^\text{th}$ block defined as $$(K(\mathbf{w}))_{i,j} = \left[\nabla_\mathbf{w} \mathbf{f}_i\right]_{m\times P}\left[\nabla_\mathbf{w} \mathbf{f}_j\right]_{P\times m}^T,\ i\in[T] \text{ and }j\in[T].$$
\end{definition}
From the above definitions, we have the following lemma, which is called $\mu$-uniform conditioning \cite{LossLandscape} of a multi-output model $\mathcal{F}(\mathbf{w})$:
% Satisfying $\mu$-PL$^*$ condition on set $C$ also implies the uniform conditioning of $\mathcal{F}(\mathbf{w})$. In particular, we have
\begin{Lemma}
    $\mathcal{F}(\mathbf{w})$ satisfies $\mu$-PL$^*$ on set $C$ if the minimum eigenvalue of the tangent kernel matrix, $K(\mathbf{w})$, is greater than or equal to $\mu$, i.e., $\lambda_\text{min}(K(\mathbf{w}))\geq\mu,\ \forall \mathbf{w}\in C$. 
\label{Lemma:Uniform_Conditioning}
\end{Lemma}
\begin{proof} 
\vspace{-\topsep}
From \eqref{Loss_function}, we have
$\begin{aligned}
    \|\nabla_\mathbf{w} L(\mathbf{w})\|^2 &= \left[\hat{\mathbf{f}}-\hat{\mathbf{x}}\right]^T \left[\nabla_\mathbf{w}\hat{\mathbf{f}}\right]_{mT\times P} \left[\nabla_\mathbf{w}\hat{\mathbf{f}}\right]^T_{P\times mT}\left[\hat{\mathbf{f}}-\hat{\mathbf{x}}\right]\\
    & = \left[\hat{\mathbf{f}}-\hat{\mathbf{x}}\right]^T \left[K(\mathbf{w})\right]_{mT\times mT}\left[\hat{\mathbf{f}}-\hat{\mathbf{x}}\right],
\end{aligned}$
where $\hat{\mathbf{f}} = \text{Vec}\left(\mathcal{F}(\mathbf{w})\right)$ and $\hat{\mathbf{x}} = \text{Vec}\left(X\right)$.
The above equation can be lower-bounded as 
\begin{equation*}
\|\nabla_\mathbf{w} L(\mathbf{w})\|^2\geq \lambda_{\text{min}}\left(K(\mathbf{w})\right)\|\hat{\mathbf{f}}-\hat{\mathbf{x}}\|_2^2\geq \mu L(\mathbf{w}).
\end{equation*}
\end{proof}

Observe that $K(\mathbf{w})$ is a positive semi-definite matrix. Thus, a necessary condition to satisfy the PL$^*$ condition (that is, a necessary condition to obtain a full rank $K(\mathbf{w})$), for a multi-output model is $P\gg mT$.
For a scalar output model, the equivalent condition is  $P\gg T$ \cite{LossLandscape}. Note that if $P\ll T$, i.e., an UP model with a scalar output, then $\lambda_\text{min}(K(\mathbf{w})) = 0$, implies that an UP model does not satisfy the PL$^*$ condition.
% An important observation is that for an ML model with scalar output we should consider $P\gg T$ such that $K(\mathbf{w})$ is a full rank matrix \cite{LossLandscape}.  Whereas for a multi-output (say $m$) model we should consider $P\gg mT$ such that $K(\mathbf{w})$ is a full rank matrix, which implies that the model satisfies $\mu$-PL$^*$ condition.
% Note that $K(\mathbf{w})$ is a positive semi-definite matrix. Moreover, we should consider $P\gg mT$ such that $K(\mathbf{w})$ is a full rank matrix.

Practically, computing $\lambda_\text{min}(K(\mathbf{w}))$ for every $\mathbf{w}\in C$, to verify the PL$^*$ condition, is not feasible. 
One can overcome this by using the Hessian spectral norm of the model $\|\mathbf{H}_\mathcal{F}(\mathbf{w})\|$ \cite{LossLandscape}:
% To overcome this, the authors in \cite{LossLandscape} proposed a way to verify the PL$^*$ condition via the Hessian spectral norm of the model. Hence, motivated by this,  
\begin{theorem}
Let $\mathbf{w}_0\in\mathbb{R}^{P\times 1}$ be the parameter initialization of an $L_\mathcal{F}$-Lipschitz and $\beta_\mathcal{F}$-smooth model $\mathcal{F}(\mathbf{w})$, and $B(\mathbf{w}_0,R) = \{\mathbf{w}|\ \|\mathbf{w}-\mathbf{w}_0\|\leq R\}$ be a ball with radius $R>0$. Assume that $K(\mathbf{w}_0)$ is well conditioned, i.e., $\lambda_\text{min}(K(\mathbf{w}_0))=\lambda_0$ for some $\lambda_0>0$.
If $\|\mathbf{H}_\mathcal{F}(\mathbf{w})\| \leq \frac{\lambda_0 - \mu}{2{L_\mathcal{F}}\sqrt{T}R}$ for all $\mathbf{w}\in B(\mathbf{w}_0,R)$, then the model satisfies $\mu$-uniform conditioning in $B(\mathbf{w}_0,R)$; this also implies that $L(\mathbf{w})$ satisfies $\mu$-PL$^*$ in the ball $B(\mathbf{w}_0,R)$.
\label{PL_Condition}
\end{theorem}
The intuition behind the above theorem is that small $\|\mathbf{H}_\mathcal{F}(\mathbf{w})\|$ leads to a small change in the tangent kernel.
Precisely, if the tangent kernel is well conditioned at the initialization, then a small $\|\mathbf{H}_\mathcal{F}(\mathbf{w})\|$ in  $B(\mathbf{w}_0,R)$ guarantees that the tangent kernel is well conditioned within $B(\mathbf{w}_0,R)$.
The following theorem states that satisfying PL$^*$ guarantees the existence of a global minimum and exponential convergence to the global minimum from $\mathbf{w}_0$ using GD:
\begin{theorem}
Consider a model $\mathcal{F}(\mathbf{w})$ that is $L_\mathcal{F}$-Lipschitz continuous and $\beta_\mathcal{F}$-smooth. If the square loss function $L(\mathbf{w})$ satisfies the $\mu$-PL$^*$ condition in $B(\mathbf{w}_0,R)$ with $R = \frac{2L_\mathcal{F}\|\mathcal{F}(\mathbf{w}_0)-X\|_F}{\mu} = O\left(\frac{1}{\mu}\right)$, then we have the following:
\begin{itemize}
    \item There exist a global minimum, $\mathbf{w}^*$, in $B(\mathbf{w}_0,R)$ such that $\mathcal{F}(\mathbf{w}^*) = X$.
    \item GD with step size $\eta\leq \frac{1}{L_\mathcal{F}^2 + \beta_\mathcal{F} \|\mathcal{F}(\mathbf{w}_0)-X\|_F}$ converges to a global minimum at an exponential convergence rate, specifically,
$L(\mathbf{w}_t)\leq (1-\eta\mu)^t L(\mathbf{w}_0).$
\end{itemize}
\label{PL_Convergence}
\end{theorem}
The proofs of Theorems \ref{PL_Condition} and \ref{PL_Convergence} are similar to the proofs of Theorems $2$ and $6$, respectively, in \cite{LossLandscape}. 
However, as linear inverse problems deal with vector recovery,
% we are dealing with the multi-output model in a linear inverse problem,
the proofs rely on Frobenius norms instead of Euclidean norms. 
% \textcolor{red}{can you elaborate on how this affects proofs e.g. in terms of technical complexity?}
% at places we switch from the Euclidean norm to the Frobenius norm in the proofs.
% gradient of model output for a particular training sample $i$, i.e., $\nabla_\mathbf{w} \mathbf{f}_i$, is a matrix here, whereas it is a vector in \cite{LossLandscape}, accordingly we switch from Euclidean norm to Frobenius norm in the proof.
\section{Optimization Guarantees}
\label{Optimization_Guarantees}
% \section{Hessian Spectral Norm of Unfolded ISTA and ADMM Networks}
% In \cite{Linearity} the authors derived the Hessian spectral norm of a fully connected deep neural network with a scalar output.
We now analyze the optimization guarantees of both LISTA and ADMM-CSNet by considering them in the OP regime. 
% In general, this is the case in an unfolded network, \textcolor{blue}{as it usually needs less data than the standard network.}
Hence, the aim is further simplified to study under what conditions LISTA and ADMM-CSNet satisfy the PL$^*$ condition.
% Hence, the aim is further refined to explore the conditions under which LISTA and ADMM-CSNet satisfy the PL$^*$ condition.
As mentioned in Theorem \ref{PL_Condition}, one can verify the PL$^*$ condition using the Hessian spectral norm of the network. Thus, in this section, we first compute the Hessian spectral norm of both LISTA and ADMM-CSNet.
The mathematical analysis performed here is motivated by \cite{Linearity}, where the authors derived the Hessian spectral norm of an FFNN with a scalar output.
Then, we provide the conditions on both the network width and the number of training samples to hold the PL$^*$ condition.
Subsequently, we provide a comparative analysis among unfolded networks and FFNN to evaluate the threshold on the number of training samples.

\subsection{Assumptions}
\label{Assumptions}
For the analysis, we consider certain assumptions on the unfolded ISTA and ADMM networks.
The inputs of the networks are bounded, i.e., there exist some constants $C_x$, $C_u$, $C_z$, and $C_y$ such that $|x_i^0|\leq C_x,$ $|u_i^0|\leq C_u$, $|z_i^0|\leq C_z$, $\forall i\in[m]$, and $|y_i|\leq C_y,\ \forall i\in[n]$.
As the computation of the Hessian spectral norm involves a second-order derivative, we approximate the soft-thresholding activation function, $S_\lambda(\cdot)$, in the unfolded network with the double-differentiable/smooth soft-thresholding activation function, $\sigma_\lambda(\cdot)$, formulated using soft-plus,
% \footnote{Henceforth, unless otherwise specified the activation function in both LISTA and ADMM-CSNet is a soft-plus-based smooth soft-thresholding.},
where
$\sigma_\lambda(x) = \log\left(1+e^{x-\lambda}\right) - \log\left(1+e^{-x-\lambda}\right).$
\begin{figure}[t!]
    \centering
    \includegraphics[height = 5cm, width = 7.5cm]{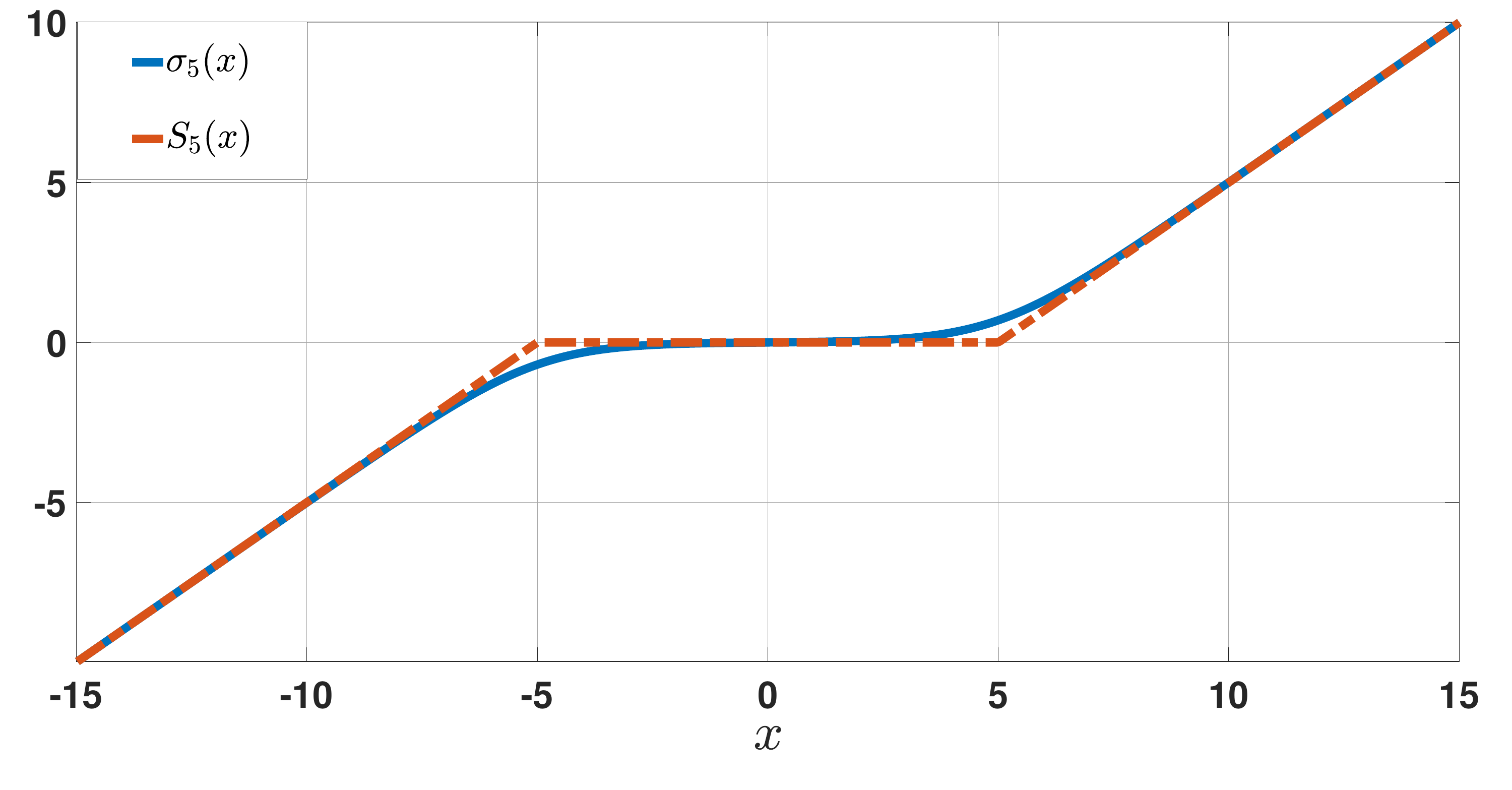}
    \caption{Soft-threshold function, $S_\lambda(x)$, and its smooth approximation, $\sigma_\lambda(x)$ (formulated using the soft-plus function), with $\lambda=5$.}
    \label{fig:Soft_thr_vs_soft_plus}
\end{figure}
Fig. \ref{fig:Soft_thr_vs_soft_plus} depicts $S_\lambda(x)$ and $\sigma_\lambda(x)$ for $\lambda=5$. Observe that $\sigma_\lambda(x)$ approximates well to the shape of $S_\lambda(x)$. 
There are several works in the literature that approximate the soft-thresholding function with a smooth version of it \cite{SmoothActivation1, SmoothActivation2, SmoothActivation3, SmoothActivation4, SmoothActivation5, SmoothActivation6, SmoothActivation7}.
% Though we considered a smooth function constructed using the soft-plus, one can use any smooth function that well approximates soft-thresholding, because, the analysis proposed in this work is independent of the function type.
The analysis proposed in this work can be extended as is to other smooth approximations.
Further, since $\lambda$ is assumed to be a constant (refer to Section \ref{Problem_Formulation}), henceforth, we write $\sigma_\lambda(\cdot)$ as $\sigma(\cdot)$.
It is well known that $\sigma(\cdot)$ is $L_\sigma$-Lipschitz continuous and $\beta_\sigma$-smooth.
% This implies, $\mathbf{w}_{P\times 1} = \text{Vec}\left([\mathbf{W}]_{L\times m \times (m+n)}\right)$ is the learnable parameter vector in an $L$-layer unfolded network, where
% \begin{equation}
%     \mathbf{W} = \left[W^1\ W^2\ \dots\ W^L\right], 
% \end{equation}
% and $\left[W^l\right]_{m\times (m+n)} = \left[W_1^l\ \ W_2^l\right]$ is the parameter matrix corresponds to $l^\text{th}$-layer. Alternatively, we can write 
% \begin{equation}
%     \mathbf{W} = \left[\left[\mathbf{W}_1\right]_{L\times m\times n}\ \ \left[\mathbf{W}_2\right]_{L\times m\times m}\right],
% \end{equation}
% $\mathbf{W}_1 = \left[W_1^1\ \dots\ W_1^L\right]$ and
% $\mathbf{W}_2 = \left[W_2^1\  \dots\ W_2^L\right]$.
% \begin{equation}
%     \mathbf{W} = \left[\left[\mathbf{W}_1\right]_{L\times m\times n}\ \ \left[\mathbf{W}_2\right]_{L\times m\times m}\right],
% \end{equation}
% $\mathbf{W}_1 = \left\{W_1^1, \dots, W_1^L\right\}$ and
% $\mathbf{W}_2 = \left\{W_2^1, \dots, W_2^L\right\}$. Alternatively, we can write
% \begin{equation}
%     \mathbf{W} = \left\{W^1, W^2,\dots, W^L\right\}, 
% \end{equation}
% where $\left[W^l\right]_{m\times (m+n)} = \left[W_1^l\ \ W_2^l\right]$ is the parameter matrix corresponds to $l^\text{th}$-layer.

Let $\mathbf{W}_{0}, \mathbf{W}_{10}, \mathbf{W}_{20}, W_{10}^{l}$ and $W_{20}^{l}$ denote the initialization of $\mathbf{W}, \mathbf{W}_{1}, \mathbf{W}_{2}$, $W_{1}^{l}$ and $W_{2}^{l}$, respectively.
We initialize each parameter using random Gaussian initialization with mean $0$ and variance $1$, i.e., $\left(W_{10}^{l}\right)_{i, j} \sim \mathcal{N}(0,1)$ and $\left(W_{20}^{l}\right)_{i, j} \sim \mathcal{N}(0,1)$, $\forall l\in[L]$.
This guarantees well conditioning of the tangent kernel at initialization \cite{GD1, LossLandscape}.
Moreover, the Gaussian initialization imposes certain bounds, with high probability, on the spectral norm of the weight matrices. In particular, we have the following:
\begin{Lemma}
    If $\left(W_{10}^{l}\right)_{i, j} \sim \mathcal{N}(0,1)$ and $\left(W_{20}^{l}\right)_{i, j} \sim \mathcal{N}(0,1)$, $\forall l\in[L]$, then with probability at least $1-2 \exp \left(-\frac{m}{2}\right)$ we have $\left\|W_{10}^{l}\right\| \leq c_{10} \sqrt{n} = O(\sqrt{n})$ and $\left\|W_{20}^{l}\right\| \leq c_{20} \sqrt{m} = O(\sqrt{m})$, $ \forall l \in [L]$, where   $c_{10}=1+2 \sqrt{m} / \sqrt{n}$ and $c_{20}=3$.
    \label{BoundOnWeightInitialization}
\end{Lemma}
\begin{proof}
Any matrix ${W}\in \mathbb{R}^{m_1\times m_2}$ with Gaussian initialization satisfies the following inequality with probability at least $1-2\exp\left(-\frac{t^{2}}{2}\right)$, where $t\geq 0$, \cite{WeightBound}:
$\|W\| \leq \sqrt{m_{1}}+\sqrt{m_{2}}+t $.
Using this fact and considering $t=\sqrt{m}$, we get $\|W_{10}^{l}\| = O(\sqrt{n})$ and $\|W_{20}^{l}\| = O(\sqrt{m})$.
\end{proof}
% \vspace{-1cm}
% Let $m=20$ and $n=10$, then with probability $0.9999$ we have $\|W_{10}\|_2\leq O(\sqrt{10})$ and $\|W_{20}\|_2\leq O(\sqrt{20})$.
The following lemma shows that the spectral norm of the weight matrices within a finite radius ball is of the same order as at the initialization.
\begin{Lemma}
    If $\mathbf{W}_{10}$ and $\mathbf{W}_{20}$ are initialized as stated in Lemma \ref{BoundOnWeightInitialization}, then for any $\mathbf{W}_1\in B(\mathbf{W}_{10},R_1)$ and $\mathbf{W}_2\in B(\mathbf{W}_{20},R_2)$, where $R_1$ and  $R_2$ are positive scalars, we have $\left\|W_{1}^{l}\right\|= O(\sqrt{n})$ and $\left\|W_{2}^{l}\right\| = O(\sqrt{m})$, $ \forall l \in [L]$.
    \label{BoundOnWeight}
\end{Lemma}
\begin{proof}
    From triangular inequality, we have 
    \begin{equation*}
    \footnotesize
    \begin{aligned}    &\left\|W_{1}^{l}\right\| \leq\left\|W_{10}^{l}\right\|+\left\|W_{1}^{l}-W_{10}^{l}\right\|_{F} \leq c_{10} \sqrt{n}+R_{1}= O(\sqrt{n}),\\   &\left\|W_{2}^{l}\right\| \leq\left\|W_{20}^{l}\right\| +\left\|W_{2}^{l}-W_{20}^{l}\right\|_{F} \leq c_{20} \sqrt{m}+R_{2} = O(\sqrt{m}).
    \end{aligned} 
    \end{equation*}
\end{proof}
% \vspace{-1cm}
% \begin{equation}
%     \mathbf{W} = \left\{W^1, W^2,\dots, W^L\right\}, 
% \end{equation}
% and $W^l = \left[W_1^l\ \ W_2^l\right]$ denotes the parameter matrix corresponds to $l^\text{th}$-layer. 
% Alternatively, we can write $\mathbf{W} = \left[\mathbf{W}_1\ \ \mathbf{W}_2\right]$, 
% where
% $\mathbf{W}_1 = \left\{W_1^1, W_1^2,\dots, W_1^L\right\}$ and
% $\mathbf{W}_2 = \left\{W_2^1, W_2^2,\dots, W_2^L\right\}$.
% \begin{equation}
% [\mathbf{W}]=\left[\begin{array}{cc}[\mathbf{W}_1]_{L\times m\times n} & [\mathbf{W}_2]_{L\times m\times m}
% \end{array}\right].
% \end{equation}

As the width of the network can be very high (dimension of the target vector), to obtain the constant asymptotic behavior, the learnable parameters  $W_1^l$ and $W_2^l$ are normalized by $\frac{1}{\sqrt{n}}$ and $\frac{1}{\sqrt{m}}$, respectively, and the output of the model is normalized by $\frac{1}{\sqrt{m}}$. This way of normalization is called neural tangent kernel (NTK) parameterization \cite{TangentKernel, NEURIPS2019_0d1a9651}.
With these assumptions, the output of a finite $L$-layer LISTA network is 
\begin{equation}
\mathbf{f}=\frac{1}{\sqrt{m}} \mathbf{x}^L, 
\label{LISTA_Model}
\end{equation}
where
\begin{equation*}
\begin{aligned}
    &\mathbf{x}^{l} = \sigma(\tilde{\mathbf{x}}^{l}) = \sigma\left(\frac{W_{1}^{l}}{\sqrt{n}} \mathbf{y}+\frac{W_{2}^{l}}{\sqrt{m}} \mathbf{x}^{l-1}\right)\ \in \mathbb{R}^{m\times 1},\ l \in [L].
\end{aligned}
\end{equation*}
Likewise, the output of a finite $L$-layer ADMM-CSNet is
\begin{equation}
\mathbf{f}=\frac{1}{\sqrt{m}} \mathbf{z}^L,
\label{ADMM_CSNet_Model}
\end{equation}
where
\begin{equation*}
\begin{aligned}    
     &\mathbf{z}^l = \sigma\left(\tilde{\mathbf{z}}^{l}\right) =\sigma\left( \mathbf{x}^l+\mathbf{u}^{l-1} \right),\\
     &\mathbf{x}^{l} = \frac{1}{\sqrt{n}} W_1^l\mathbf{y} + \frac{1}{\sqrt{m}} W_2^l\left(\mathbf{z}^{l-1}-\mathbf{u}^{l-1}\right), \\
     &
     \mathbf{u}^l = \mathbf{u}^{l-1}+\left(\mathbf{x}^l-\mathbf{z}^l\right), \ l \in [ L].
\end{aligned}
\end{equation*}
To maintain uniformity in notation, hereafter, we denote the output of the network as $\mathbf{f}=\frac{1}{\sqrt{m}} \mathbf{g}^L$, where $\mathbf{g}^l = \mathbf{x}^l$ for LISTA and $\mathbf{g}^l = \mathbf{z}^l$ for ADMM-CSNet.
% the $l^\text{th}$-layer output of unfolded ISTA and ADMM networks is
% \begin{equation}
%         \mathbf{x}^{l} = \sigma\left(\frac{1}{\sqrt{m}}{W}_2^l\mathbf{x}^{l-1} + \frac{1}{\sqrt{n}} W_1^l\mathbf{y}\right),
%         \label{UpdatedISTA}
% \end{equation}
% and 
% \begin{equation}
%     \begin{aligned}
%     \mathbf{x}^{l} = \frac{1}{\sqrt{n}} W_1^l\mathbf{y} + &\frac{1}{\sqrt{m}} W_2^l\left(\mathbf{z}^{l-1}-\mathbf{u}^{l-1}\right), \\
%      \mathbf{z}^l = \sigma\left(\mathbf{x}^l+\mathbf{u}^{l-1}\right), &\  \mathbf{u}^l = \mathbf{u}^{l-1}+\left(\mathbf{x}^l-\mathbf{z}^l\right).
%      \label{UpdatedADMM}
%      \end{aligned}
% \end{equation}

\subsection{Hessian Spectral Norm}
\label{Hessian_Spectral_Norm}
For better understanding, we first compute the Hessian spectral norm of one layer, i.e., $L=1$, unfolded network. 
% Later, we generalize the analysis to an $L$-layer network.
\subsubsection{Analysis of $1$-Layer Unfolded Network}
The Hessian matrix of a $1$-layer LISTA or ADMM-CSNet for a given training sample $i$ is\footnote{Note that, to simplify the notation, we denoted $\mathbf{H}_{\mathcal{F}_i}$ as $\mathbf{H}$.} 
\begin{equation}
\left[\mathbf{H}_{\mathcal{F}_i}\right]= \left[\mathbf{H}\right]_{m\times P\times P}=\left[\begin{array}{llll}
{H}_{1} & {H}_{2} & \cdots & {H}_{m}
\end{array}\right],
\label{Hessian_1_Layer}
\end{equation}
where $[H_s]_{P\times P} = \frac{\partial^{2} f_{s}}{\partial \mathbf{w}^2}$, $\mathbf{w} = \text{Vec}(W^1) = \text{Vec}\left([W_1^1, W_2^1]\right)$, $f_s$ denotes the $s^\text{th}$ component in the network output vector $\mathbf{f}$, i.e., $f_s = \frac{1}{\sqrt{m}}\mathbf{v}_s^T\mathbf{g}^1$, and $\mathbf{v}_s$ is a vector with $s^\text{th}$ element set to be $1$ and others to be $0$.
The Hessian spectral norm given in \eqref{Hessian_1_Layer} can be bounded as $\underset{s\in[m]}{\text{max}}\left\{\left\|{H}_{s}\right\|\right\}\leq \|\mathbf{H}\| \leq \sum_{s}\left\|{H}_{s}\right\|$. By leveraging the chain rule, we have
% one can decompose the matrix $H_s$ as 
\begin{equation}
    H_s = \frac{\partial f_s}{\partial \mathbf{g}^1} \frac{\partial^2 \mathbf{g}^1}{\partial \mathbf{w}^2}.
\end{equation}
We can bound $H_s$, as given below, by using the inequality given in \eqref{Holders_inequality},
\begin{equation}
    \|H_s\| \leq \left\|\frac{\partial f_s}{\partial \mathbf{g}^1}\right\|_\infty \left\|\frac{\partial^2 \mathbf{g}^1}{\partial \mathbf{w}^2}\right\|_{2,2,1}.
\end{equation}
From \eqref{LISTA_Model} or \eqref{ADMM_CSNet_Model}, we get 
\begin{equation}
    \left\|\frac{\partial f_s}{\partial \mathbf{g}^1}\right\|_\infty  = \left\|\frac{1}{\sqrt{m}}\mathbf{v}_s^T\right\|_\infty = O\left(\frac{1}{\sqrt{m}}\right).
\end{equation}
In addition,
\begin{equation}
    \begin{aligned}
         &\left\|\frac{\partial^{2} \mathbf{g}^{1}}{\left(\partial \mathbf{w}\right)^{2}}\right\|_{2,2,1}=\left\|\left[\begin{array}{cc}\partial^{2} \mathbf{g}^{1} /\left(\partial W_{1}^{1}\right)^{2} & \partial^{2} \mathbf{g}^{1} / \partial W_{1}^{1} \partial W_{2}^{1} \\\partial^{2} \mathbf{g}^{1} / \partial W_{2}^{1} \partial W_{1}^{1} & \partial^{2} \mathbf{g}^{1} /\left(\partial W_{2}^{1}\right)^{2}\end{array}\right]\right\|_{2,2,1} \\
        &\leq\left\|\frac{\partial^{2} \mathbf{g}^{1}}{\left(\partial W_{1}^{1}\right)^{2}}\right\|_{2,2,1}+2\left\|\frac{\partial^{2} \mathbf{g}^{1}}{\partial W_{1}^{1} \partial W_{2}^{1}}\right\|_{2,2,1}+\left\|\frac{\partial^{2} \mathbf{g}^{1}}{\left(\partial W_{2}^{1}\right)^{2}}\right\|_{2,2,1}.
    \end{aligned}
    \label{2_2_1_norm_bound}
\end{equation}
We now compute the $(2,2,1)$-norms in the above equation for both LISTA and ADMM-CSNet. To begin with, for LISTA, 
we have the following second-order partial derivatives of layer-wise output, $\mathbf{g}^1$, w.r.t. parameters:
\begin{equation*}
    \begin{aligned}
        \left(\frac{\partial^{2} \mathbf{g}^{1}}{\left(\partial W_{1}^{1}\right)^{2}}\right)_{i, j j^{\prime}, k k^{\prime}}&= \frac{\partial^2 \mathbf{x}_i^1}{\partial(W_1^1)_{jj^\prime}\partial(W_1^1)_{kk^\prime}}\\
        &=\frac{1}{n} \sigma^{\prime \prime}\left(\tilde{\mathbf{x}}_{i}^{1}\right) \mathbf{y}_{j^{\prime}} \mathbf{y}_{k^{\prime}} \mathbb{I}_{i=k=j},
    \end{aligned}
\end{equation*}
\begin{equation*}
    \begin{aligned}
        \left(\frac{\partial^{2} \mathbf{g}^{1}}{\left(\partial W_{2}^{1}\right)^{2}}\right)_{i, j j^{\prime}, k k^{\prime}}=\frac{1}{m}  \sigma^{\prime \prime}\left(\tilde{\mathbf{x}}_{i}^{1}\right) \mathbf{x}_{j^{\prime}}^{0} \mathbf{x}_{k^{\prime}}^{0} \mathbb{I}_{i=k=j},
    \end{aligned}
\end{equation*}
\begin{equation*}
    \begin{aligned}
    \left(\frac{\partial^{2} \mathbf{g}^{1}}{\partial W_{2}^{1} \partial W_{1}^{1}}\right)_{i, j j^{\prime}, k k^{\prime}}=\frac{1}{\sqrt{m n}} \sigma^{\prime \prime}\left(\tilde{\mathbf{x}}_{i}^{1}\right) \mathbf{x}_{j^{\prime}}^{0} \mathbf{y}_{k^{\prime}} \mathbb{I}_{i=k=j},
    \end{aligned}
\end{equation*}
where $\mathbb{I}_{\{\}}$ denotes the indicator function.
By utilizing the definition of $(2,2,1)$-norm given in \eqref{2_2_2_Norm}, bounds on inputs of the network, and smoothness of the activation function, the $(2,2,1)$-norms of the above quantities are obtained as shown below:
% By using the definition of $(2,2,1)$-norm given in \eqref{2_2_2_Norm}, bounds on the inputs of the network, and smoothness of the activation function, we get
\begin{equation*}
\footnotesize
\begin{aligned}
 \left\|\frac{\partial^{2} \mathbf{g}^{1}}{\left(\partial W_{1}^{1}\right)^{2}}\right\|_{2,2,1}&=\sup _{\left\|V_{1}\right\|_F=\left\|V_{2}\right\|_F=1} \frac{1}{n} \sum_{i=1}^{m}\left|\sigma^{\prime \prime}\left(\tilde{\mathbf{x}}_{i}^{1}\right)\left(V_{1} \mathbf{y}\right)_{i}\left(V_{2} \mathbf{y}\right)_{i}\right| \\
& \leq \sup _{\left\|V_{1}\right\|_F=\left\|V_{2}\right\|_F=1} \frac{1}{2 n} \beta_{\sigma}\left(\left\|V_{1} \mathbf{y}\right\|^{2}+\left\|V_{2} \mathbf{y}\right\|^{2}\right) \\
& \leq \frac{1}{2 n} \beta_{\sigma}\left(\|\mathbf{y}\|^{2}+\|\mathbf{y}\|^{2}\right)\leq\beta_{\sigma} C_{y}^{2} = O(1) \\
\end{aligned}    
\end{equation*}
\begin{equation*}
\footnotesize
\begin{aligned}
\left\|\frac{\partial^{2} \mathbf{g}^{1}}{\left(\partial W_{2}^{1}\right)^{2}}\right\|_{2,2,1}&=\sup _{\left\|V_{1}\right\|_F=\left\|V_{2}\right\|_F=1} \frac{1}{m} \sum_{i=1}^{m}\left|\sigma^{\prime \prime}\left(\tilde{\mathbf{x}}_i^{1}\right)\left(V_{1}\mathbf{x}^{0}\right)_{i}\left(V_{2}\mathbf{x}^{0}\right)_{i}\right|\\
& \leq \frac{1}{2 m} \beta_{\sigma}\left(\left\|\mathbf{x}^{0}\right\|^{2}+\left\|\mathbf{x}^{0}\right\|^{2}\right) 
 \leq \beta_{\sigma}{C_x}^{2} = O(1)
\end{aligned}    
\end{equation*}
\begin{equation*}
\footnotesize
\begin{aligned}
&\left\|\frac{\partial^{2} \mathbf{g}^{1}}{\partial W_{2}^{1} \partial W_{1}^{1}}\right\|_{2,2,1}\\ &=\sup _{\left\|V_{1}\right\|_F=\left\|V_{2}\right\|_F=1} \frac{1}{\sqrt{m n}} \sum_{i=1}^{m}\left|\sigma^{\prime \prime}\left(\tilde{\mathbf{x}}_{i}^{l}\right)\left(V_{1} \mathbf{x}_{i}^{0}\right)_{i}\left(V_{2} \mathbf{y}\right)_{i}\right| \\
& \leq \frac{1}{2 \sqrt{m n}} \beta_{\sigma}\left(\left\|\mathbf{x}^{0}\right\|^{2}+\|\mathbf{y}\|^{2}\right) \leq \sqrt{\frac{m}{4n}} \beta_{\sigma} C_{x}^{2}+\sqrt{\frac{n}{4 m}} \beta_{\sigma} C_{y}^{2} = O(1).
\end{aligned}    
\end{equation*}
Substituting the above bounds in \eqref{2_2_1_norm_bound} implies $\left\|\frac{\partial^{2} \mathbf{g}^{1}}{\left(\partial W^{1}\right)^{2}}\right\|_{2,2,1}= O(1)$. 

Similarly, for ADMM-CSNet, the equivalent second-order partial derivatives are
\begin{equation*}
\footnotesize
    \begin{aligned}
    &\left(\frac{\partial^{2} \mathbf{g}^{1}}{\left(\partial W_{1}^{1}\right)^{2}}\right)_{i, j j^{\prime}, k k^{\prime}}=\frac{1}{n} \sigma^{\prime \prime}\left(\tilde{\mathbf{z}}_{i}^{1}\right) \mathbf{y}_{j^{\prime}} \mathbf{y}_{k^{\prime}} \mathbb{I}_{i=k=j},\\
    &\left(\frac{\partial^{2} \mathbf{g}^{1}}{\left(\partial W_{2}^{1}\right)^{2}}\right)_{i, j j^{\prime}, k k^{\prime}}=\frac{1}{m} \sigma^{\prime \prime}\left(\tilde{\mathbf{z}}_{i}^{1}\right) (\mathbf{z}^{0}-\mathbf{u}^{0})_{j^{\prime}} (\mathbf{z}^{0}-\mathbf{u}^{0})_{k^{\prime}} \mathbb{I}_{i=k=j}, \\
    &\left(\frac{\partial^{2} \mathbf{g}^{1}}{\partial W_{2}^{1} \partial W_{1}^{1}}\right)_{i, j j^{\prime}, k k^{\prime}}=\frac{1}{\sqrt{mn}} \sigma^{\prime \prime}\left(\tilde{\mathbf{z}}_{i}^{1}\right) (\mathbf{z}^{0}-\mathbf{u}^{0})_{j^{\prime}} \mathbf{y}_{k^{\prime}} \mathbb{I}_{i=k=j}.
\end{aligned}
\end{equation*}
The corresponding $(2,2,1)$-norm bounds are
\begin{equation*}
        \left\|\frac{\partial^{2} \mathbf{g}^{1}}{\left(\partial W_{1}^{1}\right)^{2}}\right\|_{2,2,1} \leq \frac{1}{2n} \beta_{\sigma}\left(\|\mathbf{y}\|^{2}+\|\mathbf{y}\|^{2}\right) \leq \beta_{\sigma} C_{y}^{2} = O(1), 
\end{equation*}
\begin{equation*}
         \left\|\frac{\partial^{2} \mathbf{g}^{1}}{\left(\partial W_{2}^{1}\right)^{2}}\right\|_{2,2,1}
    \leq \frac{1}{2 m} \beta_{\sigma}\left(2 m C_z^2 +2m C_u^2\right)  = O(1),
\end{equation*}
\begin{equation*}
        \left\|\frac{\partial^{2} \mathbf{g}^{1}}{\partial W_{1}^{1} \partial W_{2}^{1}}\right\|_{2,2,1}  \leq \beta_\sigma \sqrt{\frac{m}{4 n}}\left(C_y^2+\left(C_z+C_u\right)^2\right) =  O(1).
\end{equation*}
Using the above bounds, we get $\left\|\frac{\partial^{2} \mathbf{g}^{1}}{\left(\partial W^{1}\right)^{2}}\right\|_{2,2,1}= O(1)$.
From the above analysis, we conclude that the $(2,2,1)$-norm of the tensor, $\frac{\partial^{2} \mathbf{g}^{1}}{\left(\partial W^{1}\right)^{2}}$, is of the order of $O(1)$ and the $\infty$-norm of the vector, $\frac{\partial f_s}{\partial \mathbf{g}^1}$, is of the order of $O\left(\frac{1}{\sqrt{m}}\right)$.
This implies, 
\begin{equation}
    \|H_s\|  = O\left(\frac{1}{\sqrt{m}}\right) \text{ and } \|\mathbf{H}\| = \Omega\left(\frac{1}{\sqrt{m}}\right) = O\left(\sqrt{m}\right).
\end{equation}
Therefore, 
% for a random Gaussian initialization of parameters, 
the Hessian spectral norm of a 1-layer LISTA or ADMM-CSNet depends on the 
% in $O\left(\sqrt{m}\right)$,
% with high probability in a ball of a specific radius
% that is, it is 
width (dimension of the target vector) of the network. 
We now generalize the above analysis for an $L$-layer unfolded network.
\subsubsection{Analysis of L-Layer Unfolded Network}
The Hessian matrix of an $L$-layer unfolded ISTA or ADMM network for a given $i^{\text{th}}$ training sample is written as 
\begin{equation}
    \left[\mathbf{H}\right]_{m\times P\times P}=\left[\begin{array}{llll}{H}_{1} & {H}_{2} & \cdots & {H}_{m}
\end{array}\right],
\label{Hessian_matrix_L_Layer}
\end{equation}
where ${H}_{s}$ for $s\in[m]$ is
\begin{equation}
    \left[{H}_{s}\right]_{P\times P}=\left[\begin{array}{cccc}
{H}_{s}^{1,1} & {H}_{s}^{1,2} & \ldots & {H}_{s}^{1, L} \\
{H}_{s}^{2,1} & {H}_{s}^{2,2} & \cdots & {H}_{s}^{2, L} \\
\vdots & \vdots & \ddots & \vdots \\
{H}_{s}^{L, 1} & {H}_{s}^{L, 2} & \cdots & {H}_{s}^{L, L}
\end{array}\right],
\label{Hessian_sub_block_matrix}
\end{equation}
$\left[{H}_{s}^{l_{1}, l_{2}}\right]_{P_1\times P_1}=\frac{\partial^{2} f_{s}}{\partial \mathbf{w}^{l_{1}} \partial \mathbf{w}^{l_{2}}}$, where $P_1=m^2+mn$,  $l_1\in[L]$, $l_2 \in[L]$, $\mathbf{w}^{l} = \text{Vec}(W^l) = \text{Vec}\left([W_1^l\ W_2^l]\right)$ denotes the weights of $l^\text{th}$-layer, and $f_s = \frac{1}{\sqrt{m}}\mathbf{v}_s^T\mathbf{g}^L$.
% , and $f_s^L$ is the common symbol used to denote the $s^\text{th}$ component in the network output vector\footnote{Note that $\mathbf{f}^L = \mathbf{x}^L\in \mathbb{R}^{m\times 1}$ and $\mathbf{f}^L = \mathbf{z}^L\in \mathbb{R}^{m\times 1}$ be the outputs of $L^\text{th}$-layer in LISTA and ADMM-CSNet, respectively.}, $\mathbf{f}^L$. For LISTA $f_s^L = x_s^L$ and for ADMM-CSNet $f_s^L = z_s^L$.
From \eqref{Hessian_matrix_L_Layer} and \eqref{Hessian_sub_block_matrix}, the spectral norm of $\mathbf{H}$, $\|\mathbf{H}\|$, is bounded by its block-wise spectral norm, $\|H_s\|$, as stated in the following theorem:
% , i.e., $\underset{s\in[m]}{\text{max}}\left\{\left\|{H}_{s}\right\|\right\}\leq \|\mathbf{H}\| \leq \sum_{s \in[m]}\left\|{H}_{s}\right\|.$
% By computing this block-wise spectral norm, the $\|\mathbf{H}\|$ can be bounded as stated in the following theorem:
\begin{theorem}
    The Hessian spectral norm, $\|\mathbf{H}\|$, of an $L$-layer unfolded ISTA (ADMM) network, defined as in \eqref{LISTA_Model} (\eqref{ADMM_CSNet_Model}), is bounded as 
$\underset{s\in[m]}{\text{max}}\left\{\left\|{H}_{s}\right\|\right\}\leq \|\mathbf{H}\| \leq \sum_{s\in[m]} \left\|{H}_{s}\right\|,$
where 
\begin{equation}
\begin{aligned}
\left\|{H}_{s}\right\| \leq  \sum_{l_{1}, l_{2}}\left\|{H}_{s}^{l_{1}, l_{2}}\right\| &\leq \sum_{l_{1}, l_{2}} C_1 \mathcal{Q}_{2,2,1}\left(f_{s}\right) \mathcal{Q}_{\infty}\left(f_{s}\right)\\
    &\leq C \mathcal{Q}_{2,2,1}\left(f_{s}\right) \mathcal{Q}_{\infty}\left(f_{s}\right).
\end{aligned} 
\end{equation}
% \begin{equation}
%     \begin{aligned}        
%         \|\mathbf{H}\| &\leq \sum_{s, l_{1}, l_{2}} C_1 \mathcal{Q}_{2,2,1}\left(f_{s}\right) \mathcal{Q}_{\infty}\left(f_{s}\right)\\ &\leq \sum_{s=1}^{m} C \mathcal{Q}_{2,2,1}\left(f_{s}\right) \mathcal{Q}_{\infty}\left(f_{s}\right),
%     \end{aligned}
% \end{equation}
The constant $C_1$ depends on $L$ and $L_\sigma$, $C = L^2C_1$,
\begin{equation}
\mathcal{Q}_{\infty}\left(f_s\right)= \max _{1 \leq l \leq L}\left\{\left\|\frac{\partial f_s}{\partial \mathbf{g}^{l}}\right\|_{\infty}\right\},\text{ and }
    \label{Qinf}
\end{equation}
\begin{equation}
\begin{aligned}
\mathcal{Q}_{2,2,1}\left(f_s\right)=& \max _{1 \leq l_{1} \leq l_{2} < l_{3} \leq L}\Bigg\{\left\|\frac{\partial^{2} \mathbf{g}^{l_1}}{\left(\partial \mathbf{w}^{l_1}\right)^{2}}\right\|_{2,2,1},\\
&\left\|\frac{\partial \mathbf{g}^{l_1}}{\partial \mathbf{w}^{l_{1}}}\right\|\left\|\frac{\partial^{2} \mathbf{g}^{l_2}}{\partial \mathbf{g}^{\left(l_{2}-1\right)} \partial \mathbf{w}^{l_2}}\right\|_{2,2,1}, \\
&\left\|\frac{\partial \mathbf{g}^{l_{1}}}{\partial \mathbf{w}^{l_{1}}}\right\|\left\|\frac{\partial \mathbf{g}^{l_2}}{\partial \mathbf{w}^{l_2}}\right\|\left\|\frac{\partial^{2} \mathbf{g}^{l_{3}}}{\left(\partial \mathbf{g}^{l_{3}-1}\right)^{2}}\right\|_{2,2,1}\Bigg\}.
    \end{aligned} 
    \label{Q221}
\end{equation}
\label{Bound_Hessian_L_Layer}
\end{theorem}
Proof of the above theorem is given in the Appendix. Similar to $1$-layer case, the bound on $\|\mathbf{H}\|$ depends on the  $\infty$-norms of $\frac{\partial f_s}{\partial \mathbf{g}^{l}},\ l\in[L]$ and $(2,2,1)$-norms of layer-wise derivatives (basically these are order $3$ tensors).
We now aim to derive the bounds on the quantities $\mathcal{Q}_{2,2,1}\left(f_{s}\right)$ and $\mathcal{Q}_{\infty}\left(f_{s}\right)$ for both unfolded ISTA and ADMM networks.

Similar to Lemma \ref{BoundOnWeightInitialization} and \ref{BoundOnWeight}, the Gaussian initialization of the weight matrices imposes a bound on the hidden layer output of the unfolded network, which is stated in the following lemma:
\begin{Lemma}
If $\left(W_{10}^{l}\right)_{i, j} \sim \mathcal{N}(0,1)$ and $\left(W_{20}^{l}\right)_{i, j} \sim \mathcal{N}(0,1)$, $\forall l\in[L]$, then 
% If $\mathbf{W}_{10}$ and $\mathbf{W}_{20}$ satisfies Lemma \ref{BoundOnWeightInitialization}, then 
    for any $\mathbf{W}_1\in B(\mathbf{W}_{10},R_1)$ and $\mathbf{W}_2\in B(\mathbf{W}_{20},R_2)$, we have $\left\|\mathbf{x}^{l}\right\| \leq c_{\mathrm{ISTA};\mathbf{x}}^{l}$ for LISTA, and $\left\|\mathbf{z}^{l}\right\| \leq c_{\mathrm{ADMM};\mathbf{z}}^{l}$ and $\left\|\mathbf{u}^{l}\right\| \leq c_{\mathrm{ADMM;\mathbf{u}}}^{l}$ for ADMM-CSNet.
    The updating rules are
    \footnotesize
    \begin{equation*}
        \begin{aligned}
            c_{\mathrm{ISTA} ; \mathbf{x}}^{l}&=L_{\sigma}\left(c_{10}+\frac{R_{1}}{\sqrt{n}}\right) \sqrt{n} C_{y}+L_{\sigma}\left(c_{20}+\frac{R_{2}}{\sqrt{m}}\right) c_{\mathrm{ISTA} ; \mathbf{x}}^{l-1}+\sigma(0)\\
            & = O\left(\sqrt{m}\right)
        \end{aligned}
    \end{equation*}
    \begin{equation*}
    \begin{aligned}
    c_{\mathrm{ADMM} ; \mathbf{z}}^{l}&=L_{\sigma}\left(c_{10}+\frac{R_{1}}{\sqrt{n}}\right) \sqrt{n} C_{y}+L_{\sigma}\left(c_{20}+\frac{R_{2}}{\sqrt{m}}\right) c_{\mathrm{ADMM} ; \mathbf{z}}^{l-1}\\ 
    &+L_{\sigma}\left(1+c_{20}+\frac{R_{2}}{\sqrt{m}}\right) c_{\mathrm{ADMM} ; \mathbf{u}}^{l-1}+\sigma(0) = O\left(\sqrt{m}\right), \\
    c_{\mathrm{ADMM} ; \mathbf{u}}^{l}&=\left(c_{10}+\frac{R_{1}}{\sqrt{n}}\right) \sqrt{n} C_{y}+\left(c_{20}+\frac{R_{2}}{\sqrt{m}}\right) c_{\mathrm{ADMM} ; \mathbf{z}}^{l-1}\\
    &+\left(c_{20}+\frac{R_{2}}{\sqrt{m}}+1\right) c_{\mathrm{ADMM} ; \mathbf{u}}^{l-1}+ c_{\mathrm{ADMM} ; \mathbf{z}}^{l} = O\left(\sqrt{m}\right),
    \end{aligned}
    \end{equation*}
    \normalsize
   where $c_{\mathrm{ISTA};\mathbf{x}}^{0}=\sqrt{m} C_{x}$, $c_{\mathrm{ADMM};\mathbf{z}}^{0}=\sqrt{m} C_{z}$, $c_{\mathrm{ADMM};\mathbf{u}}^{0}=\sqrt{m} C_{u}$, $|x_i^0|\leq C_x,$ $|u_i^0|\leq C_u$, and $|z_i^0|\leq C_z$, $\forall i\in[m]$. 
   \label{BoundOnHiddenLayer}
\end{Lemma}
Refer to the Appendix for proof of the above lemma. The three updating rules in Lemma \ref{BoundOnHiddenLayer} are of the order of $\sqrt{m}$ and $\sqrt{n}$ w.r.t. $m$ and $n$, respectively. However, as the width of the unfolded network is controlled by $m$, we consider the bounds on $\mathcal{Q}_{2,2,1}\left(f_{s}\right)$ and $\mathcal{Q}_\infty\left(f_{s}\right)$ w.r.t. $m$ in this work.

The following theorem gives the bound on $\|\mathbf{H}\|$ by deriving the bounds on
the quantities $\mathcal{Q}_{2,2,1}\left(f_{s}\right)$ and $\mathcal{Q}_\infty\left(f_{s}\right)$. The proof of Theorem \ref{Bound_on_Hessian_L_layer} basically uses the bounds on the weight matrices (Lemma \ref{BoundOnWeightInitialization} and Lemma \ref{BoundOnWeight}), bound on the hidden layer output (Lemma \ref{BoundOnHiddenLayer}), and properties of the activation function ($L_\sigma$-Lipschitz continuous and $\beta_\sigma$-smooth).
% Now using bounds on the weight matrices (Lemma \ref{BoundOnWeightInitialization} and Lemma \ref{BoundOnWeight}), bound on the hidden layer output (Lemma \ref{BoundOnHiddenLayer}), and properties of the activation function ($L_\sigma$-Lipschitz continuous and $\beta_\sigma$-smooth), we state the following theorem, which proposes the bound on $\|\mathbf{H}\|_2$ by deriving the bounds on the quantities $\mathcal{Q}_{2,2,1}\left(f_{s}\right)$ and $\mathcal{Q}_\infty\left(f_{s}\right)$.
% \begin{Lemma}
%     Consider an $L$-layer unfolded ISTA or ADMM network, $\mathbf{F}(\mathbf{W})$, with random Gaussian initialization $\mathbf{W}_0$. Then, with high probability over initialization the quantities $\mathcal{Q}_{2,2,1}\left(f_{s}\right) = O(1)$ and $\mathcal{Q}_{\infty}\left(f_{s}\right) = O\left(\frac{1}{\sqrt{m}}\right)$ w.r.t. $m$, at any point $\mathbf{W}\in B\left(\mathbf{W}_0, R\right)$ for some fixed $R>0$.
%     \label{BoundOn221}
% \end{Lemma}
\begin{theorem}
    Consider an $L$-layer unfolded ISTA or ADMM network, $\mathbf{F}(\mathbf{W})$, with random Gaussian initialization $\mathbf{W}_0$. Then, the quantities $\mathcal{Q}_{2,2,1}\left(f_{s}\right)$ and $\mathcal{Q}_{\infty}\left(f_{s}\right)$ satisfy the following equality w.r.t. $m$, over initialization, at any point $\mathbf{W}\in B\left(\mathbf{W}_0, R\right)$, for some fixed $R>0$:
\begin{equation}
\mathcal{Q}_{2,2,1}\left(f_{s}\right) = O(1) \text{ and }\mathcal{Q}_{\infty}\left(f_{s}\right) = \Tilde{O}\left(\frac{1}{\sqrt{m}}\right),
\end{equation}
with probabilities $1$ and $1-m e^{-c\ln ^{2}(m)}$ for some constant $c>0$, respectively.
This implies
    \begin{equation}
        \left\|{H}_{s}\right\| \leq  
        \sum_{l_{1}, l_{2}}\left\|{H}_{s}^{l_{1}, l_{2}}\right\| = \Tilde{O}\left(\frac{1}{\sqrt{m}}\right)
    \end{equation}
    and the Hessian spectral norm satisfies
    \begin{equation}
        \|\mathbf{H}\| = \Tilde{\Omega}\left(\frac{1}{\sqrt{m}}\right) = \Tilde{O}\left({\sqrt{m}}\right).
    \end{equation}
\label{Bound_on_Hessian_L_layer}
\end{theorem}
The proof of Theorem \ref{Bound_on_Hessian_L_layer} is 
motivated by \cite{Linearity} and is 
lengthy. Thus, the readers are directed to the supplementary material \cite{SupMaterial}, which provides the complete proof.
In summary, from both $1$-layer and $L$-layer analyses, we claim that the Hessian spectral norm bound of an unfolded network is proportional to the square root of the width of the network.
% From Theorem \ref{PL_Condition}, a lower $\|\mathbf{H}\|$ in $B(\mathbf{w}_0,R)$ implies a smaller change in the tangent kernal matrix in $B(\mathbf{w}_0,R)$. Then, there is a high chance that the PL$^*$ condition holds true in $B(\mathbf{w}_0,R)$.
% In contrast, a high $\|\mathbf{H}\|$ value leads to a significant change in the tangent kernel matrix. Then, the model may or may not satisfy PL$^*$ in $B(\mathbf{w}_0,R)$. 
% In the following section, we show that there exists a trade-off between the width of the network and the number of training samples to satisfy the PL$^*$ condition.
% Intuitively, an increase in the target vector dimension (for a fixed observation vector length) increases the complexity of the linear inverse problem at hand.
% In the following section, by using the bound on $\|\mathbf{H}\|$,  we provide a constraint on the width of the unfolded network such that it satisfies the PL$^*$ condition.
% We now use the bound on $\|\mathbf{H}\|$ to provide the constraint on the width of the unfolded network such that it satisfies the PL$^*$ condition.
% By using the above theorem, in the following section, we propose a constraint on the width of the unfolded network to satisfy the PL$^*$ condition.
% \subsection{For Unfolded ISTA Network}
% \subsection{For Unfolded ADMM Network}
% \section{LISTA and ADMM Satisfy PL$^*$ Condition}
% \section{Bound On the Number of Training Samples}

\subsection{Conditions on Unfolded Networks to Satisfy PL$^*$}
\label{Conditions_for_Unfolded_Networks_to_Satisfy_PL$^*$}
% \section{Unfolded Network Satisfying PL$^*$ Condition}
From Theorem \ref{PL_Condition}, the Hessian spectral norm of a model should hold the following condition to satisfy $\mu$-uniform conditioning in a ball $B(\mathbf{w}_0, R)$:
$\|\mathbf{H}_\mathcal{F}(\mathbf{w})\| \leq \frac{\lambda_0 - \mu}{2{L_\mathcal{F}}\sqrt{T}R},\ \forall \mathbf{w}\in B(\mathbf{w}_0,R).$
Since $\|\mathbf{H}_\mathcal{F}(\mathbf{w})\| = \underset{i\in[T]}{\text{max}}\|\mathbf{H}_{\mathcal{F}_i}(\mathbf{w})\|$, the above condition can be further simplified as
\begin{equation}
\|\mathbf{H}_{\mathcal{F}_i}(\mathbf{w})\| \leq \frac{\lambda_0 - \mu}{2{L_\mathcal{F}}\sqrt{T}R},\ \forall i\in[T] \text{ and }\mathbf{w}\in B(\mathbf{w}_0,R).
\label{Hessain_Bound}
\end{equation}
Substituting the Hessian spectral norm bound of LISTA and ADMM-CSNet, stated in Theorem  \ref{Bound_on_Hessian_L_layer}, in \eqref{Hessain_Bound} provides a constraint on the network width such that the square loss function satisfies the $\mu$-PL$^*$ condition in $B(\mathbf{w}_0, R)$: 
\begin{equation}
    m = \tilde{\Omega}\left(\frac{TR^2}{ (\lambda_{0}-\mu)^2}\right), \text{ where }\mu\in(0,\lambda_{0}).
    \label{Bound_on_m}
\end{equation}
Therefore, from Theorem \ref{PL_Convergence}, we claim that for a given fixed $T$ one should consider the width of the unfolded network as given in \eqref{Bound_on_m} to achieve near-zero training loss.
However, the $m$ (target vector dimension) value is generally fixed for a given linear inverse problem.
% Hence, one needs a constraint on the number of training samples instead of network width.
Hence, we provide the constraint on $T$ instead of $m$.
Substituting the $\|\mathbf{H}_{\mathcal{F}_i}(\mathbf{w})\|$ bound in \eqref{Hessain_Bound} also provides a threshold on $T$, which is summarized in the following theorem:
\begin{theorem}
    Consider a finite $L$-layer unfolded network as given in \eqref{LISTA_Model} or \eqref{ADMM_CSNet_Model} with $m$ as the network width. Assume that the model is well-conditioned at initialization, i.e., $\lambda_\text{min}(K_{\text{Unfolded}}(\mathbf{w}_0))=\lambda_{0,\text{Unfolded}}$, for some $\lambda_{0, \text{Unfolded}}>0$. Then, the loss landscape corresponding to the square loss function satisfies the $\mu$-PL$^*$ condition in a ball $B(\mathbf{w}_0, R)$, if the number of training samples, $T_{\text{Unfolded}}$, satisfies the following condition:
    \begin{equation}
    T_{\text{Unfolded}} = \tilde{O}\left(\frac{m(\lambda_{0,\text{Unfolded}}-\mu)^2}{R^2}\right), \ \mu\in(0,\lambda_{0,\text{Unfolded}}).
    \label{Bound_on_T}
\end{equation}
\end{theorem}
Thus, while addressing a linear inverse problem using unfolded networks, one should consider the number of training samples as given in \eqref{Bound_on_T}, to obtain zero training loss as the number of GD epochs increases to infinity. 
Observe that the threshold on $T$ increases with the increase in the network width. 
We attribute this to the fact that a high network width is associated with more trainable parameters in the network, which provides the ability to handle/memorize more training samples. 
Conversely, a smaller network width leads to fewer trainable parameters, thereby impacting the network's performance in handling training samples.
% In general, training using more training samples gives a good generalization performance. Hence, a higher threshold leads to a better generalization. 
% Moreover, getting a higher threshold will lead to handling a higher number of training samples, which gives a better generalization.
% Hence, an unfolded network trained using $T_1$ number of training samples gives a better generalization than an FFNN trained using $T_2$ samples, where $T_1>T_2$.
% In contrast, a smaller network width has fewer trainable parameters, which reduces performance in terms of handling the training samples of the network. 

\begin{figure}
\centering
\includegraphics[width=9.5cm, height=6cm]{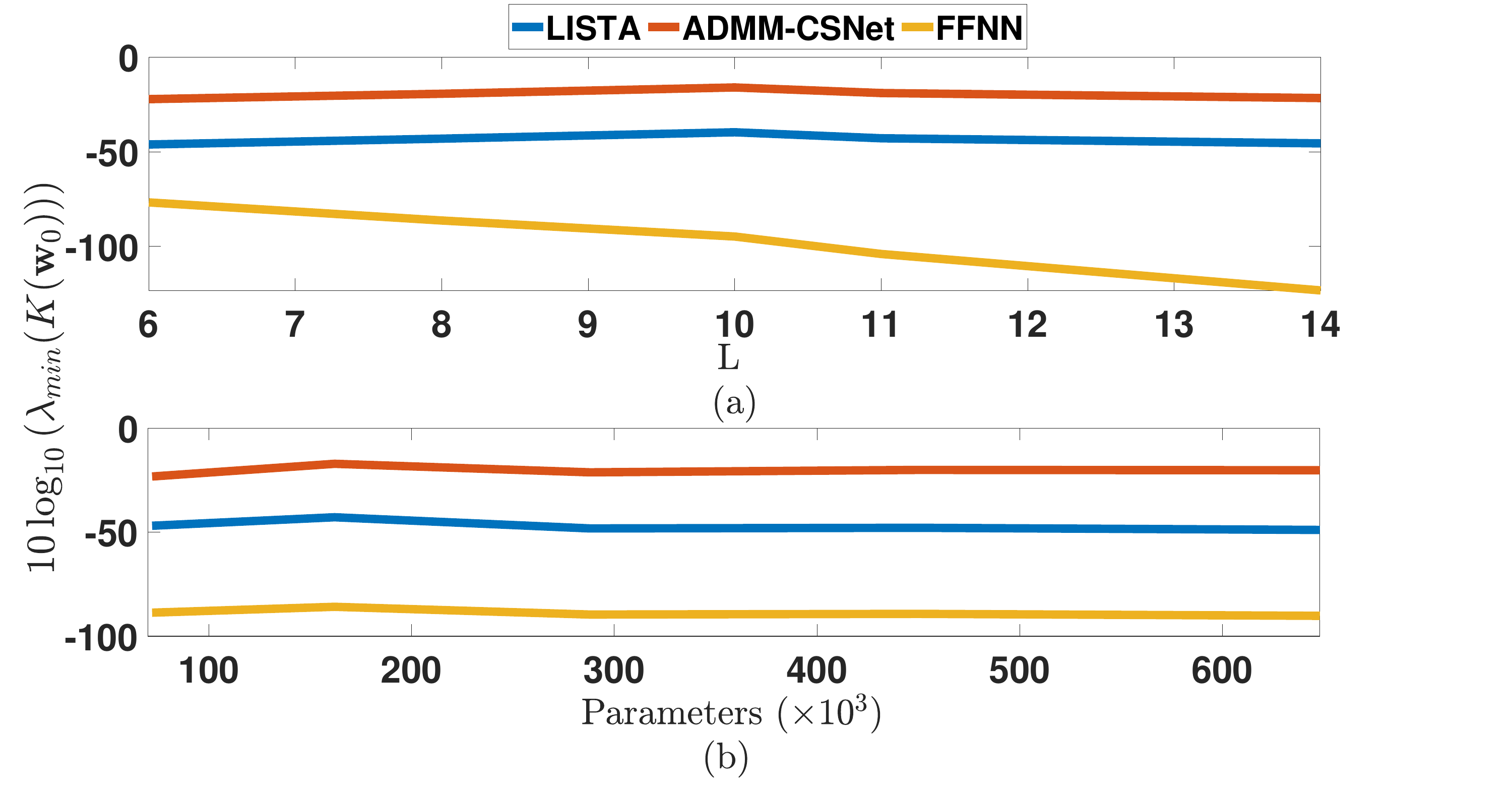}
         \caption{Variation of the minimum eigenvalue of tangent kernel matrix at initialization: (a) With respect to the number of layers. (b) With respect to the network learnable parameters.}
    \label{Minimum_Eigenvalue_NTK}
\end{figure}

\textbf{Comparison with FFNN:} In \cite{LossLandscape}, the authors computed the Hessian spectral norm of an FFNN with a scalar output, which is of the order of $\tilde{O}\left(\frac{1}{\sqrt{m}}\right)$. Following the analysis procedure of an $m$-output model given in Section \ref{Hessian_Spectral_Norm}, one can obtain the Hessian spectral norm of an FFNN with $m$-output and smoothed soft-thresholding non-linearity as given below:
\begin{equation}
    \|\mathbf{H}\| = \Tilde{\Omega}\left(\frac{1}{\sqrt{m}}\right) = \Tilde{O}\left({\sqrt{m}}\right).
\end{equation}
This implies that the bound on the number of training samples, $T_{\text{FFNN}}$, for an $m$-output FFNN to satisfy the $\mu$-PL$^*$ is
    \begin{equation}
    T_{\text{FFNN}} = \tilde{O}\left(\frac{m(\lambda_{0,\text{FFNN}}-\mu)^2}{R^2}\right), \ \mu\in(0,\lambda_{0,\text{FFNN}})
    \label{Bound_on_T_FFNN}
\end{equation}
Note that $m$ is a fixed value in both \eqref{Bound_on_T} and \eqref{Bound_on_T_FFNN}, $R$ is of the order of $O\left(\frac{1}{\mu}\right)$ (refer to Theorem \ref{PL_Convergence}), and $\mu$ depends on $\lambda_0=\lambda_{\text{min}}\left(K\left(\mathbf{w}_0\right)\right)$. 
Therefore, from \eqref{Bound_on_T} and \eqref{Bound_on_T_FFNN}, the parameter that governs the number of training samples of a network is the minimum eigenvalue of the tangent kernel matrix at initialization.
Hence, we compare both $T_\text{Unfolded}$ and $T_\text{FFNN}$ by deriving the upper bounds on $\lambda_{0,\text{Unfolded}}$ and $\lambda_{0,\text{FFNN}}$.
% Specifically, by using the following theorem, we provide the mathematical intuition to justify $T_\text{Unfolded}>T_\text{FFNN}$.
Specifically, in the following theorem, we show that the upper bound of $\lambda_{0,\text{Unfolded}}$ is higher compared to $\lambda_{0,\text{FFNN}}$.
\begin{theorem}
    Consider an L-layered FFNN, defined as
    \begin{equation}
    \label{L-layered_ffnn}
    \begin{aligned}
    & \mathbf{f}_{\text{FFNN}}=\frac{1}{\sqrt{m}} \mathbf{x}^L,
    % \\ &\text{where,} \\
    \mathbf{x}^{l} =  \sigma\left(\frac{W^{l}}{\sqrt{m}} \mathbf{x}^{l-1}\right)\ \in \mathbb{R}^m,\ l \in [L],
    \end{aligned}
    \end{equation}
    with $\mathbf{x}^0 = \sqrt{\frac{m}{n}}\mathbf{y} \in \mathbb{R}^n,\ W^1 \in \mathbb{R}^{m \times n}, \text{ and } W^l \in \mathbb{R}^{m \times m} \ \  \forall l \in [L]-\{1\} $. Also, consider the unfolded network defined in \eqref{LISTA_Model} or \eqref{ADMM_CSNet_Model}. 
    Then, the upper bound on the minimum eigenvalue of the tangent kernel matrix at initialization for unfolded network, $\text{UB}_{\text{Unfolded}}$ (either $\text{UB}_{\text{LISTA}}$ or $\text{UB}_{\text{ADMM-CSNet}}$), is greater than that of FFNN, $\text{UB}_{\text{FFNN}}$, i.e., $\text{UB}_{\text{Unfolded}}>\text{UB}_{\text{FFNN}}$.
    \label{EigenValue_Bounds}
\end{theorem}
Proof of the above theorem is given in the Appendix.
To better understand Theorem \ref{EigenValue_Bounds}, substitute $L=2$ in equations \eqref{Minimum_eigenvalue_NTK_FFNN},  \eqref{Minimum_eigenvalue_NTK_Unfolded}, and \eqref{minimum_eig_value_NTK_admm}, this leads to
$$\text{UB}_{\text{FFNN}} = \hat{L}^4 \hat{y} \left[\|W_0^1\|^2+\|\mathbf{v}_s^{T} W_0^2\|^2\right],$$ 
\begin{equation*}
\begin{aligned}
    \text{UB}&_{\text{LISTA}} = 
    \hat{L}^4 \hat{y} \left[\|W_{10}^1\|^2+\|\mathbf{v}_s^{T}W_{20}^2\|^2\right] + \hat{L}^2\hat{y}+\\ &\hat{L}^4 \hat{x} \left[\|W_{20}^1\|^2+\|\mathbf{v}_s^{T}W_{20}^2\|^2\right]
    +2\hat{L}^4 \sqrt{\hat{x}\hat{y}} \|W_{10}^1\|\|W_{20}^1\|,
\end{aligned}
\end{equation*}
and 
\begin{equation*}
\footnotesize
\begin{aligned}
    \text{UB}&_{\text{ADMM-CSNet}} = 
    \hat{L}^4 \hat{y} \left[\|W_{10}^1\|^2+\|\mathbf{v}_s^{T}W_{20}^2\|^2\right] +\hat{L}^2\hat{y}+ \frac{\|\mathbf{u}^{(1)}\|^2}{m}+\\
    &\hspace{-0.5cm}\hat{L}^4 \hat{a}^{(0)} \left[\|W_{20}^1\|^2+ \|\mathbf{v}_s^{T}W_{20}^2\|^2\right]+ 2 \hat{L}\|\tilde{\mathbf{z}}^{(l)}\| \|\mathbf{u}^{(1)}\| +\hat{L}^4 \|\mathbf{u}^{(0)}\|^2+\hat{L}^4\\
    &\hspace{-0.5cm}\left[ 2 \sqrt{\hat{y} \hat{a}^{(0)}}\|W_{10}^1\|\|W_{20}^1\|+2 \sqrt{\hat{a}^{(0)}} \|W_{20}^1\| \|\mathbf{u}^{(0)}\| +2 \sqrt{\hat{y}} \|W_{10}^1\| \|\mathbf{u}^{(0)}\|  \right].
\end{aligned}
\end{equation*}
Since the dimension of $W_1^1$ ($W_2^2$) of unfolded is same as $W^1$ ($W^2$) of FFNN, we conclude that $\text{UB}_{\text{Unfolded}}>\text{UB}_{\text{FFNN}}$ for $L=2$.  
% Note that the weight matrix $W_1^1$ ($W_2^2$) of unfolded is having the same dimension as $W^1$ ($W^2$) of FFNN and these matrices are initialized with $\mathcal{N}(0,1)$.
% Hence, we conclude that the upper bound on $\lambda_{0,\text{Unfolded}}$ is greater than the upper bound on $\lambda_{0,\text{Unfolded}}$ for $L=2$. 
One can verify that this relation holds for any $L$ value using the generalized expressions given in \eqref{Minimum_eigenvalue_NTK_FFNN}, \eqref{Minimum_eigenvalue_NTK_Unfolded}, and \eqref{minimum_eig_value_NTK_admm}.
Figures \ref{Minimum_Eigenvalue_NTK} (a) and \ref{Minimum_Eigenvalue_NTK} (b)  depict the variation of $10\log_{10}\left(\lambda_{\text{min}}\left({K}(\textbf{w}_0)\right)\right)$ w.r.t. $L$ (here we considered $T=10$, $m=100$, $n=20$, and $k=2$) and $P$ (here we vary $m$, $n$, and $k$ values by fixing $T=10$, $L=6$ for unfolded, and $L=8$ for FFNN), respectively, for LISTA, ADMM-CSNet, and FFNN.
From these figures, we see that $\lambda_{0,\text{Unfolded}}>\lambda_{0,\text{FFNN}}$.
Consequently, from Theorem \ref{EigenValue_Bounds}, \eqref{Bound_on_T}, and \eqref{Bound_on_T_FFNN}, we also claim that the upper bound of $T_{\text{Unfolded}}$ is high compared to $T_{\text{FFNN}}$.
As a result, $T_{\text{Unfolded}}>T_{\text{FFNN}}$ whenever $\lambda_{0,\text{Unfolded}}>\lambda_{0,\text{FFNN}}$.
% Furthermore, from the aforementioned equations we $\text{UB}_{\text{ADMM-CSNet}}>\text{UB}_{\text{LISTA}}$. Thus, one can expect $\lambda_{0,\text{ADMM-CSNet}}>\lambda_{0,\text{LISTA}}$, the same is justified from figures \ref{Minimum_Eigenvalue_NTK} (a) and \ref{Minimum_Eigenvalue_NTK} (b).
Moreover, from the aforementioned equations, it is evident that $\text{UB}_{\text{ADMM-CSNet}}$ exceeds $\text{UB}_{\text{LISTA}}$. Consequently, it is reasonable to anticipate that $\lambda_{0,\text{ADMM-CSNet}}$ will surpass $\lambda_{0,\text{LISTA}}$. This inference is substantiated by the data depicted in figures \ref{Minimum_Eigenvalue_NTK} (a) and \ref{Minimum_Eigenvalue_NTK} (b).
This implies that the upper bound on $T_{\text{ADMM-CSNet}}$ exceeds the upper bound on $T_{\text{LISTA}}$.
Through simulations, we show that $T_{\text{ADMM-CSNet}}>T_{\text{LISTA}}>T_{\text{FFNN}}$ in the following section.
Since the threshold on $T$ — guaranteeing memorization — is higher for unfolded networks than FFNN, we should obtain a better expected error, which is upper bounded by the sum of generalization and training error \cite{Expected_error}, for unfolded networks than FFNN for a given $T$ value such that $T_{\text{FFNN}}< T\leq T_{\text{Unfolded}}$. Because in such scenarios the training error is zero and the generalization error is smaller  for unfolded networks \cite{Avner}.

% \textcolor{red}{Indeed, training with a larger set of diverse training samples tends to yield better generalization performance. As a result, a higher threshold on the number of training samples can enhance the network's ability to generalize\footnote{Refer to Fig. 2 in \cite{Avner}, where the authors showed that the estimation (test) error of unfolded is better than FFNN w.r.t. number of training samples; further, this error reduces with the increase in the number of training samples.} well to unseen data.}

\section{Numerical Experiments}
\label{experiment}
We perform the following simulations to support the proposed theory. 
% All simulations are executed using Python version 3.9 on multiple AMD Ryzen 7 3700x 8-core processors with 16 GB RAM.
For all the simulations in this section, we fix the following for LISTA, ADMM-CSNet, and FFNN: $1.$ Parameters are initialized independently and identically (i.i.d.) from a Gaussian distribution with zero mean and unit variance, i.e., $\mathcal{N}(0,1)$.
$2.$ Networks are trained with the aim of minimizing the square loss function \eqref{Loss_function} using stochastic GD.
Note that the theoretical analysis proposed in this work is for GD, however, to address the computation and storage issues, we considered stochastic GD for the numerical analysis. 
$3.$ Modified soft-plus activation function (refer to \ref{Assumptions}) with $\lambda=1$ is used as the non-linear activation function. $4.$ A batch size of $\frac{T}{5}$ is considered. $5.$ All the simulations are repeated for $10$ trials.

\begin{figure}
         \centering
         \includegraphics[width=9cm, height=5cm]{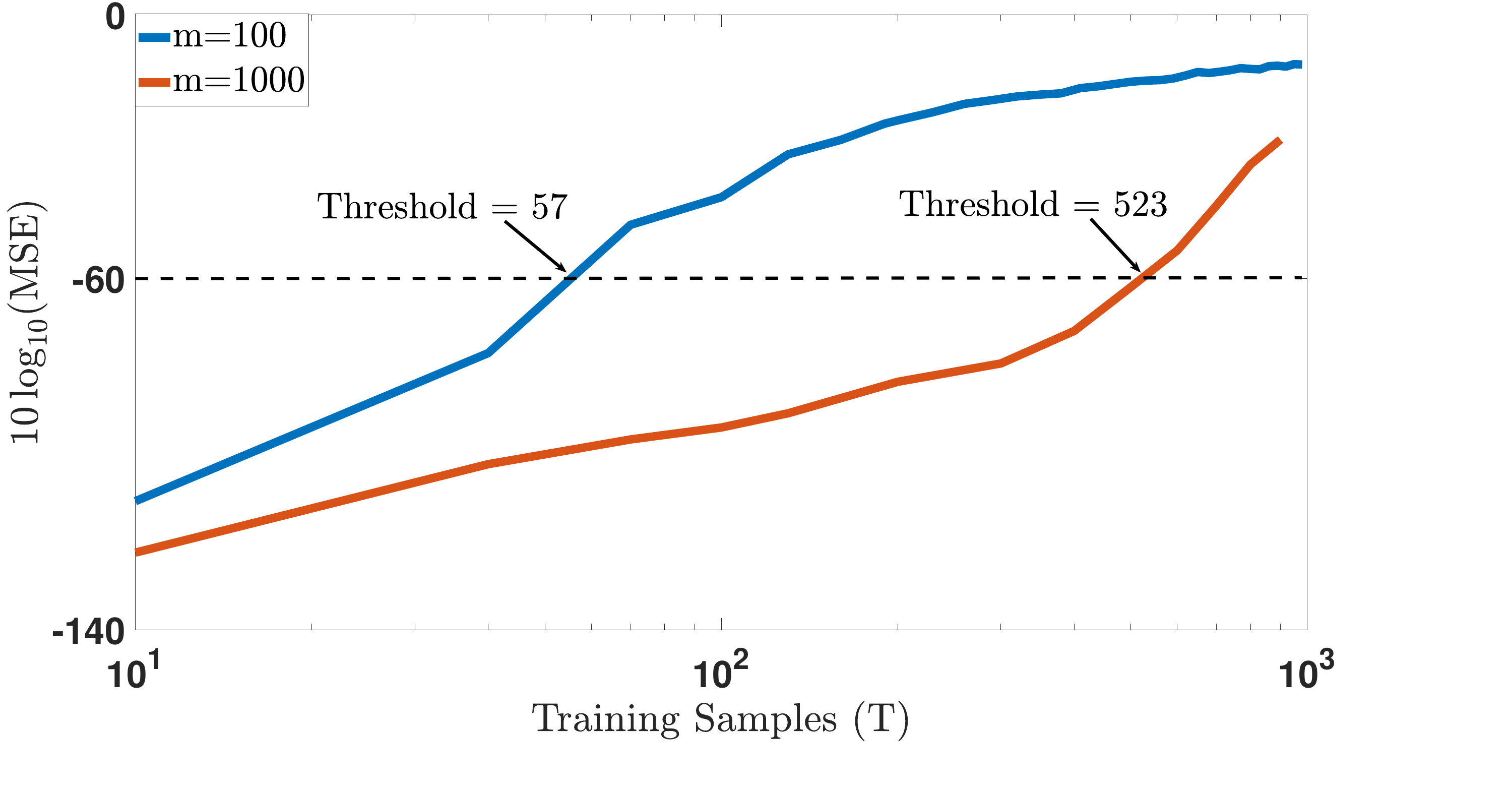}
         \caption{Training loss vs $T$ for LISTA.}
         \label{Threshold_for_LISTA}
\end{figure}

\begin{figure}
         \centering
         \includegraphics[width=9cm, height=5cm]{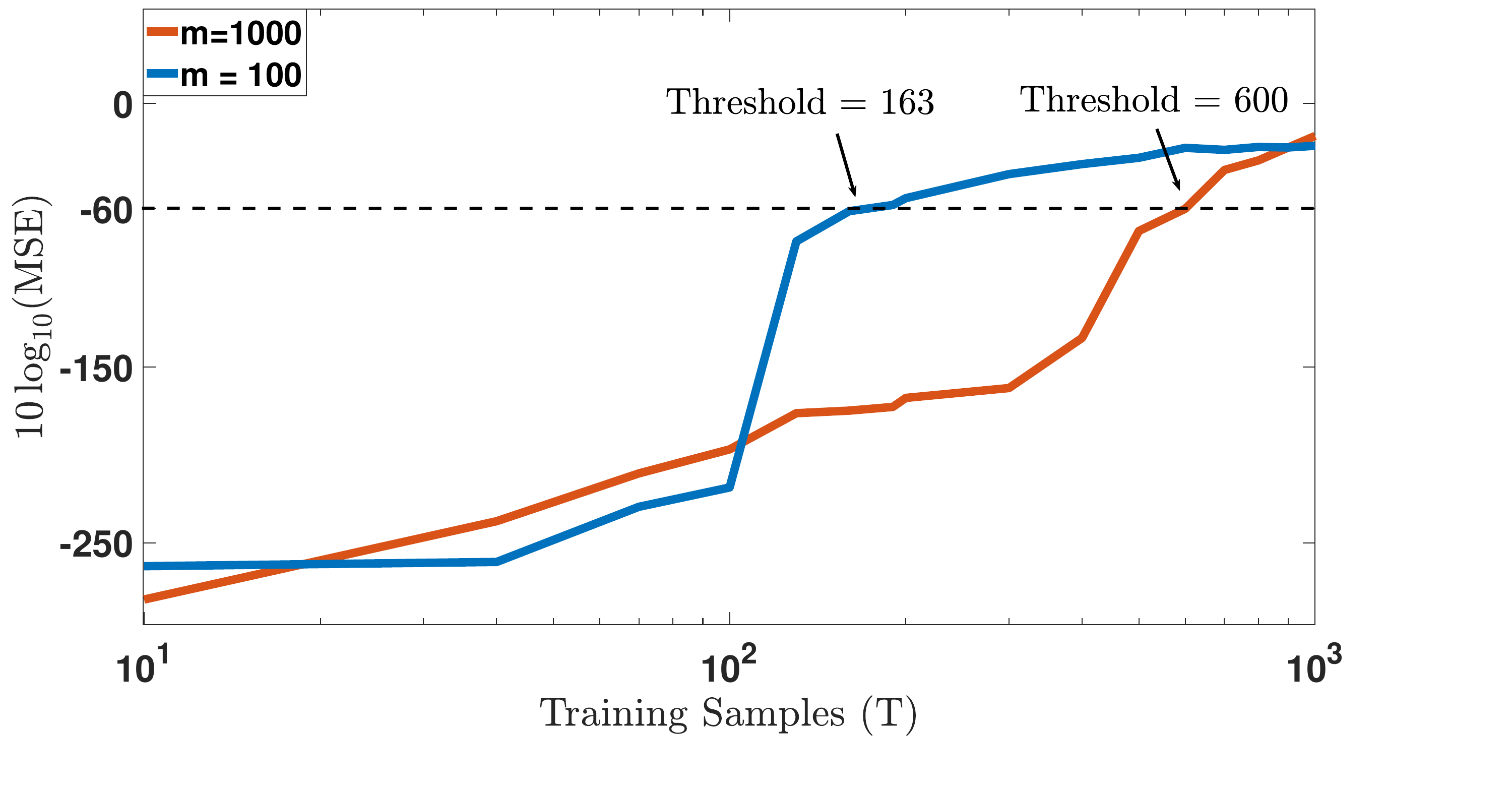}
         \caption{Training loss vs $T$ for ADMM-CSNet.}
         \label{Threshold_for_ADMM}
\end{figure}

\textbf{Threshold on $T$:}
From \eqref{Bound_on_T}, the choice of $T$ plays a vital role in achieving near-zero training loss. 
To illustrate this, consider two linear inverse models: $\mathbf{y}_1 = {A}_1\mathbf{x}_1 + \mathbf{e}_1$ and $\mathbf{y}_2 = {A}_2\mathbf{x}_2 + \mathbf{e}_2$,
% We now demonstrate this with the following setup: Consider two linear inverse models $\mathbf{y}_1 = {A}_1\mathbf{x}_1 + \mathbf{e}$ and $\mathbf{y}_2 = {A}_2\mathbf{x}_2 + \mathbf{e}$,
where $\mathbf{y}_1\in \mathbb{R}^{20\times 1}$, $\mathbf{x}_1\in \mathbb{R}^{100\times 1}$, $A_1\in \mathbb{R}^{20\times 100}$, $\|\mathbf{x}_1\|_0 = 2$, $\mathbf{y}_2\in \mathbb{R}^{200\times 1}$, $\mathbf{x}_2\in \mathbb{R}^{1000\times 1}$, $A_2\in \mathbb{R}^{200\times 1000}$, and $\|\mathbf{x}_2\|_0 = 10$.
Generate synthetic data using a random linear operator matrix, which follows the uniform distribution, and then normalize it to ensure $\|A_1\|_F = \|A_2\|_F = 10$. 
Both models are subjected to Gaussian noise $(\mathbf{e}_1$ and $\mathbf{e}_2$) with a signal-to-noise ratio (SNR) of $10$ dB.
Construct an $L$-layer LISTA and ADMM-CSNet with $L=11$.  Here, we train LISTA for $30$K epochs and ADMM-CSNet for $40$K epochs.
For the first model, we choose $0.12$ and $0.09$ as learning rates for LISTA and ADMM-CSNet, respectively.
For the second model, we choose $1.2$ for LISTA and $0.9$ for ADMM-CSNet.
Figures \ref{Threshold_for_LISTA} and \ref{Threshold_for_ADMM} depict the variation of mean square loss/error (MSE) w.r.t. $T$ for both LISTA and ADMM-CSNet, respectively.
Note that for a fixed $m$ there exists a threshold (by considering a specific MSE value) on $T$ such that choosing a $T$ value that is less than this threshold leads to near-zero training loss. Moreover, observe that this threshold increases as the network width grows.

\begin{figure}
    \centering
    \includegraphics[width=9cm, height=5cm]{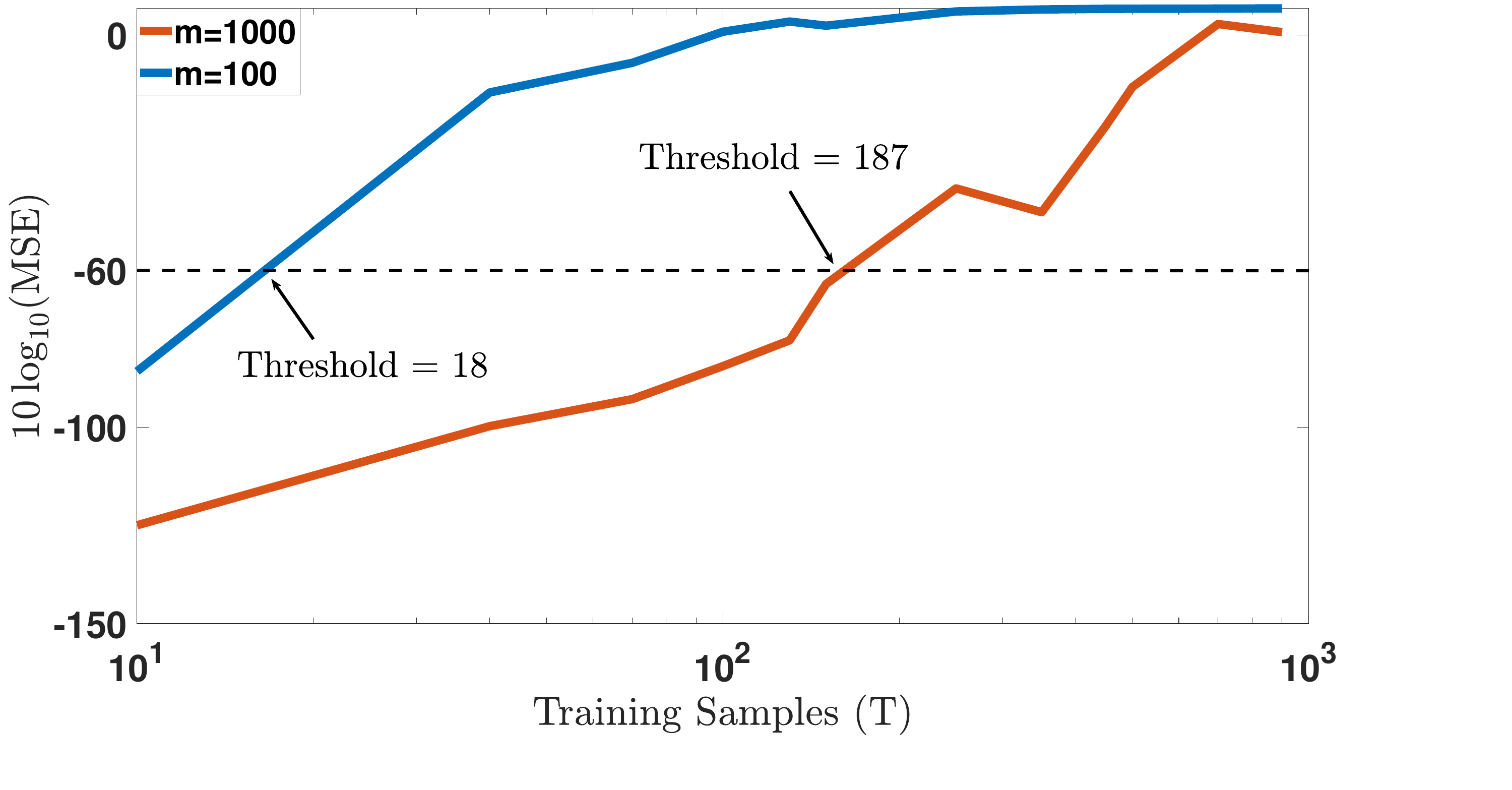}
    \caption{Training loss vs $T$ for FFNN.}
    \label{Threshold_for_Ffnn}
\end{figure}

For comparison, construct an $L$-layer FFNN, to recover $\mathbf{x}_1$ and $\mathbf{x}_2$, that has the same number of parameters as that of unfolded, hence, we choose $L=14$. Here, we train the network for $40K$ epochs with a learning rate of $0.04$ for the first model and $0.3$ for the second model.
Fig. \ref{Threshold_for_Ffnn} shows the variation of MSE w.r.t. $T$. From Fig. \ref{Threshold_for_Ffnn}, we can conclude that the threshold for FFNN is lower compared to LISTA and ADMM-CSNet.

% have learning rates of respectively, while for the second inverse problem setup, the learning rates are 1.2 for LISTA and 0.9 for ADMM-CSNet.
% Here we generate the synthetic data by considering a random measurement matrix that follows the uniform distribution and we normalize it such that $\|A_1\|_F = \|A_2\|_F = 10$. 
% Both models are added with Gaussian noise $\mathbf{e}$ with an SNR of $10$ dB.
% % Assume that both models are added with noise $e$ that follows Gaussian distribution with SNR $10 $ dB.
% We construct an $L$-layer LISTA and ADMM-CSNet with $L=11$, initializing their parameters from a Gaussian distribution with zero mean and unit variance. 
% We train the networks using a stochastic gradient descent (GD) with square loss function \eqref{Loss_function}, by considering 30K and 20K number of epochs for LISTA and ADMM-CSNet respectively. 
%Additionally, we consider learning rates of 0.12 and 0.09 in the first inverse problem setup for LISTA and ADMM-CSNet respectively. Similarly learning rates of 1.2 and 0.9 in the second inverse problem setup for LISTA and ADMM-CSNet respectively. We consider the modified soft plus (from the section \ref{Assumptions}) as an activation function with $\lambda=1$ for both networks.
%For optimization, we employ a stochastic gradient descent (GD) algorithm, minimizing the square loss function defined  \eqref{Loss_function}.

\begin{figure}
        \centering
         \includegraphics[width=9cm, height=5cm]{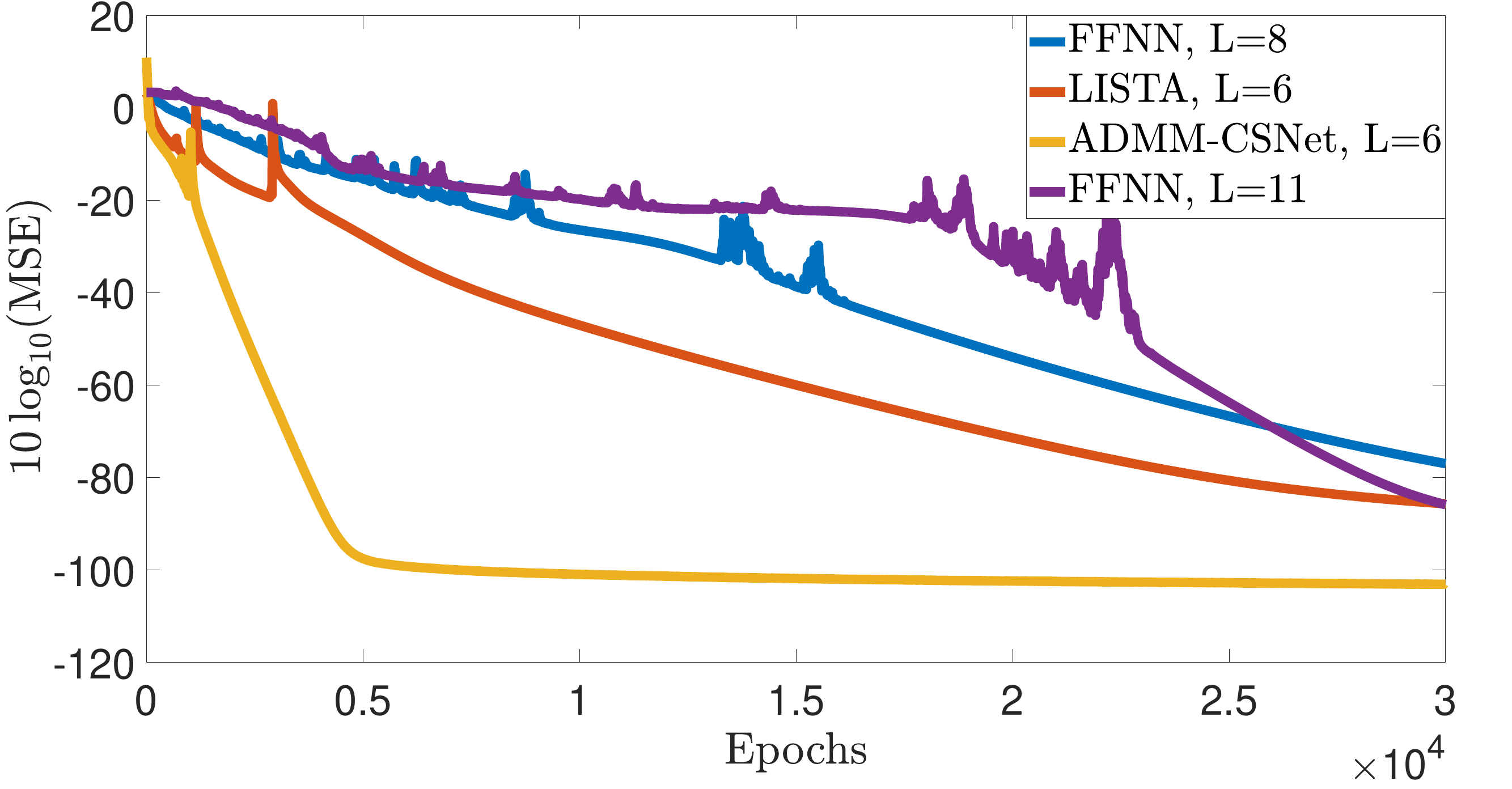}
     \caption{\small  Comparison between LISTA, ADMM-CSNet, and FFNN in terms of the required number of parameters, $P$, for training loss convergence.}
     \label{compare}
\end{figure}

\textbf{Comparison Between Unfolded and Standard Networks:} We compare LISTA and ADMM-CSNet with FFNN in terms of parameter efficiency. 
To demonstrate this, consider the first linear inverse model given in the above simulation. Then, construct LISTA, ADMM-CSNet, and FFNN with a fixed number of parameters and consider $T = 30$.
Also, consider the same learning rates that are associated with the first model in the above simulation for LISTA, ADMM-CSNet, and FFNN.
Here we choose $L = 6$ for both LISTA and ADMM-CSNet, and $L=8$ for FFNN,  resulting in a total of $72K$ parameters. 
As shown in Fig. \ref{compare}, the convergence of training loss to zero is better for LISTA and ADMM-CSNet compared to FFNN. 
Fig. \ref{compare} also shows the training loss convergence of FFNN with $L=11$. Now, FFNN has $102K$ learnable parameters, and its performance is comparable to LISTA for higher epoch values.
Therefore, to achieve a better training loss FFNN requires more trainable parameters.
% Observe that the loss convergence rate of ADMM-CSNet/LISTA is better compared to FFNN. We now give the theoretical intuition behind this. 
% Figure \ref{compare_and_generalize} (b) also depicts the training loss convergence of FFNN with $L=11$, in this case, the number of learnable parameters is $102K$ and the performance is comparable with LISTA.
% Therefore, to achieve a better training loss convergence we require more trainable parameters for FFNNs than LISTA/ADMM-CSNet.

\begin{figure}
    \centering
    \includegraphics[width=9.5cm, height=5cm]{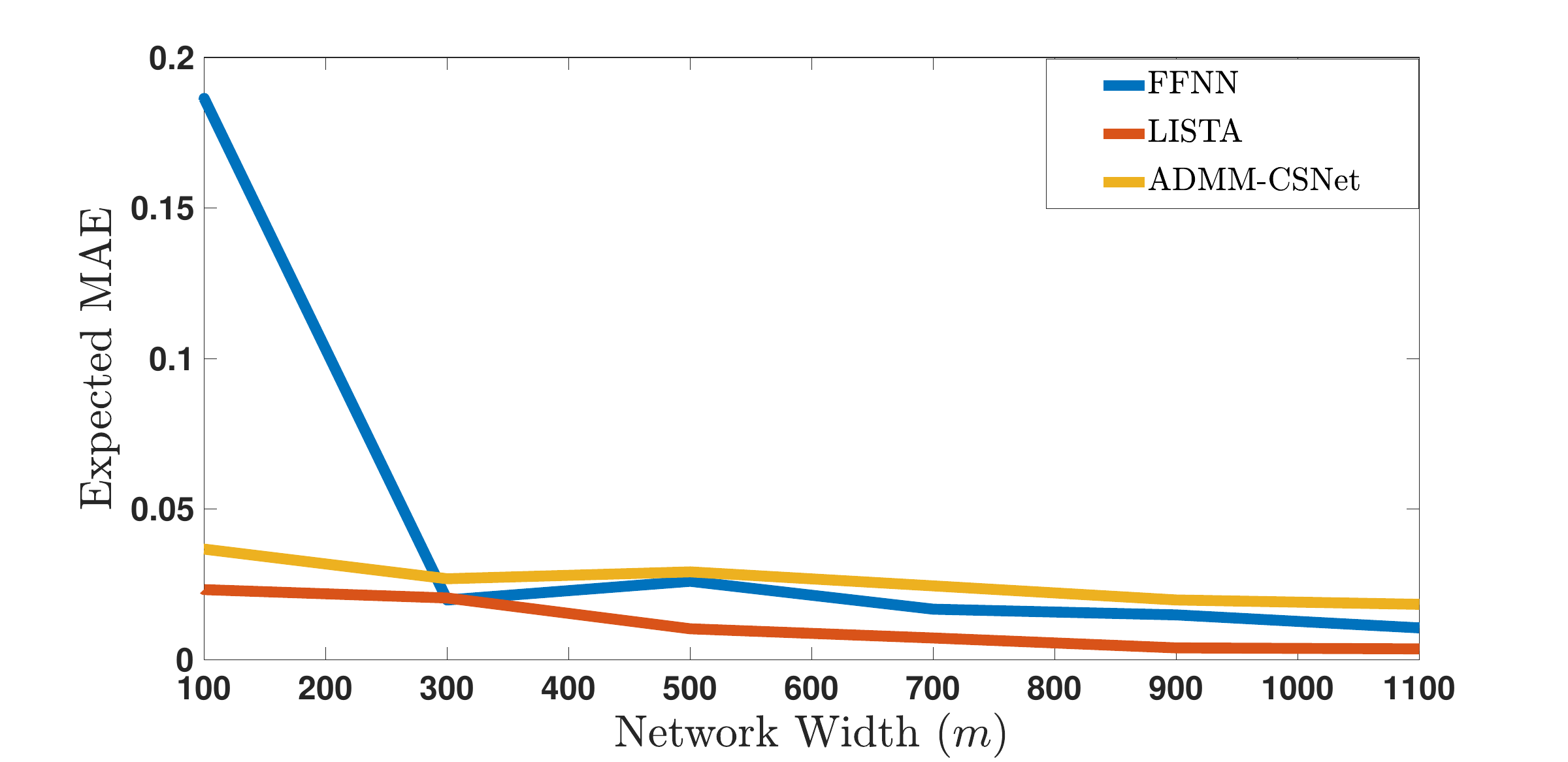}
    \caption{\small Variation of the expected MAE w.r.t. $m$ for both LISTA and ADMM-CSNet.}
    \label{generalize}
\end{figure}

\textbf{Generalization:}
In this simulation, we show that zero-training error leads to better generalization.
To demonstrate this, consider LISTA/ADMM-CSNet/FFNN with a fixed $T$ and observe the variation of the expected mean absolute error (MAE) w.r.t. $m$. 
If the generalization performance is better, then it is anticipated that the expected MAE reduces as the $m$ increases. Because an increase in $m$ improves the possibility of getting near-zero training loss for a fixed $T$.
In Fig. \ref{generalize}, we present the results for LISTA, ADMM-CSNet, and FFNN with $T=100$. Notably, the expected MAE diminishes as $m$ increases, i.e., as the number of parameters grows. 
Further, it is observed that for this choice of $T$, the training error is near-zero for $m$ values exceeding approximately $300$ for FFNN, and approximately $250$ for both LISTA and ADMM-CSNet.
This finding underscores the importance of zero-training error in generalization.

However, it is important to note that the generalization results presented here are preliminary and require a rigorous analysis for more robust conclusions. Because considering a smaller value of $T$ may not yield satisfactory generalization performance. Thus, it is important to find a lower bound on $T$ to optimize both the training process and overall generalization capability, which we consider as a future work of interest.
% if the threshold on $T$ is very small, then it will not give a good generalization.
% So it is important to give a lower bound on the no. of training samples to achieve a good generalization.
% It is expected that for lower $m$ values, where the training error is not equal to zero, we get a high test error.
% On the other hand, for higher values of $m$, the test error is significantly reduced, as these networks achieve near-zero training loss.
% In Figure \ref{compare_and_generalize} (a), we present the results for both LISTA and ADMM-CSNet with $T=100$. Notably, the test MAE diminishes as $m$ increases, i.e., as the number of parameters grows. This finding underscores the importance of zero-training error in generalization.
% In contrast, for higher $m$ values, the test error is very low, as these networks provide a near-zero training loss. Figure \ref{compare_and_generalize} (a) depicts the results for both LISTA and ADMM-CSNet by considering $T=100$. Notice that the test error reduces as we increase the width of the network, i.e., increase in the number of parameters.

\section{Conclusion}
In this work, we provided optimization guarantees for finite-layer LISTA and ADMM-CSNet with smooth nonlinear activation.
% that have been used to solve the linear inverse problem.
% In the process, we first derived the Hessian spectral norm of these unfolded networks. 
We begin by deriving the Hessian spectral norm of these unfolded networks.
Based on this, we provided conditions on both the network width and the number of training samples, such that the empirical training loss converges to zero as the number of learning epochs increases using the GD approach.
%We also provided a threshold on the number of training samples to get the zero training error for both unfolded networks.
Additionally, we showed that LISTA and ADMM-CSNet outperform the standard FFNN in terms of threshold on the number of training samples and parameter efficiency.
% we provided a comparative analysis of the threshold on the number of training samples among LISTA, ADMM-CSNet, and FFNN. In particular, we proved that LISTA and ADMM-CSNet have higher threshold values on the number of training samples in comparison with FFNN.
We provided simulations to support the theoretical findings.
% proposed in this work. 
% Finally, we demonstrated that unfolded networks require less number of parameters to achieve near-zero training error than the standard FFNNs. 
%A number of numerical results are provided for backing up the theoretical findings proposed in this work.

The work presented in this paper is an initial step to understand the theory behind the performance of unfolded networks.
While considering certain assumptions, our work raises intriguing questions for future research.
% , several intriguing questions for future research arise.
% In the process, we considered several assumptions, which raise the following interesting questions for future research:
For instance, we approximated the soft-threshold activation function with a double-differentiable function formulated using soft-plus. 
However, it is important to analyze the optimization guarantees without relying on any such approximations. 
Additionally, we assumed a constant value for $\lambda$ in $\sigma_\lambda(\cdot)$. It is interesting to explore the impact of treating $\lambda$ as a learnable parameter.
% Furthermore, the theoretical findings proposed in this work are obtained by considering the loss function as a square loss function. Thus, 
Furthermore, analyzing the changes in the analysis for other loss functions presents an intriguing avenue for further research.

\appendix
\textit{Proof of \textbf{Theorem} \ref{Bound_Hessian_L_Layer}:}
The Hessian block ${H}_{s}^{l_{1}, l_{2}}$ can be decomposed as given in \eqref{HessianDecomp}, using the following chain rule:
\footnotesize
\begin{equation*}
    \frac{\partial f_{s}}{\partial \mathbf{w}^{l}}=\frac{\partial \mathbf{g}^{l}}{\partial \mathbf{w}^{l}}\left(\prod_{l^{\prime}=l+1}^{L} \frac{\partial \mathbf{g}^{l}}{\partial \mathbf{g}^{l^{\prime}-1}}\right) \frac{\partial f_{s}}{\partial \mathbf{g}^{L}}.
\end{equation*}
\normalsize
\begin{equation}
\footnotesize
\begin{aligned}
{H}_{s}^{l_{1}, l_{2}}=& \frac{\partial^{2} \mathbf{g}^{l_{1}}}{\left(\partial \mathbf{w}^{l_{1}}\right)^{2}} \frac{\partial f_{s}}{\partial \mathbf{g}^{l_{1}}} \mathbb{I}_{l_{1}=l_{2}}+\left(\frac{\partial \mathbf{g}^{l_{1}}}{\partial \mathbf{w}^{l_{1}}} \prod_{l^{\prime}=l_{1}+1}^{l_{2}-1} \frac{\partial \mathbf{g}^{l^{\prime}}}{\partial \mathbf{g}^{l^{\prime}-1}}\right) \frac{\partial^{2} \mathbf{g}^{l_{2}}}{\partial \mathbf{w}^{l_{2}} \partial \mathbf{g}^{l_{2}-1}}\\
& \left(\frac{\partial f_{s}}{\partial \mathbf{g}^{l_{2}}}\right) +\sum_{l=l_{2}+1}^{L}\left(\frac{\partial \mathbf{g}^{l_{1}}}{\partial \mathbf{w}^{l_{1}}} \prod_{l^{\prime}=l_{1}+1}^{l-1} \frac{\partial \mathbf{g}^{l^{\prime}}}{\partial \mathbf{g}^{l^{\prime}-1}}\right) \frac{\partial^{2} \mathbf{g}^{l^{\prime}}}{\left(\partial \mathbf{g}^{l^{\prime}-1}\right)^{2}}\\
&\left(\frac{\partial \mathbf{g}^{l_{2}}}{\partial \mathbf{w}^{l_{2}}} \prod_{l^{\prime}=l_{2}+1}^{l} \frac{\partial \mathbf{g}^{l^{\prime}}}{\partial \mathbf{g}^{l^{\prime}-1}}\right)\left(\frac{\partial f_{s}}{\partial \mathbf{g}^{l}}\right).
\end{aligned}
\label{HessianDecomp}
\end{equation}
From \eqref{HessianDecomp}, the spectral norm of ${H}_{s}^{l_{1}, l_{2}}$ can be bounded as
% written in \eqref{HessianBlockBound}.
\begin{equation}
\footnotesize
\begin{aligned}
\left\|{H}_{s}^{l_{1}, l_{2}}\right\|_2
% & \leq\left\|\frac{\partial^{2} \mathbf{g}^{l_{1}}}{\left(\partial \mathbf{w}^{l_{1}}\right)^{2}}\right\|_{2,2,1}\left\|\frac{\partial f_{s}}{\partial \mathbf{g}^{l_{1}}}\right\|_{\infty}+\left\|\frac{\partial \mathbf{g}^{l_{1}}}{\partial \mathbf{w}^{l_{1}}}\right\|_F \prod_{l^{\prime}=l_{1}+1}^{l_{2}-1}\left\|\frac{\partial \mathbf{g}^{l^{\prime}}}{\partial \mathbf{g}^{l^{\prime}-1}}\right\|_F\left\|\frac{\partial^{2} \mathbf{g}^{l_{2}}}{\partial \mathbf{w}^{l_{2}} \partial \mathbf{g}^{l_{2}-1}}\right\|_{2,2,1}\left\|\frac{\partial f_{s}}{\partial \mathbf{g}^{l_{2}}}\right\|_{\infty} \\
% & +\sum_{l=l_{2}+1}^{L}\left\|\frac{\partial \mathbf{g}^{l_{1}}}{\partial \mathbf{w}^{l_{1}}}\right\|_F \prod_{l^{\prime}=l_{1}+1}^{l-1}\left\|\frac{\partial \mathbf{g}^{l^{\prime}}}{\partial \mathbf{g}^{l^{\prime}-1}}\right\|_F\left\|\frac{\partial^{2} \mathbf{g}^{l^{\prime}}}{\left(\partial \mathbf{g}^{l^{\prime}-1}\right)^{2}}\right\|_{2,2,1}\left\|\frac{\partial \mathbf{g}^{l_{2}}}{\partial \mathbf{w}^{l_{2}}}\right\|_F\prod_{l^{\prime}=l_{2}+1}^{l}\left\|\frac{\partial \mathbf{g}^{l^{\prime}}}{\partial \mathbf{g}^{l^{\prime}-1}}\right\|_F\left\|\frac{\partial f_{s}}{\partial \mathbf{g}^{l}}\right\|_{\infty} \\
& \leq\left\|\frac{\partial^{2} \mathbf{g}^{l_{1}}}{\left(\partial \mathbf{w}^{\left(l_{1}\right)}\right)^{2}}\right\|_{2,2,1}\left\|\frac{\partial f_{s}}{\partial \mathbf{g}^{l_{1}}}\right\|_{\infty}+L_\sigma^{l_{2}-l_{1}-1}\left\|\frac{\partial \mathbf{g}^{l_{1}}}{\partial \mathbf{w}^{l_{1}}}\right\|_F\\
&\left\|\frac{\partial^{2} \mathbf{g}^{l_{2}}}{\partial \mathbf{w}^{l_{2}} \partial \mathbf{g}^{l_{2}-1}}\right\|_{2,2,1}\left\|\frac{\partial f_{s}}{\partial \mathbf{g}^{l_{2}}}\right\|_{\infty} +\sum_{l=l_{2}+1}^{L} L_\sigma^{2 l-l_{1}-l_{2}}\left\|\frac{\partial \mathbf{g}^{l_{1}}}{\partial \mathbf{w}^{l_{1}}}\right\|_F\\
&\left\|\frac{\partial^{2} \mathbf{g}^{l}}{\left(\partial \mathbf{g}^{l^{\prime}-1}\right)^{2}}\right\|_{2,2,1}\left\|\frac{\partial \mathbf{g}^{l_{2}}}{\partial \mathbf{w}^{l_{2}}}\right\|_F\left\|\frac{\partial f_{s}}{\partial \mathbf{g}^{l}}\right\|_{\infty}.
\end{aligned}
\label{HessianBlockBound}
\end{equation}
Note that \eqref{HessianBlockBound} uses the fact that $\left\|\frac{\partial \mathbf{g}^{l^{\prime}}}{\partial \mathbf{g}^{l^{\prime}-1}}\right\|_F \leq L_\sigma$.
By using the notations given in \eqref{Qinf} and \eqref{Q221}, we get
$$\left\|{H}_{s}^{l_{1}, l_{2}}\right\| \leq C_1 \mathcal{Q}_{2,2,1}\left(f_s\right)\mathcal{Q}_{\infty}\left(f_s\right),$$
where $C_1$ is a constant depend on $L$ and $L_\sigma$. \hfill\qedsymbol

\textit{Proof of \textbf{Lemma \ref{BoundOnHiddenLayer}}:}
For $l=0$, $\|\mathbf{x}^0\|\leq \sqrt{m}\|\mathbf{x}^0\|_\infty\leq \sqrt{m}C_x$, $\|\mathbf{z}^0\|\leq \sqrt{m}\|\mathbf{z}^0\|_\infty\leq \sqrt{m}C_z$, and $\|\mathbf{u}^0\|\leq \sqrt{m}\|\mathbf{u}^0\|_\infty\leq \sqrt{m}C_u$. Whereas for $l = 1, 2,\dots,L$, we have
\begin{equation*}
\footnotesize
    \begin{aligned}
    \left\|\mathbf{x}^{l}\right\| &=\left\|\sigma\left(\frac{W_{1}^{l}}{\sqrt{n}} \mathbf{y}+\frac{W_{2}^{l}}{\sqrt{m}} \mathbf{x}^{l-1}\right)\right\| \\
    % & \leq L_{\sigma}\left\|\frac{W_{1}^{l}}{\sqrt{n}} \mathbf{y}+\frac{W_{2}^{l}}{\sqrt{m}} \mathbf{x}^{l- 1}\right\|+\sigma(0) \\
    & \leq L_{\sigma}\left\|\frac{W_{1}^{l}}{\sqrt{n}}\right\| \|\mathbf{y}\|+L_{\sigma}\left\|\frac{W_{2}^{l}}{\sqrt{m}}\right\| \left\|\mathbf{x}^{l-1}\right\|+\sigma(0) \\
    & \leq L_{\sigma}\left(c_{10}+\frac{R_{1}}{\sqrt{n}}\right) \sqrt{n} C_{\mathbf{y}}+L_{\sigma}\left(c_{20}+\frac{R_{2}}{\sqrt{m}}\right) c_{\mathrm{ISTA} ; x}^{l-1}+\sigma(0)\\
    &=c_{\mathrm{ISTA} ; \mathbf{x}}^{l}.
\end{aligned}
\normalsize
\end{equation*}  
Here, we used  Lemma \ref{BoundOnWeight} and $L_\sigma$-Lipschitz continuous of the activation function $\sigma(\cdot)$. Similarly, 
\begin{equation*}
\footnotesize
    \begin{aligned}
    & \left\|\mathbf{z}^{l}\right\|=\left\|\sigma\left(\frac{1}{\sqrt{n}} W_{1}^{l} \mathbf{y}+\frac{1}{\sqrt{m}} W_{2}^{l}\left(\mathbf{z}^{l-1}-\mathbf{u}^{l-1}\right)+\mathbf{u}^{l-1}\right)\right\| \\
    & \leq L_{\sigma} \frac{1}{\sqrt{n}}\left\|W_{1}^{l}\right\|\|\mathbf{y}\|+L_{\sigma} \frac{1}{\sqrt{m}}\left\|W_{2}^{l}\right\|\left\|\mathbf{z}^{l-1}\right\|+\frac{1}{\sqrt{m}} L_{\sigma}\left\|W_{2}^{l}\right\|\left\|\mathbf{u}^{l-1}\right\|\\
    &\quad+L_{\sigma}\left\|\mathbf{u}^{l-1}\right\|+\sigma(0) \\
    & \leq L_{\sigma}\left(c_{10}+\frac{R_{1}}{\sqrt{n}}\right) \sqrt{n} C_{y}+L_{\sigma}\left(c_{20}+\frac{R_{2}}{\sqrt{m}}\right) c_{\mathrm{ADMM} ; \mathbf{z}}^{l-1}\\
    &\quad+L_{\sigma}\left(1+c_{20}+\frac{R_{2}}{\sqrt{m}}\right) c_{\mathrm{ADMM} ; \mathbf{u}}^{l-1}+\sigma(0) \\
    & =c_{\mathrm{ADMM} ; \mathbf{z}}^{l} 
\end{aligned}
\normalsize
\end{equation*}
and 
\begin{equation*}
\footnotesize
    \begin{aligned}
    & \left\|\mathbf{u}^{l}\right\|=\left\|\mathbf{u}^{l-1}+\left(\frac{1}{\sqrt{n}} W_{1}^{l} \mathbf{y}+\frac{1}{\sqrt{m}} W_{2}^{l}\left(\mathbf{z}^{l-1}-\mathbf{u}^{l-1}\right)-\mathbf{z}^{l}\right)\right\| \\
    & \leq\left\|\mathbf{u}^{l-1}\right\|+\left\|\frac{1}{\sqrt{n}} W_{1}^{l} \mathbf{y}\right\|+\left\|\frac{1}{\sqrt{m}} W_{2}^{l} \mathbf{z}^{l-1}\right\|+\left\|\frac{1}{\sqrt{m}} W_{2}^{l} \mathbf{u}^{l-1}\right\|+\left\|\mathbf{z}^{l}\right\| \\
    & \leq \left(c_{10}+\frac{R_{1}}{\sqrt{n}}\right) \sqrt{n} C_{y}+\left(c_{20}+\frac{R_{2}}{\sqrt{m}}\right) c_{\mathrm{ADMM} ; \mathbf{z}}^{l-1}\\
    &\quad +\left(c_{20}+\frac{R_{2}}{\sqrt{m}}+1\right) c_{\mathrm{ADMM} ; \mathbf{u}}^{l-1}+ c_{\mathrm{ADMM} ; \mathbf{z}}^{l} \\
    & =c_{\mathrm{ADMM} ; \mathbf{u}}^{l}
    \end{aligned}
    \normalsize
\end{equation*}
\hfill\qedsymbol

\textit{Proof of \textbf{Theorem} \ref{EigenValue_Bounds}:}
Consider the real symmetric NTK matrix $[K\left(\mathbf{w}_0\right)]_{mT\times mT}$.
Utilizing the Rayleigh quotient of $K\left(\mathbf{w}_0\right)$, we can write the following for any $\mathbf{x}$ such that $\|\mathbf{x}\|_{2}=1$:
$$\lambda_{\min }\left(K\left(\mathbf{w}_0\right)\right) \leq \mathbf{x}^{\top} K\left(\mathbf{w}_0\right)\mathbf{x} \leq \lambda_{\max }(K\left(\mathbf{w}_0\right)).$$
Let $\mathbf{x}$ be a vector having all zeros except the $s^{\text{th}}$ component to be $1$. Thus $\lambda_{\min }(K\left(\mathbf{w}_0\right)) \leq[K\left(\mathbf{w}_0\right)]_{\mathrm{s,s}}$, for any $s\in[mT]$. Assume $s=1$, this implies,
\begin{equation}
    \lambda_{\min }\left(K\left(\mathbf{w}_{0}\right)\right) \leq\left\langle\nabla_{\mathbf{w}_0} \mathbf{f}_{1}, \nabla_{\mathbf{w}_0} \mathbf{f}_{1}\right\rangle,
    \label{min_eig}
\end{equation}
where $\mathbf{f}_1$ is the $1^{\text{st}}$ component in the the model output vector $\mathbf{f}$ corresponding to the first training sample.
We now aim to compute $\left\langle\nabla_{\mathbf{w}_0} \mathbf{f}_{1}, \nabla_{\mathbf{w}_0} \mathbf{f}_{1}\right\rangle$ for FFNN, LISTA, and ADMM-CSNet.

Consider a one-layer FFNN, then  from \eqref{L-layered_ffnn}, the $s^{\text{th}}$ component of $\mathbf{f}_{\text{FFNN}}$ is, $   \mathbf{f}_{s}=\frac{1}{\sqrt{m}} \sigma\left(\frac{1}{\sqrt{n}} W_0^{1}(s,:) \mathbf{y}\right),$  where $W_0^1(s,:)$ represents the $s^{\text{th}}$ row of $W_0^1.$
This implies,
\begin{equation*}
    \left\langle\nabla_{W_0^1} \mathbf{f}_{s}, \nabla_{W_0^1} \mathbf{f}_{s}\right\rangle
    =\left[\frac{\sigma^{\prime}(\tilde{\mathbf{x}}_{s}^{1})}{\sqrt{m n}}\right]^{2}\|\mathbf{y}\|^{2} \leq \hat{L}^2 \hat{y},
 \label{1-layered_ffnn_bound}
\end{equation*}
where $\hat{L}=\frac{L_{\sigma}}{\sqrt{m}}, \text{ and } \hat{y} = \frac{\|\mathbf{y}\|^2}{n}$.
Similarly, for a 2-layered FFNN, we have
% defined as $ \mathbf{f}=\frac{1}{\sqrt{m}} \sigma\left(\frac{W^{2}}{\sqrt{m}} \sigma\left(\frac{W^{1}}{\sqrt{n}} \mathbf{y}\right)\right) \in \mathbb{R}^m ; W^{2} \in \mathbb{R}^{m \times m}, \ \  W^{1} \in \mathbb{R}^{m \times n}.$
\begin{equation*}
    \label{2-layered_ffnn_bound}
    \begin{aligned}    \left\langle\nabla_{\mathbf{W}_0} \mathbf{f}_{s}, \nabla_{\mathbf{W}_0} \mathbf{f}_{s}\right\rangle &= \left\langle\nabla_{W_0^{1}} \mathbf{f}_{s}, \nabla_{W_0^{1}} \mathbf{f}_{s}\right\rangle + \left\langle\nabla_{W_0^{2}} \mathbf{f}_{s}, \nabla_{W_0^{2}} \mathbf{f}_{s}\right\rangle\\
    &\leq  (\hat{L}^2)^2 \hat{y} \left[ \left\|W_0^{1}\right\|^{2} + \left\|W_0^{2}(s, :)\right\|^{2} \right].
\end{aligned}
\end{equation*}
Generalizing the above equations, one can derive the upper bound on $\lambda_{0, \text{FFNN}}$ for an L-layer FFNN as
    \begin{equation}
        \footnotesize
        \begin{aligned}
        \lambda_{0, \text{FFNN}}&\leq\text{UB}_{\text{FFNN}} \\& = \hat{L}^{2L} \hat{y}  \left[ \sum_{i=1}^{L-1} \|\mathbf{v}_s^{T} W_0^L\|^2\prod_{j=1, j \neq i}^{L-1} \|W_0^j\|^2  +\prod_{j=1}^{L-1}\|W_0^j\|^2 \right]. 
        \end{aligned}
    \label{Minimum_eigenvalue_NTK_FFNN}
    \end{equation} 
Likewise, consider $L=1$, then from \eqref{LISTA_Model}, the $s^{\text{th}}$ component of $\mathbf{f}_{\text{LISTA}}$ is
$$\mathbf{f}_{s}=\frac{1}{\sqrt{m}} \sigma\left(\frac{1}{\sqrt{n}} W_{10}^{1}(s,:) \mathbf{y}+\frac{1}{\sqrt{m}} W_{20}^{1}(s,:) \mathbf{x}\right).$$
 This implies,
\begin{equation*}
    \label{1-layerd_unfolded_bound}
    \begin{aligned}
    \left\langle\nabla_{\mathbf{w}_0} \mathbf{f}_{s}, \nabla_{\mathbf{w}_0} \mathbf{f}_{s}\right\rangle &=\left\langle\nabla_{W_{10}^1} \mathbf{f}_{s}, \nabla_{W_{10}^1} \mathbf{f}_{s}\right\rangle + \left\langle\nabla_{W_{20}^1} \mathbf{f}_{s}, \nabla_{W_{20}^1} \mathbf{f}_{s}\right\rangle \\
    &\leq \hat{L}^2\left[\hat{y} + \hat{x}\right],
    \end{aligned}
\end{equation*}
where $\hat{x} = \frac{\|\mathbf{x}\|^2}{m}$.
If $L=2$, then the $s^{\text{th}}$ component of $\mathbf{f}_{\text{LISTA}}$ is
\begin{equation*}
    \begin{aligned}
    \left\langle\nabla_{\mathbf{w}_0} \mathbf{f}_{s}, \nabla_{\mathbf{w}_0} \mathbf{f}_{s}\right\rangle 
    &=\left\langle\nabla_{W_{10}^{2}} \mathbf{f}_{s}, \nabla_{W_{10}^{2}} \mathbf{f}_{s}\right\rangle+\left\langle\nabla_{W_{20}^{2}} \mathbf{f}_{s}, \nabla_{W_{20}^{2}} \mathbf{f}_{s}\right\rangle  \\ &+\left\langle\nabla_{W_{10}^{1}} \mathbf{f}_{s}, \nabla_{W_{10}^{1}} \mathbf{f}_{s}\right\rangle+\left\langle\nabla_{W_{20}^{1}} \mathbf{f}_{s}, \nabla_{W_{20}^{1}} \mathbf{f}_{s}\right\rangle \\
    &\hspace{-0.4cm}\leq \hat{L}^{2}\left[\hat{y}+\hat{L}^{2}\|\tilde{\mathbf{x}}^{(1)}\|^{2}\right]  +\hat{L}^{4}\left[\hat{y} + \hat{x}\right]\left\|\mathbf{v}_{s}^{\top} W_{20}^{2}\right\|^{2}.
\end{aligned}
\end{equation*}
By extending the above equations, we obtain the upper bound on $\lambda_{0, \text{LISTA}}$ for an $L$-layer LISTA as 
    \begin{equation}
        \footnotesize
        \begin{aligned}
        &\lambda_{0, \text{LISTA}}\leq \text{UB}_{\text{LISTA}} = \hat{L}^2 \left( \hat{y}+ \hat{x} \right), \ \ \text{for} \ \ L=1 \\
         & \lambda_{0, \text{LISTA}}\leq\text{UB}_{\text{LISTA}} = \hat{L}^{2L} \left( \hat{y}+ \hat{x} \right) \|\mathbf{v}_s^{T} W_{20}^L\|^2 \prod_{l=2}^{L-1} \|W_{20}^l\|^2\\
         &+\sum_{k=2}^{L-1}  \hat{L}^{2L-2k+2} \left[  \hat{y} + \hat{L}^2 \left\|\tilde{\mathbf{x}}^{(k-1)} \right\|^2 \right] \|\mathbf{v}_s^{T} W_{20}^L\|^2\prod_{l=k+1}^{L-1} \|W_{20}^l\|^2 \\ 
         % & \\
         &+  \hat{L}^{2} \left[ \hat{y} +\hat{L}^2 \|\tilde{x}^{(L-1)}\|^2\right], \text{ for } L>1,
        \end{aligned}
\label{Minimum_eigenvalue_NTK_Unfolded}
\end{equation}
where $\hat{L}=\frac{L_{\sigma}}{\sqrt{m}},\  \hat{y} = \frac{\|\mathbf{y}\|^2}{n},\ \text{ and } \hat{x} =\frac{\|\mathbf{x}\|^2}{m}.$
Repeating the same analysis, one can derive the upper bound on $\lambda_{0, \text{ADMM-CSNet}} $ of an $L$-layer ADMM-CSNet as
 \begin{equation}
\label{minimum_eig_value_NTK_admm}
        \footnotesize
        \begin{aligned}
            &\lambda_{0, \text{ADMM-CSNet}}\leq \text{UB}_{\text{ADMM-CSNet}}= \hat{L}^2\left[ \hat{y} + \hat{a}^{(L-1)}\right] \\& + \sum_{k=1}^{L-1} \hat{L}^{2L-2k+2} \left[ \hat{y} + \hat{a}^{(k-1)}\right] \|\mathbf{v}_s^{T} W_{20}^L\|^2\prod_{l=k+1}^{L-1} \|W_{20}^l\|^2 ,
        \end{aligned}
    \end{equation}
where $\hat{a}^{(l)} = \frac{\|\mathbf{z}^{(l)}- \mathbf{u}^{(l)} \|^2}{m},\ \forall l \in [L-1]\cup \{0\}$.
\hfill\qedsymbol

\ifCLASSOPTIONcaptionsoff
  \newpage
\fi
\bibliographystyle{ieeetr}
\bibliography{bibs_1}

\begin{thebibliography}{10}

\bibitem{EldarCS}
Y.~C. Eldar and G.~Kutyniok, {\em Compressed sensing: theory and applications}.
\newblock Cambridge University Press, 2012.

\bibitem{DonohoCS}
D.~Donoho, ``Compressed sensing,'' {\em IEEE Trans. Inf. Theory}, vol.~52,
  no.~4, pp.~1289--1306, 2006.

\bibitem{Nir1}
N.~Shlezinger, J.~Whang, Y.~C. Eldar, and A.~G. Dimakis, ``Model-{Based} {Deep}
  {Learning},'' arXiv:2012.08405, 2020.

\bibitem{Nir2}
N.~Shlezinger, J.~Whang, Y.~C. Eldar, and A.~G. Dimakis, ``{Model-Based Deep
  Learning:} {Key} {Approaches} {and} {Design} {Guidelines},'' in {\em Proc.
  IEEE Data Sci. Learn. Workshop (DSLW)}, pp.~1--6, 2021.

\bibitem{UnrollingPerformance}
K.~Gregor and Y.~LeCun, ``Learning fast approximations of sparse coding,'' in
  {\em Proc. Int. Conf. Mach. Learn.}, pp.~399--406, 2010.

\bibitem{Unrolling}
V.~Monga, Y.~Li, and Y.~C. Eldar, ``{Algorithm Unrolling: Interpretable,
  Efficient Deep Learning for Signal and Image Processing},'' {\em IEEE Signal
  Process. Mag.}, vol.~38, no.~2, pp.~18--44, 2021.

\bibitem{ADMM-CSNet}
Y.~Yang, J.~Sun, H.~Li, and Z.~Xu, ``{ADMM-CSNet: A Deep Learning Approach for
  Image Compressive Sensing},'' {\em IEEE Trans. Pattern Anal. Mach. Intell.},
  vol.~42, no.~3, pp.~521--538, 2020.

\bibitem{Deblurring}
Y.~Li, M.~Tofighi, J.~Geng, V.~Monga, and Y.~C. Eldar, ``{Efficient and
  Interpretable Deep Blind Image Deblurring Via Algorithm Unrolling},'' {\em
  IEEE Trans. Med. Imag.}, vol.~6, pp.~666--681, 2020.

\bibitem{ImageSuperResolution}
Z.~Wang, D.~Liu, J.~Yang, W.~Han, and T.~Huang, ``Deep {Networks} for {Image}
  {Super}-{Resolution} {With} {Sparse} {Prior},'' in {\em Proc. IEEE Int. Conf.
  Comput. Vis.}, December 2015.

\bibitem{LearnedSparcom}
G.~Dardikman-Yoffe and Y.~C. Eldar, ``Learned {SPARCOM}: unfolded deep
  super-resolution microscopy,'' {\em Opt. Express}, vol.~28, pp.~27736--27763,
  Sep 2020.

\bibitem{Ultrasound}
O.~Solomon, R.~Cohen, Y.~Zhang, Y.~Yang, Q.~He, J.~Luo, R.~J.~G. van Sloun, and
  Y.~C. Eldar, ``Deep {Unfolded} {Robust} {PCA} {With} {Application} to
  {Clutter} {Suppression} in {Ultrasound},'' {\em IEEE Trans. Med. Imag.},
  vol.~39, no.~4, pp.~1051--1063, 2020.

\bibitem{PowerSystem}
L.~Zhang, G.~Wang, and G.~B. Giannakis, ``{Real-Time Power System State
  Estimation and Forecasting via Deep Unrolled Neural Networks},'' {\em IEEE
  Trans. Signal Process.}, vol.~67, no.~15, pp.~4069--4077, 2019.

\bibitem{Avner}
A.~Shultzman, E.~Azar, M.~R.~D. Rodrigues, and Y.~C. Eldar, ``{Generalization
  and Estimation Error Bounds for Model-based Neural Networks},'' in {\em Proc.
  Int. Conf. Learn. Represent.}, 2023.

\bibitem{Generalization1}
E.~Schnoor, A.~Behboodi, and H.~Rauhut, ``{Generalization Error Bounds for
  Iterative Recovery Algorithms Unfolded as Neural Networks},'' {\em
  arXiv.2112.04364}, 2022.

\bibitem{Generalization2}
A.~Behboodi, H.~Rauhut, and E.~Schnoor, ``{Compressive Sensing and Neural
  Networks from a Statistical Learning Perspective},'' {\em arXiv.2010.15658},
  2021.

\bibitem{liu2018alista}
J.~Liu, X.~Chen, Z.~Wang, and W.~Yin, ``{ALISTA: Analytic Weights Are As Good
  As Learned Weights in LISTA},'' in {\em Proc. Int. Conf. Learn. Represent.},
  2019.

\bibitem{NEURIPS2018_cf8c9be2}
X.~Chen, J.~Liu, Z.~Wang, and W.~Yin, ``{Theoretical Linear Convergence of
  Unfolded ISTA and Its Practical Weights and Thresholds},'' in {\em Proc. Adv.
  Neural Inf. Process. Syst.}, vol.~31, Curran Associates, Inc., 2018.

\bibitem{LISTACOnvergence1}
X.~Chen, J.~Liu, Z.~Wang, and W.~Yin, ``{Hyperparameter Tuning is All You Need
  for LISTA},'' in {\em Proc. Adv. Neural Inf. Process. Syst.}, vol.~34,
  pp.~11678--11689, Curran Associates, Inc., 2021.

\bibitem{ClassicalML}
T.~Hastie, R.~Tibshirani, J.~H. Friedman, and J.~H. Friedman, {\em {The
  elements of statistical learning: data mining, inference, and prediction}},
  vol.~2.
\newblock Springer, 2009.

\bibitem{Double_Descent1}
M.~Belkin, D.~Hsu, S.~Ma, and S.~Mandal, ``{Reconciling modern machine-learning
  practice and the classical bias–variance trade-off},'' {\em Proc. Nat.
  Acad. Sci.}, vol.~116, no.~32, pp.~15849--15854, 2019.

\bibitem{Double_Descent2}
M.~Belkin, ``{Fit without fear: remarkable mathematical phenomena of deep
  learning through the prism of interpolation},'' {\em Acta Numerica}, vol.~30,
  p.~203–248, 2021.

\bibitem{Double_Descent3}
C.~Zhang, S.~Bengio, M.~Hardt, B.~Recht, and O.~Vinyals, ``{Understanding Deep
  Learning (Still) Requires Rethinking Generalization},'' {\em Commun. ACM},
  vol.~64, p.~107–115, feb 2021.

\bibitem{Double_Descent4}
P.~Nakkiran, G.~Kaplun, Y.~Bansal, T.~Yang, B.~Barak, and I.~Sutskever, ``{Deep
  Double Descent: Where Bigger Models and More Data Hurt},'' in {\em Proc. Int.
  Conf. Learn. Represent.}, 2020.

\bibitem{Double_Descent5}
S.~Spigler, M.~Geiger, S.~d’Ascoli, L.~Sagun, G.~Biroli, and M.~Wyart, ``{A
  jamming transition from under- to over-parametrization affects generalization
  in deep learning},'' {\em J. Phys. A}, vol.~52, p.~474001, oct 2019.

\bibitem{Double_Descent6}
M.~Belkin, S.~Ma, and S.~Mandal, ``{To Understand Deep Learning We Need to
  Understand Kernel Learning},'' in {\em Proc. Int. Conf. Mach. Learn.},
  vol.~80, pp.~541--549, PMLR, 10--15 Jul 2018.

\bibitem{LossLandscape}
C.~Liu, L.~Zhu, and M.~Belkin, ``{Loss landscapes and optimization in
  over-parameterized non-linear systems and neural networks},'' {\em Appl.
  Comput. Harmon. Anal.}, vol.~59, pp.~85--116, 2022.

\bibitem{GD1}
S.~S. Du, X.~Zhai, B.~Poczos, and A.~Singh, ``{Gradient Descent Provably
  Optimizes Over-parameterized Neural Networks},'' in {\em Proc. Int. Conf.
  Learn. Represent.}, 2019.

\bibitem{GD2}
S.~Du, J.~Lee, H.~Li, L.~Wang, and X.~Zhai, ``{Gradient Descent Finds Global
  Minima of Deep Neural Networks},'' in {\em Int. Conf. Mach. Learn.}, vol.~97,
  pp.~1675--1685, PMLR, 09--15 Jun 2019.

\bibitem{GD3}
Z.~Allen-Zhu, Y.~Li, and Z.~Song, ``{A convergence theory for deep learning via
  over-parameterization},'' in {\em Proc. Int. Conf. Mach. Learn.},
  pp.~242--252, PMLR, 2019.

\bibitem{GD4}
D.~Zou, Y.~Cao, D.~Zhou, and Q.~Gu, ``{Stochastic Gradient Descent Optimizes
  Over-parameterized Deep ReLU Networks},'' {\em CoRR}, vol.~abs/1811.08888,
  2018.

\bibitem{Linearity}
C.~Liu, L.~Zhu, and M.~Belkin, ``On the linearity of large non-linear models:
  when and why the tangent kernel is constant,'' {\em Proc. Adv. Neural Inf.
  Process. Syst.}, vol.~33, pp.~15954--15964, 2020.

\bibitem{Expected_error}
D.~Jakubovitz, R.~Giryes, and M.~R. Rodrigues, ``Generalization error in deep
  learning,'' in {\em Compressed Sensing and Its Applications: Third
  International MATHEON Conference 2017}, pp.~153--193, Springer, 2019.

\bibitem{LASSO}
R.~Tibshirani, ``{Regression Shrinkage and Selection via the Lasso},'' {\em J.
  Roy. Statist Soc. Ser. B (Methodol.)}, vol.~58, no.~1, pp.~267--288, 1996.

\bibitem{ProximalAlg}
N.~Parikh and S.~Boyd, ``{Proximal Algorithms},'' {\em Found. Trends Optim.},
  vol.~1, no.~3, pp.~127--239, 2014.

\bibitem{ISTA}
I.~Daubechies, M.~Defrise, and C.~De~Mol, ``An iterative thresholding algorithm
  for linear inverse problems with a sparsity constraint,'' {\em Communications
  on Pure and Applied Mathematics: A Journal Issued by the Courant Institute of
  Mathematical Sciences}, vol.~57, no.~11, pp.~1413--1457, 2004.

\bibitem{ADMMBook}
S.~Boyd, N.~Parikh, E.~Chu, B.~Peleato, and J.~Eckstein, {\em {Distributed
  Optimization and Statistical Learning via the Alternating Direction Method of
  Multipliers}}.
\newblock 2011.

\bibitem{PL}
B.~T. Polyak, ``{Gradient methods for minimizing functionals},'' {\em Ž.
  Vyčisl. Mat. Mat. Fiz.}, vol.~3, no.~4, pp.~643--653, 1963.

\bibitem{PL1}
S.~Lojasiewicz, ``{A topological property of real analytic subsets},'' {\em
  Coll. du CNRS, Les {\'e}quations aux d{\'e}riv{\'e}es partielles}, vol.~117,
  no.~87-89, p.~2, 1963.

\bibitem{SmoothActivation1}
Y.~Ben~Sahel, J.~P. Bryan, B.~Cleary, S.~L. Farhi, and Y.~C. Eldar, ``{Deep
  Unrolled Recovery in Sparse Biological Imaging: Achieving fast, accurate
  results},'' {\em IEEE Signal Process. Mag.}, vol.~39, no.~2, pp.~45--57,
  2022.

\bibitem{SmoothActivation2}
A.~M. Atto, D.~Pastor, and G.~Mercier, ``{Smooth sigmoid wavelet shrinkage for
  non-parametric estimation},'' in {\em Proc. IEEE Int. Conf. Acoust., Speech,
  Signal Process.}, pp.~3265--3268, 2008.

\bibitem{SmoothActivation3}
X.-P. Zhang, ``{Thresholding neural network for adaptive noise reduction},''
  {\em IEEE Trans. Neural Netw.}, vol.~12, no.~3, pp.~567--584, 2001.

\bibitem{SmoothActivation4}
X.-P. Zhang, ``{Space-scale adaptive noise reduction in images based on
  thresholding neural network},'' in {\em Proc. IEEE Int. Conf. Acoust.,
  Speech, Signal Process.}, vol.~3, pp.~1889--1892 vol.3, 2001.

\bibitem{SmoothActivation5}
H.~Pan, D.~Badawi, and A.~E. Cetin, ``{Fast} {Walsh}-{Hadamard} {Transform} and
  {Smooth}-{Thresholding} {Based} {Binary} {Layers} in {Deep} {Neural}
  {Networks},'' in {\em Proc. IEEE/CVF Conf. Comput. Vis. Pattern Recognit.
  (CVPR)}, pp.~4650--4659, June 2021.

\bibitem{SmoothActivation6}
J.~Youn, S.~Ravindran, R.~Wu, J.~Li, and R.~van Sloun, ``{Circular
  Convolutional Learned ISTA for Automotive Radar DOA Estimation},'' in {\em
  Proc. 19th Eur. Radar Conf. (EuRAD)}, pp.~273--276, 2022.

\bibitem{SmoothActivation7}
K.~Kavukcuoglu, P.~Sermanet, Y.-l. Boureau, K.~Gregor, M.~Mathieu, and Y.~Cun,
  ``{Learning} {Convolutional} {Feature} {Hierarchies} for {Visual}
  {Recognition},'' in {\em Proc. Adv. Neural Inf. Process. Syst.}, vol.~23,
  Curran Associates, Inc., 2010.

\bibitem{WeightBound}
R.~Vershynin, ``{Introduction to the non-asymptotic analysis of random
  matrices},'' {\em arXiv.1011.3027}, 2010.

\bibitem{TangentKernel}
A.~Jacot, F.~Gabriel, and C.~Hongler, ``{Neural Tangent Kernel: Convergence and
  Generalization in Neural Networks},'' in {\em Proc. Adv. Neural Inf. Process.
  Syst.}, vol.~31, 2018.

\bibitem{NEURIPS2019_0d1a9651}
J.~Lee, L.~Xiao, S.~Schoenholz, Y.~Bahri, R.~Novak, J.~Sohl-Dickstein, and
  J.~Pennington, ``{Wide Neural Networks of Any Depth Evolve as Linear Models
  Under Gradient Descent},'' in {\em Proc. Adv. Neural Inf. Process. Syst.},
  vol.~32, Curran Associates, Inc., 2019.

\bibitem{SupMaterial}
S.~B. Shah, P.~Pradhan, W.~Pu, R.~Ramunaidu, M.~R.~D. Rodrigues, and Y.~C.
  Eldar, ``{Supporting Material: Optimization Guarantees of Unfolded ISTA and
  ADMM Networks With Smooth Soft-Thresholding},'' {\em 2023}.

\end{thebibliography}


\begin{thebibliography}{1}

\bibitem{appendixref}
B.~Laurent and P.~Massart, ``Adaptive estimation of a quadratic functional by model selection,'' {\em Annals of statistics}, pp.~1302--1338, 2000.

\end{thebibliography}

\newpage
\onecolumn

\center\Large{\textbf{Supporting Material: Optimization Guarantees of Unfolded ISTA and ADMM Networks With Smooth Soft-Thresholding}}
\normalsize
\vspace{1cm}

\justifying

From Theorem 3, the Hessian spectral norm, $\|\mathbf{H}\|_2$, of an $L$-layer unfolded ISTA (ADMM) network is bounded as 
\begin{equation}
    \begin{aligned}
        \|\mathbf{H}\|_2 &\leq \sum_{s, l_{1}, l_{2}} C_1 \mathcal{Q}_{2,2,1}\left(f_{s}\right) \mathcal{Q}_{\infty}\left(f_{s}\right)\\ &\leq \sum_{s=1}^{m} C \mathcal{Q}_{2,2,1}\left(f_{s}\right) \mathcal{Q}_{\infty}\left(f_{s}\right),
    \end{aligned}
\end{equation}
where the constant $C_1$ depends on $L$ and $L_\sigma$, $C = L^2C_1$,
\begin{equation}
    \mathcal{Q}_{\infty}\left(f_s\right)= \max _{1 \leq l \leq L}\left\{\left\|\frac{\partial f_s}{\partial \mathbf{g}^{l}}\right\|_{\infty}\right\}
    \label{Qinf}\text{ and}
\end{equation}
\begin{equation}
\mathcal{Q}_{2,2,1}\left(f_s\right)= \max _{1 \leq l_{1} \leq l_{2} < l_{3} \leq L}\Bigg\{\left\|\frac{\partial^{2} \mathbf{g}^{l_1}}{\left(\partial \mathbf{w}^{l_1}\right)^{2}}\right\|_{2,2,1},
\left\|\frac{\partial \mathbf{g}^{l_1}}{\partial \mathbf{w}^{l_{1}}}\right\|\left\|\frac{\partial^{2} \mathbf{g}^{l_2}}{\partial \mathbf{g}^{\left(l_{2}-1\right)} \partial \mathbf{w}^{l_2}}\right\|_{2,2,1}, \left\|\frac{\partial \mathbf{g}^{l_{1}}}{\partial \mathbf{w}^{l_{1}}}\right\|\left\|\frac{\partial \mathbf{g}^{l_2}}{\partial \mathbf{w}^{l_2}}\right\|\left\|\frac{\partial^{2} \mathbf{g}^{l_{3}}}{\left(\partial \mathbf{g}^{l_{3}-1}\right)^{2}}\right\|_{2,2,1}\Bigg\}.
    \label{Q221}
\end{equation}
   Note that $\mathbf{g}^l = \mathbf{x}^l$ for LISTA and $\mathbf{g}^l = \mathbf{z}^l$ for ADMM-CSNet.
\section{Proof of Theorem $4$}
Theorem $4$ aims to provide bounds on $\mathcal{Q}_{\infty}\left(f_s\right)$ and $\mathcal{Q}_{2,2,1}\left(f_s\right)$. The proof of this theorem has been divided into two parts: First, we prove the bound on $Q_{2,2,1}$ in sub-sections \ref{Q_2_2_1_For_LISTA} and \ref{Q_2_2_1_For_ADMMCSNet}, respectively. Then, we prove the bound on $Q_\infty$ in sub-sections \ref{Q_Infinity_For_LISTA} and \ref{Q_Infinity_For_ADMMCSNet}, respectively.
Here we denote $\|\cdot \|$ as $l_2$-norm for vectors and spectral norm for matrices. We also denote $\|\cdot \|_F$ as the Frobenious norm of matrices.

\subsection{Bound on $Q_{2,2,1}$ For LISTA Network}
Consider an L-layer unfolded ISTA network with output 
\label{Q_2_2_1_For_LISTA}
\begin{equation}
\begin{aligned}
    \label{eq:1}
    \mathbf{f}&=\frac{1}{\sqrt{m}} \mathbf{x}^L, \text{ where }\\
    \mathbf{x}^{l} &= \sigma(\tilde{\mathbf{x}}^{l}) = \sigma\left(\frac{W_{1}^{l}}{\sqrt{n}} \mathbf{y}+\frac{W_{2}^{l}}{\sqrt{m}} \mathbf{x}^{l-1}\right)\ \in \mathbb{R}^m,\ l \in [L].
\end{aligned}
\end{equation}
Now the first derivatives of $\mathbf{x}^{l}$ are
\begin{equation*}
\begin{aligned}
&\left(\frac{\partial \mathbf{x}^{l}}{\partial \mathbf{x}^{l-1}}\right)_{i, j}=\frac{1}{\sqrt{m}} \sigma^{\prime}\left(\tilde{\mathbf{x}}_{i}^{l}\right)\left(W_{2}\right)_{i, j}^{l}, \left(\frac{\partial \mathbf{x}^{l}}{\partial W_{1}^{l}}\right)_{i, j j^{\prime}}=\frac{1}{\sqrt{n}} \sigma^{\prime}\left(\tilde{\mathbf{x}}_{i}^{l}\right) \mathbf{y}_{j^{\prime}} \mathbb{I}_{i=j}, \left(\frac{\partial \mathbf{x}^{l}}{\partial W_{2}^{l}}\right)_{i, j j^{\prime}}=\frac{1}{\sqrt{m}} \sigma^{\prime}\left(\tilde{\mathbf{x}}_{i}^{l}\right) \mathbf{x}_{j^{\prime}}^{l-1} \mathbb{I}_{i=j}.
\end{aligned}
\end{equation*} 
By the definition of spectral norm, $\|A\|_2=\sup _{\|\mathbf{v}\|_2=1}\|A \mathbf{v}\|_2$, we have
%\begin{equation*}
%\begin{aligned}
%\left\|\frac{\partial \mathbf{x}^{l}}{\partial \mathbf{x}^{l-1}}\right\|_2^{2} &=\sup _{\|\mathbf{v}\|_2=1} \frac{1}{m} \sum_{i}\left(\sigma^{\prime}\left(\tilde{\mathbf{x}}_{i}^{l}\right)\left(W_{2}\right)_{i, j}^{l} \mathbf{v}_{j}\right)^{2} =\sup _{\|\mathbf{v}\|_2=1} \frac{1}{m}\left\|\Sigma^{\prime l} W_{2}^{l} \mathbf{v}\right\|^{2} \leq \frac{1}{m}\left\|\Sigma^{\prime l}\right\|^{2}\left\|W_{2}^{l}\right\|^{2} \leq  L_{\sigma}^{2}\left(c_{20}+R_{2} / \sqrt{m}\right)^{2} \\

%\end{aligned}
%\end{equation*}

\begin{equation*}
  \begin{aligned}
\left\|\frac{\partial \mathbf{x}^{l}}{\partial W_{1}^{l}}\right\|^{2} =\sup _{\|V\|_{F}=1} \frac{1}{n} \sum_{i}\left(\sum_{j, j^{\prime}} \sigma^{\prime}\left(\tilde{\mathbf{x}}_{i}^{l}\right) \mathbf{y}_{j^{\prime}}  V_{j, j^{\prime}}\mathbb{I}_{i=j}\right)^{2}  =\sup _{\|V\|_{F}=1} \frac{1}{n}\left\|\Sigma^{\prime l} V \mathbf{y}\right\|^{2} \leq \frac{1}{n}\left\|\Sigma^{\prime l}\right\|^{2}\|\mathbf{y}\|^{2} \leq L_{\sigma}^{2} C_{y}^{2}=O(1),
\end{aligned}  
\end{equation*}
where $\Sigma^{\prime l}$ is a diagonal matrix with the diagonal entry $(\Sigma^{\prime l})_{ii}=\sigma^{\prime}\left(\tilde{\mathbf{x}}_{i}^{l}\right)$.
Similarly,
\begin{equation*}
   \begin{aligned}
\left\|\frac{\partial \mathbf{x}^{l}}{\partial W_{2}^{l}}\right\|^{2}&=\sup _{\|V\|_{F}=1} \frac{1}{m} \sum_{i}\left(\sum_{j, j^{\prime}} \sigma^{\prime}\left(\tilde{\mathbf{x}}_{i}^{l}\right) \mathbf{x}_{j^{\prime}}^{l-1}  V_{j,j^{\prime}} \mathbb{I}_{i=j}\right)^{2} =\sup _{\|V\|_{F}=1} \frac{1}{m}\left\|\Sigma^{\prime l} V \mathbf{x}^{l-1}\right\|^{2}\\
&\leq \frac{1}{m} L_{\sigma}^{2}\left\|\mathbf{x}^{l-1}\right\|^{2}  \leq \frac{1}{m} L_{\sigma}^{2}\left(c_{\text {ISTA } ; \mathbf{x}}^{l-1}\right)^{2}=O(1).
\end{aligned} 
\end{equation*}
Here we used $\left(c_{\text {ISTA } ; \mathbf{x}}^{l-1}\right)=O(\sqrt{m})$ from lemma (4).
\begin{equation}
\label{eq:2}
\left\|\frac{\partial \mathbf{x}^{l}}{\partial W^{l}}\right\| =\left\|\left[\frac{\partial \mathbf{x}^{l}}{\partial W_{1}^{l}} \quad \frac{\partial \mathbf{x}^{l}}{\partial W_{2}^{l}}\right]\right\| 
 \leq\left\|\frac{\partial \mathbf{x}^{l}}{\partial W_{1}^{l}}\right\|+\left\|\frac{\partial \mathbf{x}^{l}}{\partial W_{2}^{l}}\right\|=O(1)+O(1)=O(1).
 %\leq L_{\sigma}^{2} C_{y}^{2}+\frac{1}{m} L_{\sigma}^{2}\left(c_{\mathrm{ISTA} ; \mathbf{x}}^{l-1}\right)^{2} = O(1)
\end{equation}

The second-order derivatives of the vector-valued layer function $\mathbf{x}^{l}$, which are order 3 tensors, have the following expressions:
\begin{equation*}
\begin{aligned}
& \left(\frac{\partial^{2} \mathbf{x}^{l}}{\left(\partial \mathbf{x}^{l-1}\right)^{2}}\right)_{i, j, k}=\frac{1}{m} \sigma^{\prime \prime}\left(\tilde{\mathbf{x}}_{i}^{l}\right)\left(W_{2}\right)_{i, j}^{l}\left(W_{2}\right)_{i, k}^{l}; \quad 
 \left(\frac{\partial^{2} \mathbf{x}^{l}}{\partial \mathbf{x}^{l-1} \partial W_{2}^{l}}\right)_{i, j, k k^{\prime}}=\frac{1}{m} \sigma^{\prime \prime}\left(\tilde{\mathbf{x}}_{i}^{l}\right)\left(W_{2}\right)_{i, j}^{l} \mathbf{x}_{k^{\prime}}^{l-1} \mathbb{I}_{i=k}; \\
& \left(\frac{\partial^{2} \mathbf{x}^{l}}{\partial \mathbf{x}^{l-1} \partial W_{1}^{l}}\right)_{i, j, k k^{\prime}}=\frac{1}{\sqrt{m n}} \sigma^{\prime \prime}\left(\tilde{\mathbf{x}}_{i}^{l}\right)\left(W_{2}\right)_{i, j}^{l} \mathbf{y}_{k^{\prime}} \mathbb{I}_{i=k}; \quad 
 \left(\frac{\partial^{2} \mathbf{x}^{l}}{\left(\partial W_{2}^{l}\right)^{2}}\right)_{i, j j^{\prime}, k k^{\prime}}=\frac{1}{m} \sigma^{\prime \prime}\left(\tilde{\mathbf{x}}_{i}^{l}\right) \mathbf{x}_{j^{\prime}}^{l-1} \mathbf{x}_{k^{\prime}}^{l-1} \mathbb{I}_{i=k=j}; \\
& \left(\frac{\partial^{2} \mathbf{x}^{l}}{\partial W_{2}^{l} \partial W_{1}^{l}}\right)_{i, j j^{\prime}, k k^{\prime}}=\frac{1}{\sqrt{m n}} \sigma^{\prime \prime}\left(\tilde{\mathbf{x}}_{i}^{l}\right) \mathbf{x}_{j^{\prime}}^{l-1} \mathbf{y}_{k^{\prime}} \mathbb{I}_{i=k=j}; \quad 
 \left(\frac{\partial^{2} \mathbf{x}^{l}}{\left(\partial W_{1}^{l}\right)^{2}}\right)_{i, j j^{\prime}, k k^{\prime}}=\frac{1}{n} \sigma^{\prime \prime}\left(\tilde{\mathbf{x}}_{i}^{l}\right) \mathbf{y}_{j^{\prime}} \mathbf{y}_{k^{\prime}} \mathbb{I}_{i=k=j};
\end{aligned}
\end{equation*}

\begin{equation}
\label{eq:3}
    \begin{aligned}
 \left\|\frac{\partial^{2} \mathbf{x}^{l}}{\left(\partial \mathbf{x}^{l-1}\right)^{2}}\right\|_{2,2,1}=&\sup _{\left\|\mathbf{v}_{1}\right\|=\left\|\mathbf{v}_{2}\right\|=1} \frac{1}{m} \sum_{i=1}^{m}\left|\sigma^{\prime \prime}\left(\tilde{\mathbf{x}}_{i}^{l}\right)\left(W_2^{l} \mathbf{v}_{1}\right)_{i}\left(W_2^{l} \mathbf{v}_{2}\right)_{i}\right| 
 \leq \sup _{\left\|\mathbf{v}_{1}\right\|=\left\|\mathbf{v}_{2}\right\|=1} \frac{1}{m} \beta_{\sigma} \sum_{i=1}^{m}\left|\left(W_{2}^{l} \mathbf{v}_{1}\right)_{i}\left(W_{2}^{l} \mathbf{v}_{2}\right)_{i}\right| \\
& \leq \sup _{\left\|\mathbf{v}_{1}\right\|=\left\|\mathbf{v}_{2}\right\|=1} \frac{1}{2 m} \beta_{\sigma} \sum_{i=1}^{m}\left(W_{2}^{l} \mathbf{v}_{1}\right)_{i}^{2}+\left(W_{2}^{l} \mathbf{v}_{2}\right)_{i}^{2} 
 \leq \sup  _{\left\|\mathbf{v}_{1}\right\|=\left\|\mathbf{v}_{2}\right\|=1} \frac{1}{2 m} \beta_{\sigma}\left(\left\|W_{2}^{l} \mathbf{v}_{1}\right\|^{2}+\left\|W_{2}^{l} \mathbf{v}_{2}\right\|^{2}\right) \\
&\leq \frac{1}{2 m} \beta_{\sigma}\left(\left\|W_{2}^{l}\right\|_{2}^{2}+\left\|W_{2}^{l}\right\|_{2}^{2}\right) 
 \leq \beta_{\sigma}\left(c_{20}+\frac{R_{2}}{\sqrt{m}}\right)^{2} =O(1),
\end{aligned}
\end{equation}

\begin{equation*}
    \begin{aligned}
\left\|\frac{\partial^{2} \mathbf{x}^{l}}{\partial \mathbf{x}^{l-1} \partial W_{2}^{l}}\right\|_{2,2,1} &=\sup _{\left\|\mathbf{v}_{1}\right\|=\left\|V_{2}\right\|_F=1} \frac{1}{m} \sum_{i=1}^{m}\left|\sigma^{\prime \prime}\left(\tilde{\mathbf{x}}_{i}^{l}\right)\left(W_{2}^{l} \mathbf{v}_{1}\right)_{i}\left(V_{2} \mathbf{x}^{l-1}\right)_{i}\right|  \leq \sup _{\left\|\mathbf{v}_{1}\right\|=\left\|V_{2}\right\|_F=1} \frac{1}{2 m} \beta_{\sigma}\left(\left\|W_{2}^{l} \mathbf{v}_{1}\right\|^{2}+\left\|V_{2} \mathbf{x}^{l-1}\right\|^{2}\right) \\
& \leq \frac{1}{2 m} \beta_{\sigma}\left(\left\|W_{2}^{l}\right\|^{2}+\left\|\mathbf{x}^{l-1}\right\|^{2}\right)  \leq \frac{\beta_{\sigma}}{2 m}\left(c_{20} \sqrt{m}+R_{2}\right)^{2}+\frac{\beta_{\sigma}}{2 m}\left(c_{\text {ISTA } ; \mathbf{x}}^{l-1}\right)^{2} =O(1),
\end{aligned}
\end{equation*}

\begin{equation*}
 \begin{aligned}
\left\|\frac{\partial^{2} \mathbf{x}^{l}}{\partial \mathbf{x}^{l-1} \partial W_{1}^{l}}\right\|_{2,2,1} &=\sup _{\left\|\mathbf{v}_{1}\right\|=\left\|V_{2}\right\|_F=1} \frac{1}{\sqrt{m n}} \sum_{i=1}^{m}\left|\sigma^{\prime \prime}\left(\tilde{\mathbf{x}}_{i}^{l}\right)\left(W_{2}^{l} \mathbf{v}_{1}\right)_{i}\left(V_{2} \mathbf{y}\right)_{i}\right| 
\leq \sup _{\left\|\mathbf{v}_{1}\right\|=\left\|V_{2}\right\|_F=1} \frac{1}{2 \sqrt{m n}} \beta_{\sigma}\left(\left\|W_{2}^{l} \mathbf{v}_{1}\right\|^{2}+\left\|V_{2} \mathbf{y}\right\|^{2}\right) \\
& \leq \frac{1}{2 \sqrt{m n}} \beta_{\sigma}\left(\left\|W_{2}^{l}\right\|^{2}+\|\mathbf{y}\|^{2}\right) 
 \leq \sqrt{\frac{m}{4 n}} \beta_{\sigma}\left(c_{20}+\frac{R_{2}}{\sqrt{m}}\right)^{2}+\sqrt{\frac{n}{4 m}} \beta_{\sigma} C_{y}^{2}=O(1),
\end{aligned}   
\end{equation*}

\begin{equation*}
\begin{aligned}
\left\|\frac{\partial^{2} \mathbf{x}^{l}}{\left(\partial W_{2}^{l}\right)^{2}}\right\|_{2,2,1}&=\sup _{\left\|V_{1}\right\|_F=\left\|V_{2}\right\|_F=1} \frac{1}{m} \sum_{i=1}^{m}\left|\sigma^{\prime \prime}\left(\tilde{\mathbf{x}}_i^{l}\right)\left(V_{1} \mathbf{x}^{l-1}\right)_{i}\left(V_{2} \mathbf{x}^{l-1}\right)_{i}\right|
 \leq \sup _{\left\|V_{1}\right\|_F=\left\|V_{2}\right\|_F=1} \frac{1}{2 m} \beta_{\sigma}\left(\left\|V_{1} \mathbf{x}^{l-1}\right\|^{2}+\left\|V_{2} \mathbf{x}^{l-1}\right\|^{2}\right) \\
& \leq \frac{1}{2 m} \beta_{\sigma}\left(\left\|\mathbf{x}^{l-1}\right\|^{2}+\left\|\mathbf{x}^{l-1}\right\|^{2}\right) 
 \leq \frac{1}{m} \beta_{\sigma}\left(c_{\mathrm{ISTA} ; \mathbf{x}}^{l-1}\right)^{2}=O(1),
\end{aligned}    
\end{equation*}

\begin{equation*}
\begin{aligned}
\left\|\frac{\partial^{2} \mathbf{x}^{l}}{\partial W_{2}^{l} \partial W_{1}^{l}}\right\|_{2,2,1} &=\sup _{\left\|V_{1}\right\|_F=\left\|V_{2}\right\|_F=1} \frac{1}{\sqrt{m n}} \sum_{i=1}^{m}\left|\sigma^{\prime \prime}\left(\tilde{\mathbf{x}}_{i}^{l}\right)\left(V_{1} \mathbf{x}_{i}^{l-1}\right)_{i}\left(V_{2} \mathbf{y}\right)_{i}\right| 
 \leq \sup _{\left\|V_{1}\right\|_F=\left\|V_{2}\right\|_F=1} \frac{1}{2 \sqrt{m n}} \beta_{\sigma}\left(\left\|V_{1} \mathbf{x}^{l-1}\right\|^{2}+\left\|V_{2} \mathbf{y}\right\|^{2}\right) \\
& \leq \frac{1}{2 \sqrt{m n}} \beta_{\sigma}\left(\left\|\mathbf{x}^{l-1}\right\|^{2}+\|\mathbf{y}\|^{2}\right) 
 \leq \frac{\beta_{\sigma}}{2 \sqrt{m n}}\left(c_{\text {ISTA;$\mathbf{x}$ }}^{l-1}\right)^{2}+\sqrt{\frac{n}{4 m}} \beta_{\sigma} C_y^{2}=O(1),
\end{aligned}    
\end{equation*}

\begin{equation*}
\begin{aligned}
 \left\|\frac{\partial^{2} \mathbf{x}^{l}}{\left(\partial W_{1}^{l}\right)^{2}}\right\|_{2,2,1}&=\sup _{\left\|V_{1}\right\|_F=\left\|V_{2}\right\|_F=1} \frac{1}{n} \sum_{i=1}^{m}\left|\sigma^{\prime \prime}\left(\tilde{\mathbf{x}}_{i}^{l}\right)\left(V_{1} \mathbf{y}\right)_{i}\left(V_{2} \mathbf{y}\right)_{i}\right| 
 \leq \sup _{\left\|V_{1}\right\|_F=\left\|V_{2}\right\|_F=1} \frac{1}{2 n} \beta_{\sigma}\left(\left\|V_{1} \mathbf{y}\right\|^{2}+\left\|V_{2} \mathbf{y}\right\|^{2}\right) \\
& \leq \frac{1}{2 n} \beta_{\sigma}\left(\|\mathbf{y}\|^{2}+\|\mathbf{y}\|^{2}\right)=\beta_{\sigma} C_y^{2}=O(1),
\end{aligned}    
\end{equation*}

\begin{equation}
\label{eq:4}
    \begin{aligned}
 \left\|\frac{\partial^{2} \mathbf{x}^{l}}{\partial \mathbf{x}^{l-1} \partial W^{l}}\right\|_{2,2,1}&=\left\|\left[\begin{array}{ll}
 \frac{\partial^{2} \mathbf{x}^{l}}{\partial \mathbf{x}^{l-1} \partial W_{1}^{l}} & \frac{\partial^{2} \mathbf{x}^{l}}{\partial \mathbf{x}^{l-1} \partial W_{2}^{l}}\end{array}\right]\right\|_{2,2,1} 
 \leq\left\|\frac{\partial^{2} \mathbf{x}^{l}}{\partial \mathbf{x}^{l-1} \partial W_{1}^{l}}\right\|_{2,2,1}+\left\|\frac{\partial^{2} \mathbf{x}^{l}}{\partial \mathbf{x}^{l-1} \partial W_{2}^{l}}\right\|_{2,2,1} \\
 &=O(1)+O(1)=O(1),
%\leq\left(\sqrt{\frac{m}{4 n}}+\frac{1}{2}\right) \beta_{\sigma}\left(c_{20}+\frac{R_{2}}{\sqrt{m}}\right)^{2}+\sqrt{\frac{n}{4 m}} \beta_{\sigma} C_y^{2}+\frac{\beta_{\sigma}}{2 m}\left(c_{\text {ISTA } ; \mathbf{x}}^{l-1}\right)^{2} =O(1) \\
\end{aligned}
\end{equation}
\begin{equation}
\label{eq:5}
    \begin{aligned}
 \left\|\frac{\partial^{2} \mathbf{x}^{l}}{\left(\partial W^{l}\right)^{2}}\right\|_{2,2,1}&=\left\|\left[\begin{array}{cc}\partial^{2} \mathbf{x}^{l} /\left(\partial W_{1}^{l}\right)^{2} & \partial^{2} \mathbf{x}^{l} / \partial W_{1}^{l} \partial W_{2}^{l} \\\partial^{2} \mathbf{x}^{l} / \partial W_{1}^{l} \partial W_{2}^{l} & \partial^{2} \mathbf{x}^{l} /\left(\partial W_{2}^{l}\right)^{2}\end{array}\right]\right\|_{2,2,1} 
 \leq\left\|\frac{\partial^{2} \mathbf{x}^{l}}{\left(\partial W_{1}^{l}\right)^{2}}\right\|_{2,2,1}+2\left\|\frac{\partial^{2} \mathbf{x}^{l}}{\partial W_{1}^{l} \partial W_{2}^{l}}\right\|_{2,2,1}+\left\|\frac{\partial^{2} \mathbf{x}^{l}}{\left(\partial W_{2}^{l}\right)^{2}}\right\|_{2,2,1} \\
 &=O(1)+O(1)+O(1)=O(1).
%\leq\left(\sqrt{\frac{n}{4 m}}+1\right) \beta_{\sigma} C_y^{2}+\left(\frac{1}{2 \sqrt{m n}}+\frac{1}{m}\right) \beta_{\sigma}\left(c_{ISTA ; \mathbf{x}}^{l-1}\right)^{2} = O(1)
\end{aligned}
\end{equation}
Therefore, by using \eqref{eq:2}, \eqref{eq:3}, \eqref{eq:4}, and \eqref{eq:5}, we get $\mathcal{Q}_{2,2,1}\left(f_{s}\right)=O(1)$, for all $s \in [m]$.
\subsection{Bound on $Q_{2,2,1}$ For ADMM-CSNet}
\label{Q_2_2_1_For_ADMMCSNet}
Consider an L-layered ADMM-CSNet as
\begin{equation}
\label{eq:6}
    \begin{aligned}
    \mathbf{f}&=\frac{1}{\sqrt{m}} \mathbf{z}^L ;\\ \mathbf{z}^l &= \sigma\left(\tilde{\mathbf{z}}^{l}\right) =\sigma\left( \mathbf{x}^l+\mathbf{u}^{l-1} \right),\\
    \mathbf{x}^{l} &= \frac{1}{\sqrt{n}} W_1^l\mathbf{y} + \frac{1}{\sqrt{m}} W_2^l\left(\mathbf{z}^{l-1}-\mathbf{u}^{l-1}\right), \\
        \mathbf{u}^l &= \mathbf{u}^{l-1}+\left(\mathbf{x}^l-\mathbf{z}^l\right).
\end{aligned}
\end{equation}
where f is the output of the network. Now the first derivatives of  $\mathbf{z}^{l}$ are
\begin{equation*}
   \begin{aligned}
&\left(\frac{\partial \mathbf{z}^{l}}{\partial \mathbf{z}^{l-1}}\right)_{i, j}=\left(\frac{\partial \mathbf{z}^{l}}{\partial \mathbf{z}^{l-1}}\right)_{i, j}+\left(\frac{\partial \mathbf{z}^{l}}{\partial \mathbf{u}^{l-1}} \frac{\partial \mathbf{u}^{l-1}}{\partial \mathbf{z}^{l-1}}\right)_{i, j}=\sigma^{\prime}\left(\tilde{\mathbf{z}}_{i}^{l}\right)\left(\frac{2}{\sqrt{m}} W_{2}^{l}-I\right)_{i, j}; \\
&\left(\frac{\partial \mathbf{z}^{l}}{\partial W_{1}^{l}}\right)_{i, j j^{\prime}}=\frac{1}{\sqrt{n}} \sigma^{\prime}\left(\tilde{\mathbf{z}}_{i}^{l}\right) \mathbf{y}_{j^{\prime}} \mathbb{I}_{i=j} ;
\left(\frac{\partial \mathbf{z}^{l}}{\partial W_{2}^{l}}\right)_{i, j j^{\prime}}=\frac{1}{\sqrt{m}} \sigma^{\prime}\left(\tilde{\mathbf{z}}_{i}^{l}\right) (\mathbf{z}^{l-1}-\mathbf{u}^{l-1})_{j^{\prime}} \mathbb{I}_{i=j}.
\end{aligned}  
 \end{equation*}
%By the definition of spectral norm, $\|A\|_2=\sup _{\|\mathbf{v}\|=1}\|A \mathbf{v}\|_2$, we have
%$
%\begin{aligned}
%\left\|\frac{\partial \mathbf{z}^{l}}{\partial \mathbf{z}^{l-1}}\right\|_2^{2} &=\sup _{\|\mathbf{v}\|=1} \sum_{i}\left(\sigma^{\prime}\left(\tilde{\mathbf{z}}_{i}^{l}\right)\left(\frac{2}{\sqrt{m}} W_{2}^{l}- I\right)_{i, j} \mathbf{v}_{j}\right)^{2} 
%=\sup _{\|\mathbf{v}\|=1}\left\|\frac{2}{\sqrt{m}} \Sigma^{\prime l} W_{2}^{l} \mathbf{v}- \Sigma^{\prime l} \mathbf{v}\right\|^{2} 
 %\leq \frac{(2)^{2}}{m}\left\|\Sigma^{\prime l}\right\|^{2}\left\|W_{2}^{l}\right\|^{2}+\left\|\Sigma^{\prime l}\right\|^{2} \\
%& \leq \frac{4}{m} L_{\sigma}^{2}\left(c_{20} \sqrt{m}+R_{2}\right)^{2}+ L_{\sigma}^{2} =O(1)
%\end{aligned}
%$
Now, we have
\begin{equation*}
    \begin{aligned}
\left\|\frac{\partial \mathbf{z}^{l}}{\partial W_{1}^{l}}\right\|_2^{2} &=\sup _{\|V\|_{F}=1} \frac{1}{n} \sum_{i=1}^{m}\left(\sum_{j, j^{\prime}} \sigma^{\prime}\left(\tilde{\mathbf{z}}_{i}^{l}\right) \mathbf{y}_{j^{\prime}} \mathbb{I}_{i=j} V_{j j^{\prime}}\right)^{2} 
=\sup _{\|V\|_F=1} \frac{1}{n}\left\|\Sigma^{\prime l} V \mathbf{y}\right\|^{2} 
 \leq \frac{1}{n}\left\|\Sigma^{\prime l}\right\|^{2}\|\mathbf{y}\|^{2} 
 \leq L_{\sigma}^{2} C_y^{2}=O(1),
\end{aligned}
\end{equation*}
where $\Sigma^{\prime l}$ is a diagonal matrix with the diagonal entry $(\Sigma^{\prime l})_{ii}=\sigma^{\prime}\left(\tilde{\mathbf{z}}_{i}^{l}\right)$.
\begin{equation*}
    \begin{aligned}
\left\|\frac{\partial \mathbf{z}^{l}}{\partial W_{2}^{l}}\right\|_2^{2} &=\sup _{\|V\|_{F}=1} \frac{1}{m} \sum_{i=1}^{m}\left(\sum_{j, j^{\prime}} \sigma^{\prime}\left(\tilde{\mathbf{z}}_{i}^{l}\right) (\mathbf{z}^{l-1}-\mathbf{u}^{l-1})_{j^{\prime}} \mathbb{I}_{i=j} V_{j j^{\prime}}\right)^{2} 
=\sup _{\|V\|_F=1} \frac{1}{m}\left\|\Sigma^{\prime  l} V (\mathbf{z}^{l-1}-\mathbf{u}^{l-1})\right\|^{2} \\
& \leq \frac{1}{m}\left\|\Sigma^{\prime l}\right\|^{2}\left\|\mathbf{z}^{l-1}\right\|^{2}+ \frac{1}{m}\left\|\Sigma^{\prime l}\right\|^{2}\left\|\mathbf{u}^{l-1}\right\|^{2} 
 \leq \frac{1}{m} L_{\sigma}^{2} \left( \left(c_{\mathrm{ADMM} ; \mathbf{z}}^{l-1}\right)^{2} + \left(c_{\mathrm{ADMM} ; \mathbf{u}}^{l-1}\right)^{2} \right)=O(1).
 \end{aligned}
\end{equation*}
From lemma (4) we used $\left(c_{\mathrm{ADMM} ; \mathbf{z}}^{l-1}\right)=O(\sqrt{m})$ and $\left(c_{\mathrm{ADMM} ; \mathbf{u}}^{l-1}\right)=O(\sqrt{m})$. Therefore
 \begin{equation}
     \label{eq:7}
      \begin{aligned}
\left\|\frac{\partial \mathbf{z}^{l}}{\partial W^{l}}\right\| &=\left\|\left[\frac{\partial \mathbf{z}^{l}}{\partial W_{1}^{l}} \quad \frac{\partial \mathbf{z}^{l}}{\partial W_{2}^{l}}\right]\right\| 
 \leq\left\|\frac{\partial \mathbf{z}^{l}}{\partial W_{1}^{l}}\right\|+\left\|\frac{\partial \mathbf{z}^{l}}{\partial W_{2}^{l}}\right\| 
 =O(1)+O(1)=O(1).
% \leq L_{\sigma}^{2} C_y^{2}+\frac{1}{m} L_{\sigma}^{2} \left( \left(c_{\mathrm{ADMM} ; \mathbf{z}}^{l-1}\right)^{2} + \left(c_{\mathrm{ADMM} ; \mathbf{u}}^{l-1}\right)^{2} \right) =O(1)
\end{aligned}
 \end{equation}
The second derivatives of the vector-valued layer function $\mathbf{z}^{l}$, which are order 3 tensors, have the following expressions:
\begin{equation}
    \begin{aligned}
&\left(\frac{\partial^{2} \mathbf{z}^{l}}{\left(\partial \mathbf{z}^{l-1}\right)^{2}}\right)_{i, j, k}=\sigma^{\prime \prime}\left(\tilde{\mathbf{z}}_{i}^{l}\right)\left(\frac{2}{\sqrt{m}} W_{2}^{l}- I\right)_{i, j}\left(\frac{2}{\sqrt{m}} W_{2}^{l}- I\right)_{i, k}; \\
&\left(\frac{\partial^{2} \mathbf{z}^{l}}{\partial \mathbf{z}^{l-1} \partial W_{2}^{l}}\right)_{i, j, k k^{\prime}}=\frac{1}{\sqrt{m}} \sigma^{\prime \prime}\left(\tilde{\mathbf{z}}_{i}^{l}\right)\left(\frac{2}{\sqrt{m}} W_{2}^{l}- I\right)_{i j} (\mathbf{z}^{l-1}-\mathbf{u}^{l-1})_{k^{\prime}} \mathbb{I}_{i=k} +\frac{2}{\sqrt{m}} \sigma^{\prime}\left( \tilde{\mathbf{z}}_{i}^{l} \right) \mathbb{I}_{i=k} \mathbb{I}_{j=k^{\prime}};\\
&\left(\frac{\partial^{2} \mathbf{z}^{l}}{\partial \mathbf{z}^{l-1} \partial W_{1}^{l}}\right)_{i, j, k k^{\prime}}=\frac{1}{\sqrt{m}} \sigma^{\prime \prime}\left(\tilde{\mathbf{z}}_{i}^{l}\right)\left(\frac{2}{\sqrt{m}} W_{2}^{l}- I\right)_{i j} \mathbf{y}_{k^{\prime}} \mathbb{I}_{i=k};\\
&\left(\frac{\partial^{2} \mathbf{z}^{l}}{\left(\partial W_{2}^{l}\right)^{2}}\right)_{i, j j^{\prime}, k k^{\prime}}=\frac{1}{m} \sigma^{\prime \prime}\left(\tilde{\mathbf{z}}_{i}^{l}\right) (\mathbf{z}^{l-1}-\mathbf{u}^{l-1})_{j^{\prime}} (\mathbf{z}^{l-1}-\mathbf{u}^{l-1})_{k^{\prime}} \mathbb{I}_{i=k=j}; \\
&\left(\frac{\partial^{2} \mathbf{z}^{l}}{\partial W_{2}^{l} \partial W_{1}^{l}}\right)_{i, j j^{\prime}, k k^{\prime}}=\frac{1}{\sqrt{mn}} \sigma^{\prime \prime}\left(\tilde{\mathbf{z}}_{i}^{l}\right) (\mathbf{z}^{l-1}-\mathbf{u}^{l-1})_{j^{\prime}} \mathbf{y}_{k^{\prime}} \mathbb{I}_{i=k=j}; \\
&\left(\frac{\partial^{2} \mathbf{z}^{l}}{\left(\partial W_{1}^{l}\right)^{2}}\right)_{i, j j^{\prime}, k k^{\prime}}=\frac{1}{n} \sigma^{\prime \prime}\left(\tilde{\mathbf{z}}_{i}^{l}\right) \mathbf{y}_{j^{\prime}} \mathbf{y}_{k^{\prime}} \mathbb{I}_{i=k=j};
\end{aligned}
\end{equation}

\begin{equation}
    \label{eq:8}
    \begin{aligned}
    \left\|\frac{\partial^{2} \mathbf{z}^{l}}{\left(\partial \mathbf{z}^{l-1}\right)^{2}}\right\|_{2,2,1} &=\sup _{\left\|\mathbf{v}_{1}\right\|=\left\|\mathbf{v}_{2}\right\|=1} \sum_{i=1}^{m}\left|\sigma^{\prime \prime}\left(\tilde{\mathbf{z}}_{i}^{l}\right)\left(\left(\frac{2}{\sqrt{m}} W_{2}^{l}- I\right) \mathbf{v}_{1}\right)_{i}\left(\left(\frac{2}{\sqrt{m}} W_{2}^{l}- I\right) \mathbf{v}_{2}\right)_{i}\right| \\
&\leq \sup _{\left\|\mathbf{v}_{1}\right\|=\left\|\mathbf{v}_{2}\right\|=1} \beta_{\sigma} \sum_{i=1}^{m}\left|\left(\left(\frac{2}{\sqrt{m}} W_{2}^{l}- I\right) \mathbf{v}_{1}\right)_{i}\left(\left(\frac{2}{\sqrt{m}} W_{2}^{l}- I\right) \mathbf{v}_{2}\right)_{i}\right| \\
&\leq \sup _{\left\|\mathbf{v}_{1}\right\|=\left\|\mathbf{v}_{2}\right\|=1} \frac{1}{2} \beta_{\sigma} \sum_{i=1}^{m}\left(\left(\frac{2}{\sqrt{m}} W_{2}^{l}- I\right) \mathbf{v}_{1}\right)_{i}^{2}+\left(\left(\frac{2}{\sqrt{m}} W_{2}^{l}- I\right)^{2}\right)_{i} \\
&\leq \sup _{\left\|\mathbf{v}_{1}\right\|=\left\|\mathbf{v}_{2}\right\|=1} \frac{1}{2} \beta_{\sigma}\left(\left\|\frac{2}{\sqrt{m}} W_{2}^{l} \mathbf{v}_{1}- \mathbf{v}_{1}\right\|^{2}+\left\|\frac{2}{\sqrt{m}} W_{2}^{l} \mathbf{v}_{2}- \mathbf{v}_{2}\right\|^{2}\right) \\
&\leq \beta_{\sigma}\left(\left\|\frac{2}{\sqrt{m}} W_{2}^{l}\right\|^{2}+1^{2}\right) \leq \beta_{\sigma} +4 \beta_{\sigma} \frac{\left(c_{20} \sqrt{m}+R_{2}\right)^{2}}{m} =O(1),
    \end{aligned}
\end{equation}

\begin{equation*}
    \begin{aligned}
\left\|\frac{\partial^{2} \mathbf{z}^{l}}{\partial \mathbf{z}^{l-1} \partial W_{2}^{l}}\right\|_{2,2,1} &=\sup _{\left\|\mathbf{v}_{1}\right\|=\left\|V_{2}\right\|_F=1}  \sum_{i=1}^{m}\left| \frac{1}{m} \sigma^{\prime \prime}\left(\tilde{\mathbf{z}}_{i}^{l}\right)\left(\left(2 W_{2}^{l}- \sqrt{m} I\right) \mathbf{v}_{1}\right)_{i}\left(V_{2} (\mathbf{z}^{l-1}-\mathbf{u}^{l-1})\right)_{i} +\frac{2}{\sqrt{m}} \sigma^{\prime}\left( \tilde{\mathbf{z}}_{i}^{l} \right) \left(V_{2} \mathbf{v}_1 \right)_i \right| \\
& \leq \sup _{\left\|\mathbf{v}_{1}\right\|=\left\|V_{2}\right\|_F=1} \frac{1}{2 m} \beta_{\sigma}\left(\left\|2 W_{2}^{l} \mathbf{v}_{1}- \sqrt{m} \mathbf{v}_{1}\right\|^{2}+\left\|V_{2} \mathbf{z}^{l-1}-V_{2} \mathbf{u}^{l-1}\right\|^{2}\right) + \frac{2}{\sqrt{m}} \left\|\Sigma^{\prime l}\right\| \left\| \mathbf{v}_1\right\| \left\| V_2\right\|\\
& \leq \frac{1}{2 m} \beta_{\sigma}\left(\left\|2 W_{2}^{l}\right\|^{2}+m+\left\|\mathbf{z}^{l-1}\right\|^{2}+ \left\|\mathbf{u}^{l-1}\right\|^{2}\right) +\frac{2}{\sqrt{m}} L_{\sigma} \\
& \leq \frac{\beta_{\sigma}}{2 m}\left((2)^{2}\left(c_{20} \sqrt{m}+R_{2}\right)^{2}+m+\left(c_{\text {ADMM; $\mathbf{z}$ }}^{l-1}\right)^{2} + \left(c_{\text {ADMM; $\mathbf{u}$ }}^{l-1}\right)^{2}\right) + \frac{2}{\sqrt{m}} L_{\sigma} \\
& = O(1) +O(1/\sqrt{m})=O(1),
\end{aligned}
\end{equation*}
\begin{equation*}
    \begin{aligned}
\left\|\frac{\partial^{2} \mathbf{z}^{l}}{\partial \mathbf{z}^{l-1} \partial W_{1}^{l}}\right\|_{2,2,1}&=\sup _{\left\|\mathbf{v}_{1}\right\|=\left\|V_{2}\right\|_F=1} \frac{1}{m} \sum_{i=1}^{m}\left|\sigma^{\prime \prime}\left(\tilde{\mathbf{z}}_{i}^{l}\right)\left(\left(2 W_{2}^{l}- \sqrt{m}I\right) \mathbf{v}_{1}\right)_{i}\left(V_{2} \mathbf{y}\right)_{i}\right|\\
&\leq \sup _{\left\|\mathbf{v}_{1}\right\|=\left\|V_{2}\right\|_F=1} \frac{1}{2 m} \beta_{\sigma}\left(\left\|2 W_{2}^{l} \mathbf{v}_{1}-\sqrt{m} \mathbf{v}_{1}\right\|^{2}+\left\|V_{2} \mathbf{y}\right\|^{2}\right) 
\leq \frac{1}{2 m} \beta_{\sigma}\left(\left\|2 W_{2}^{l}\right\|^{2}+m+\|\mathbf{y}\|^{2}\right) \\
&\leq \frac{\beta_{\sigma}}{2 m}\left((2)^{2}\left(c_{20} \sqrt{m}+R_{2}\right)^{2}+m+m C_y^{2}\right) =O(1),
\end{aligned}
\end{equation*}
\begin{equation}
    \label{eq:9}
    \begin{aligned}
        \left\|\frac{\partial^{2} \mathbf{z}^{l}}{\partial \mathbf{z}^{l-1} \partial W^{l}}\right\|_{2,2,1}&=\left\|\left[\frac{\partial^{2} \mathbf{z}^{l}}{\partial \mathbf{z}^{l-1} \partial W_{1}^{l}} \quad \frac{\partial^{2} \mathbf{z}^{l}}{\partial \mathbf{z}^{l-1} \partial W_{2}^{l}}\right]\right\|_{2,2,1} 
 \leq\left\|\frac{\partial^{2} \mathbf{z}^{l}}{\partial \mathbf{z}^{l-1} \partial W_{1}^{l}}\right\|_{2,2,1}+\left\|\frac{\partial^{2} \mathbf{z}^{l}}{\partial \mathbf{z}^{l-1} \partial W_{2}^{l}}\right\|_{2,2,1} =O(1),
%& \leq \frac{\beta_{\sigma}}{2 m}\left(m C_y^{2}+\left(c_{\mathrm{ADMM} ; \mathbf{z}}^{l-1}\right)^{2}+\left(c_{\mathrm{ADMM} ; \mathbf{u}}^{l-1}\right)^{2}\right) +\frac{\beta_{\sigma}}{m}\left((2)^{2}\left(c_{20} \sqrt{m}+R_{2}\right)^{2}+m\right) 
    \end{aligned}
\end{equation}
\begin{equation*}
    \begin{aligned}
 \left\|\frac{\partial^{2} \mathbf{z}^{l}}{\left(\partial W_{2}^{l}\right)^{2}}\right\|_{2,2,1}=&\sup _{\left\|V_{1}\right\|_F=\left\|V_{2}\right\|_F=1} \frac{1}{m} \sum_{i=1}^{m}\left|\sigma^{\prime \prime}\left(\tilde{\mathbf{z}}_{i}^{l}\right)\left(V_{1} (\mathbf{z}^{l-1}-\mathbf{u}^{l-1})\right)_{i}\left(V_{2} (\mathbf{z}^{l-1}-\mathbf{u}^{l-1})\right)_{i}\right| \\
& \leq \sup _{\left\|V_{1}\right\|_F=\left\|V_{2}\right\|_F=1} \frac{1}{2 m} \beta_{\sigma} \sum_{i=1}^{m}\left(\left\|V_{1} (\mathbf{z}^{l-1}-\mathbf{u}^{l-1})\right\|^{2}+\left\|V_{2} (\mathbf{z}^{l-1}-\mathbf{u}^{l-1})\right\|^{2}\right) \\
& \leq \frac{1}{2 m} \beta_{\sigma}\left(\left\|\mathbf{z}^{l-1}\right\|^{2}+\left\|\mathbf{u}^{l-1}\right\|^{2}+\left\|\mathbf{z}^{l-1}\right\|^{2}+\left\|\mathbf{u}^{l-1}\right\|^{2}\right) 
\leq \frac{1}{m} \beta_{\sigma}\left( \left(c_{\mathrm{ADMM} ; \mathbf{z}}^{l-1}\right)^{2} + \left(c_{\mathrm{ADMM} ; \mathbf{u}}^{l-1}\right)^{2}\right) =O(1),
\end{aligned}
\end{equation*}

\begin{equation*}
    \begin{aligned}
\left\|\frac{\partial^{2} \mathbf{z}^{l}}{\partial W_{1}^{l} \partial W_{2}^{l}}\right\|_{2,2,1}&=\sup _{\left\|V_{1}\right\|_F=\left\|V_{2}\right\|_F=1} \frac{1}{\sqrt{mn}} \sum_{i=1}^{m}\left|\sigma^{\prime \prime}\left(\tilde{\mathbf{z}}_{i}^{l}\right)\left(V_{1} (\mathbf{z}^{l-1}-\mathbf{u}^{l-1})\right)_{i}\left(V_{2} \mathbf{y}\right)_{i}\right| \\
& \leq \sup _{\left\|V_{1}\right\|_F=\left\|V_{2}\right\|_F=1} \frac{1}{2 \sqrt{mn}} \beta_{\sigma} \sum_{i=1}^{m}\left(\left\|V_{1} (\mathbf{z}^{l-1}-\mathbf{u}^{l-1})\right\|^{2}+\left\|V_{2} \mathbf{y}\right\|^{2}\right) \\
& \leq \frac{1}{2 \sqrt{mn}} \beta_{\sigma}\left(\left\|\mathbf{z}^{l-1}\right\|^{2}+\left\|\mathbf{u}^{l-1}\right\|^{2}+\|\mathbf{y}\|^{2}\right) 
\leq \frac{1}{2 \sqrt{mn}} \beta_{\sigma}\left(n C_y^{2}+\left(c_{\text {ADMM } ; \mathbf{z}}^{l-1}\right)^{2}+\left(c_{\text {ADMM } ; \mathbf{u}}^{l-1}\right)^{2}\right)=O(1),
\end{aligned}
\end{equation*}

\begin{equation*}
    \begin{aligned}
\left\|\frac{\partial^{2} \mathbf{z}^{l}}{\left(\partial W_{1}^{l}\right)^{2}}\right\|_{2,2,1}&=\sup _{\left\|V_{1}\right\|_F=\left\|V_{2}\right\|_F=1} \frac{1}{n} \sum_{i=1}^{m}\left|\sigma^{\prime \prime}\left(\tilde{\mathbf{z}}_{i}^{l}\right)\left(V_{1} \mathbf{y}\right)_{i}\left(V_{2} \mathbf{y}\right)_{i}\right| 
 \leq \sup _{\left\|V_{1}\right\|_F=\left\|V_{2}\right\|_F=1} \frac{1}{2 n} \beta_{\sigma} \sum_{i=1}^{m}\left(\left\|V_{1} \mathbf{y}\right\|^{2}+\left\|V_{2} \mathbf{y}\right\|^{2}\right) \\
& \leq \frac{1}{2 n} \beta_{\sigma}\left(\|\mathbf{y}\|^{2}+\|\mathbf{y}\|^{2}\right) \leq \beta_{\sigma} C_y^{2}=O(1),
\end{aligned}
\end{equation*}
\begin{equation}
    \label{eq:10}
    \begin{aligned}
\left\|\frac{\partial^{2} \mathbf{z}^{l}}{\left(\partial W^{l}\right)^{2}}\right\|_{2,2,1} &=\left\|\left[\begin{array}{cc}
\partial^{2} \mathbf{z}^{l} /\left(\partial W_{1}^{l}\right)^{2} & \partial^{2} \mathbf{z}^{l} / \partial W_{1}^{l} \partial W_{2}^{l} \\
\partial^{2} \mathbf{z}^{l} / \partial W_{1}^{l} \partial W_{2}^{l} & \partial^{2} \mathbf{z}^{l} /\left(\partial W_{2}^{l}\right)^{2}
\end{array}\right]\right\|_{2,2,1} 
\leq\left\|\frac{\partial^{2} \mathbf{z}^{l}}{\left(\partial W_{1}^{l}\right)^{2}}\right\|_{2,2,1}+2\left\|\frac{\partial^{2} \mathbf{z}^{l}}{\partial W_{1}^{l} \partial W_{2}^{l}}\right\|_{2,2,1}+\left\|\frac{\partial^{2} \mathbf{z}^{l}}{\left(\partial W_{2}^{l}\right)^{2}}\right\|_{2,2,1} \\
%& \leq \beta_{\sigma} C_y^{2}+ \frac{1}{ \sqrt{mn}} \beta_{\sigma}\left(n C_y^{2}+\left(c_{\text {ADMM } ; \mathbf{z}}^{l-1}\right)^{2}+\left(c_{\text {ADMM } ; \mathbf{u}}^{l-1}\right)^{2}\right) + \frac{1}{m} \beta_{\sigma}\left( \left(c_{\mathrm{ADMM} ; \mathbf{z}}^{l-1}\right)^{2} + \left(c_{\mathrm{ADMM} ; \mathbf{u}}^{l-1}\right)^{2}\right)\\
&=O(1).
%\leq 2 \beta_{\sigma} C_y^{2}+\frac{2 \beta_{\sigma}}{ m} \left( \left(c_{\mathrm{ADMM} ; \mathbf{z}}^{l-1}\right)^{2} + \left(c_{\mathrm{ADMM} ; \mathbf{z}}^{l-1}\right)^{2} \right) =O(1)
\end{aligned}
\end{equation}
Therefore, from \eqref{eq:7}, \eqref{eq:8}, \eqref{eq:9}, and \eqref{eq:10}, we get that $\mathcal{Q}_{2,2,1}(f_s)=O(1)$, for all $s \in [m].$
\subsection{Bound on $Q_{\infty}$ For LISTA Network}
\label{Q_Infinity_For_LISTA}
Let $\mathbf{b}_s^l = \frac{\partial f_s}{\partial \mathbf{x}^{l}}$, then $
    \mathcal{Q}_{\infty}\left(f_s\right)= \max _{1 \leq l \leq L}\left\{\left\|\mathbf{b}_s^l\right\|_{\infty}\right\}$.
We now compute bound on $\left\|\mathbf{b}_s^l\right\|_{\infty}$.
From triangle inequality, we can write
\begin{equation}
\label{eq:inf}
    \begin{aligned}
\left\|\mathbf{b}_{s}^{l}\right\|_{\infty} & \leq\left\|\mathbf{b}_{s,0}^{l}\right\|_{\infty}+\left\|\mathbf{b}_{s}^{l}-\mathbf{b}_{s,0}^{l}\right\|_{\infty} \leq\left\|\mathbf{b}_{s,0}^{l}\right\|_{\infty}+\left\|\mathbf{b}_{s}^{l}-\mathbf{b}_{s,0}^{l}\right\|.
\end{aligned}
\end{equation}
where $\mathbf{b}_{s,0}^l$ is $\mathbf{b}_{s}^l$ at initialization.
Therefore, one can obtain the bound on $\left\|\mathbf{b}_{s}^{l}\right\|_{\infty}$ by computing the bounds on $\left\|\mathbf{b}_{s,0}^{l}\right\|_{\infty}$ and $\left\|\mathbf{b}_{s}^{l}-\mathbf{b}_{s,0}^{l}\right\|$, which are provided in
Lemma \ref{lm:8} and Lemma \ref{lm:9}, respectively.
Moreover, in order to compute the bound on $\left\|\mathbf{b}_{s,0}^{l}\right\|_{\infty}$, we require several lemmas which are stated below. In specific, Lemma \ref{lm:6} and Lemma \ref{lm:7} provide the bound on each component of the hidden layer's output at initialization and the bound on $l_2$-norm of $\mathbf{b}_s^l,\ l\in[L]$, respectively. 
% To prove $\mathcal{Q}_{\infty}(f_s)=O(1/\sqrt{m})$ for all $s \in [m]$, we require several lemmas which are stated below with proofs.
\begin{Lemma}
\label{lm:6}
     For any $l\in [L]$ and $i\in[m]$, we have $\left|\mathbf{x}_{i}^{l}\right| \leq \ln (m)+|\sigma(0)|$ at initialization
     with probability at least $1-2 e^{-c_{\mathbf{x}}^{l} l n^{2}(m)}$ for some constant $c_{\mathbf{x}}^{l}>0$.
\end{Lemma}

\begin{proof}
From \eqref{eq:1},
\begin{equation*}
 \begin{aligned}
\left|\mathbf{x}_{i}^{l}\right| =\left|\sigma\left(\frac{1}{\sqrt{m}} \sum_{k=1}^{m}\left(W_{2}^{l}\right)_{i k} \mathbf{x}_{k}^{l-1}+\frac{1}{\sqrt{n}}\sum_{k=1}^n\left(W_{1}^{l}\right)_{i k} \mathbf{y}_{k}\right)\right| \leq\left|\frac{L_{\sigma}}{\sqrt{m}} \sum_{k=1}^{m}\left(W_{2}^{l}\right)_{i k} \mathbf{x}_{k}^{l-1}+\frac{L_{\sigma}}{\sqrt{n}} \sum_{k=1}^{n}\left(W_{1}^{l}\right)_{i k} \mathbf{y}_{k}\right|+|\sigma(0)|.
\end{aligned}   
\end{equation*}
As $\left(W_{1}^{l}\right)_{ik} \sim$ $\mathcal{N}(0,1)$ and $\left(W_{2}^{l}\right)_{ik} \sim \mathcal{N}(0,1)$, so that $\sum_{k=1}^{n}\left(W_{1}^{l}\right)_{ik} \mathbf{y}_{k} \sim \mathcal{N}\left(0,\|\mathbf{y}\|^{2}\right)$ and $\sum_{k=1}^{m}\left(W_{2}^{l}\right)_{ik} \mathbf{x}_{k}^{l-1} \sim \mathcal{N}\left(0,\left\|\mathbf{x}^{l-1}\right\|^{2}\right)$. \
In addition, since $\left(W_{1}^{l}\right)_{ik}$ and $\left(W_{2}^{l}\right)_{ik}$ are independent, $\sum_{k=1}^{n}\left(W_{1}^{l}\right)_{ik} \mathbf{y}_{k}+$ $\sum_{k=1}^{m}\left(W_{2}^{l}\right)_{ik} \mathbf{x}_{k}^{l-1} \sim \mathcal{N}\left(0,\|\mathbf{y}\|^{2}+\left\|\mathbf{x}^{l-1}\right\|^{2}\right)$. Using the concentration inequality of a Gaussian random variable, we obtain
\begin{equation*}
   \begin{aligned}
\operatorname{Pr}\left[\left|\mathbf{x}_{i}^{l}\right| \geq \ln (m)+|\sigma(0)|\right] & \leq \operatorname{Pr}\left[\left|\frac{L_{\sigma}}{\sqrt{m}} \sum_{k=1}^{m}\left(W_{2}^{l}\right)_{ik} \mathbf{x}_{k}^{l-1}+\frac{L_{\sigma}}{\sqrt{n}} \sum_{k=1}^{n}\left(W_{1}^{l}\right)_{ik} \mathbf{y}_{k}\right| \geq \ln (m)\right] 
 \leq 2 e^{-\frac{m l n^{2}(m)}{2 L_{\sigma}^{2}\left(\|\mathbf{y}\|^{2}+\left\|\mathbf{x}^{l-1}\right\|^{2}\right)}}.
\end{aligned} 
\end{equation*}
This implies,
\begin{equation}    \operatorname{Pr}\left[\left|\mathbf{x}_{i}^{l}\right| \leq \ln (m)+|\sigma(0)|\right] \geq 1-2 e^{-\frac{m \ln^{2}(m)}{2 L_{\sigma}^{2}\left(\|\mathbf{y}\|^{2}+\left\|\mathbf{x}^{l-1}\right\|^{2}\right)}} = 1-2 e^{-c_{\mathbf{x}}^{l} \ln^{2}(m)},\ \forall l\in[L],
\end{equation}
where $c_{\mathbf{x}}^{l}=\frac{m }{2 L_{\sigma}^{2}\left(\|\mathbf{y}\|^{2}+\left\|\mathbf{x}^{l-1}\right\|^{2}\right)}>0.$
\end{proof}
\begin{Lemma}
    \label{lm:7}
Consider an $L$-layer LISTA network with $\left(W_{10}^{l}\right)_{i, j} \sim \mathcal{N}(0,1)$ and $\left(W_{20}^{l}\right)_{i, j} \sim \mathcal{N}(0,1)$, $\forall l\in[L]$, then, for any $\mathbf{W}_{1}$ and $\mathbf{W}_{2}$ such that $\left\|\mathbf{W}_{1}-\mathbf{W}_{10}\right\| \leq R_{1}$ and $\left\|\mathbf{W}_{2}-\mathbf{W}_{20}\right\| \leq R_{2}$, we have, 
% at all hidden layers, i.e., $l\in [L]$,
\begin{equation}
\label{eqlm71}
    \left\|\mathbf{b}_{s}^{l}\right\| \leq L_{\sigma}^{L-l}\left(c_{20}+R_{2} / \sqrt{m}\right)^{L-l}, \ l\in[L].
\end{equation}
From this at initialization, i.e., for $R_2=0$, we get
\begin{equation}
\label{eqlm72}
    \left\|\mathbf{b}_{s, 0}^{l}\right\| \leq L_{\sigma}^{L-l} c_{20}^{L-l}.
\end{equation}
\end{Lemma}
\begin{proof}
We prove this lemma by using induction on $l$.
Initially, for $l=L$, we have
\begin{equation*}
    \left\|\mathbf{b}_{s}^{L}\right\|=\left\|\frac{\partial f_s}{\partial \mathbf{x}^{L}}\right\|=(1 / \sqrt{m })\left\|\mathbf{v}_{s}\right\|=1 / \sqrt{m }<1.
\end{equation*}
That is, the inequality in \eqref{eqlm71} holds true for $l=L$.
Assume that at $l^{th}$ layer the inequality holds, i.e., $\left\|\mathbf{b}_{s}^{l}\right\| \leq L_{\sigma}^{L-l}\left(c_{0}+R_{2} / \sqrt{m}\right)^{L-l}$, then below we prove that \eqref{eqlm71} holds true even for the $(l-1)^{th}$  layer:
\begin{equation*}
    \begin{aligned}
\left\|\mathbf{b}_{s}^{l-1}\right\| &=\left\|\frac{\partial f_s}{\partial \mathbf{x}^{l-1}}\right\|=\left\|\frac{\partial \mathbf{x}^l}{\partial \mathbf{x}^{l-1}}\frac{\partial f_s}{\partial \mathbf{x}^{l}}\right\|=\left\|\frac{1}{\sqrt{m}}\left(W_{2}^{l}\right)^{T} \Sigma^{\prime l} \mathbf{b}_{s}^{l}\right\| 
 \leq \frac{1}{\sqrt{m}}\left\|W_{2}^{l}\right\|\left\|\Sigma^{\prime l}\right\|\left\|\mathbf{b}_{s}^{l}\right\| \\ &\leq\left(c_{20}+R_{2} / \sqrt{m}\right) L_{\sigma}\left\|\mathbf{b}_{s}^{l}\right\| 
\leq\left(c_{20}+R_{2} / \sqrt{m}\right)^{L-l+1} L_{\sigma}^{L-l+1}.
\end{aligned}
\end{equation*}
% i.e. For $l-1$, \eqref{eqlm71} is true.
So, from the above analysis, we claim that the inequality in \eqref{eqlm71} holds true for any $l\in [L]$.
Now, at initialization, i.e., substituting $R_2=0$ in \eqref{eqlm71} directly leads to \eqref{eqlm72}.
\end{proof}
As mentioned earlier, we now use Lemma \ref{lm:6} and Lemma \ref{lm:7} to provide bound on $\left\|\mathbf{b}_{s, 0}^{l}\right\|_{\infty}$.
\begin{Lemma}
\label{lm:8}
At initialization, the $\infty$-norm of $\mathbf{b}_s^l$ is in  $\Tilde{O}(1/\sqrt{m})$ with probability $1-m e^{-c_{b s}^{l} \ln ^{2}(m)}$ for some constant $c_{b s}^{l}>0,$
 i.e., 
 \begin{equation}
     \label{eqbinf}
     \|\mathbf{b}_{s, 0}^{l}\|_{\infty}=\Tilde{O}\left(\frac{1}{\sqrt{m}}\right).
 \end{equation}
\end{Lemma}
\begin{proof}
We prove this lemma by induction. Before proceeding, lets denote $\mathbf{s}^{l}=\mathbf{b}_{s, 0}^{l}$.
Initially, for $l=L$, we have
\begin{equation*}
    \left\|\mathbf{s}^{L}\right\|_{\infty}=1 / \sqrt{m}\left\|\mathbf{v}_{s}\right\|_{\infty}={O}(1 / \sqrt{m}).
\end{equation*} 
Implies that \eqref{eqbinf} holds true for $l=L$.
Suppose that at $l^{th}$ layer with probability at least $1-m e^{-c_{b s}^{l} \ln ^{2}(m)}$, for some constant $c_{b s}^{l}>0,\left\|\mathbf{s}^{l}\right\|_{\infty} = \Tilde{O}(\frac{1}{\sqrt{m}})$.
We now prove that equation \eqref{eqbinf} is valid for $(l-1)^{th}$ layer as well with probability at least $1-m e^{-c_{b s}^{l-1} \ln ^{2}(m)}$ for some constant $c_{b s}^{l-1}>0.$
% We analyze every component of $s^{l-1}$
In particular, the absolute value of $i^{th}$ component of $\mathbf{s}_{i}^{l-1}$ is bounded as 
\begin{equation*}
    \begin{aligned}
\left|\mathbf{s}_{i}^{l-1}\right|&=\left|\frac{1}{\sqrt{m}} \sum_{k=1}^{m}\left(W_{2}^{l-1}\right)_{ki} \sigma^{\prime}\left(\frac{1}{\sqrt{m}} \sum_{j=1}^{m}\left(W_{2}^{l-1}\right)_{k j} \mathbf{x}_{j}^{l-2}+\frac{1}{\sqrt{n}} \sum_{j=1}^{n}\left(W_{1}^{l-1}\right)_{k j} \mathbf{y}_{j}\right) \mathbf{s}_{k}^{l}\right| \\
&\leq \left|\frac{1}{\sqrt{m}} \sum_{k=1}^{m}\left(W_{2}^{l-1}\right)_{ki} \sigma^{\prime}\left(\frac{1}{\sqrt{m}} \sum_{j \neq i}^{m}\left(W_{2}^{l-1}\right)_{k j} \mathbf{x}_{j}^{l-2}+\frac{1}{\sqrt{n}} \sum_{j \neq i}^{n}\left(W_{1}^{l-1}\right)_{k j} \mathbf{y}_{j}\right) \mathbf{s}_{k}^{l}\right| \\
&+\left|\frac{1}{m} \beta_{\sigma} \mathbf{x}_{i}^{l-2} \sum_{k=1}^{m}\left(\left(W_{2}^{l-1}\right)_{ki}\right)^{2} \mathbf{s}_{k}^{l}\right| + \left|\frac{1}{\sqrt{m} \sqrt{n}} \beta_{\sigma} \mathbf{y}_{i} \sum_{k=1}^{m}\left(W_{1}^{l-1}\right)_{ki}\left(W_{2}^{l-1}\right)_{ki} \mathbf{s}_{k}^{l}\right|\\
& = |T_1| + |T_2| +|T_3|.
\end{aligned}
\end{equation*}
Now, we provide bounds on the terms ($T_1, T_2,$ and $T_3$) individually:
% and find their corresponding distribution. 

\begin{equation*}
    \begin{aligned}
T_1 = & \frac{1}{\sqrt{m}} \sum_{k=1}^{m}\left(W_{2}^{l-1}\right)_{ki} \sigma^{\prime}\left(\frac{1}{\sqrt{m}} \sum_{j \neq i}^{m}\left(W_{2}^{l-1}\right)_{k j} \mathbf{x}_{j}^{l-2}+\frac{1}{\sqrt{n}} \sum_{j \neq i}^{m}\left(W_{1}^{l-1}\right)_{k j} \mathbf{y}_{j}\right) \mathbf{s}_{k}^{l} \\
\leq & \frac{L_{\sigma}}{\sqrt{m}} \sum_{k=1}^{m}\left(W_{2}^{l-1}\right)_{ki} \mathbf{s}_{k}^{l} \sim \mathcal{N}\left(0, \frac{L_{\sigma}^{2}}{m}\left\|\mathbf{s}^{l}\right\|^{2}\right),
\end{aligned}
\end{equation*}

\begin{equation*}
    \begin{aligned}
&T_2 = \frac{1}{m} \beta_{\sigma} \mathbf{x}_{i}^{l-2} \sum_{k=1}^{m}\left(\left(W_{2}^{l-1}\right)_{ki}\right)^{2} \mathbf{s}_{k}^{l} \leq \frac{1}{m} \beta_{\sigma}\left|\mathbf{x}_{i}^{l-2}\right|\left\|\mathbf{s}^{l}\right\|_{\infty} \sum_{k=1}^{m}\left(\left(W_{2}^{l-1}\right)_{ki}\right)^{2}, \\
&T_3 = \frac{1}{\sqrt{m} \sqrt{n}} \beta_{\sigma} \mathbf{y}_{i} \sum_{k=1}^{m}\left(W_{1}^{l-1}\right)_{ki}\left(W_{2}^{l-1}\right) \mathbf{s}_{k}^{l} \leq \frac{1}{\sqrt{m} \sqrt{n}} \beta_{\sigma}\left|\mathbf{y}_{i}\right|\left\|\mathbf{s}^{l}\right\|_{\infty} \sum_{k=1}^{m}\left(W_{1}^{l-1}\right)_{ki}\left(W_{2}^{l-1}\right)_{ki},
\end{aligned}
\end{equation*}
where $\sum_{k=1}^{m}\left(\left(W_{2}^{l-1}\right)_{ki}\right)^{2} \sim \mathcal{\chi}^{2}(m)$, $\sum_{k=1}^{m}\left(W_{1}^{l-1}\right)_{ki}\left(W_{2}^{l-1}\right)_{ki} \sim \mathcal{\chi}^{2}(m)$, and $\mathcal{\chi}^{2}(m)$ denotes the chi-square distribution with degree $m$.
By using the concentration inequality on the derived $T_1$ bound, we obtain
\begin{equation}
\label{T1eq}\operatorname{Pr}\left[\left|\frac{L_{\sigma}}{\sqrt{m}} \sum_{k=1}^{m}\left(W_{2}^{l-1}\right)_{ki} \mathbf{s}_{k}^{l}\right| \geq \frac{\ln (m)}{\sqrt{m}}\right] \leq 2 e^{-\frac{\ln^{2}(m)}{2 L_{\sigma}^{2}\|\mathbf{s}^{l} \|^{2}}} \leq 2 e^{-c_{\sigma}^{l} \ln ^{2}(m)}.
\end{equation}
Substituting the bound of $\left\|\mathbf{s}^{l}\right\|$, obtained from Lemma (\ref{lm:7}), in the above inequality leads to $c_{\sigma}^{l}=1 /\left(2 L_{\sigma}^{2}\left\|\mathbf{s}^{l}\right\|^{2}\right)$ $\geq 1 /\left(2 L_{\sigma}^{2 L-2 l+2} c_{20}^{2 L-2 l}\right)$.
From Lemma 1 in \citesupp{appendixref}, there exist constants $\tilde{c}_{1}, \tilde{c}_{2}$, and $\tilde{c}_{3}>0$, such that
\begin{equation}
    \label{T2eq}
    \operatorname{Pr}\left[\left|\frac{1}{m} \beta_{\sigma} | \mathbf{x}_{i}^{l-2}|\left\|\mathbf{s}^{l}\right\|_{\infty} \sum_{k=1}^{m}\left(\left(W_{2}^{l-1}\right)_{ki}\right)^{2}\right| \geq \tilde{c}_{1} e^{-\frac{\ln^{\tilde{c}_{3}}(m)}{\sqrt{m}} }\right] \leq e^{-\tilde{c}_{2} m}.
\end{equation}
Here, by using Lemma (\ref{lm:6}), we can write $\left|\mathbf{x}_{i}^{l-2}\right| \leq \ln (m)+|\sigma(0)|$ with probability at least $1-2 e^{-c_{\mathbf{x}}^{l-2} \ln ^{2}(m)}$and by induction hypothesis we have $\left\|\mathbf{s}^{l}\right\|_{\infty}=\tilde{O}(1/\sqrt{m})$ with probability $1-m e^{-c_{b s}^{l} \ln^{2}(m)}$.  
 Similarly, there exist constants $\hat{c}_{1}$, $\hat{c}_{2}$, and $\hat{c}_{3}>0$, such that
\begin{equation}
    \label{T3eq}
    \operatorname{Pr}\left[\left|\frac{1}{\sqrt{mn}} \beta_{\sigma}| \mathbf{y}_{i}|\left\|\mathbf{s}^{l}\right\|_{\infty} \sum_{k=1}^{m}\left(W_{1}^{l-1}\right)_{ki}\left(W_{2}^{l-1}\right)_{ki}\right| \geq \hat{c}_{1} e^{-\frac{l n^{\hat{c}_{3}(\sqrt{mn})}}{\sqrt{\sqrt{mn}}}}\right] \leq e^{-\hat{c}_{2} \sqrt{mn}}.
\end{equation}
Combining probabilities in \eqref{T1eq}, \eqref{T2eq}, and \eqref{T3eq}, there exists a constant $c_{bs}^{l-1}$ such that
\begin{equation*}
    e^{-c_{bs}^{l-1} \ln ^{2}(m)} \leq m e^{-c_{b s}^{l} \ln ^{2}(m)}+2 e^{-c_{\sigma}^{l} \ln ^{2}(m)}+2 e^{-c_{\mathbf{x}}^{l} \ln ^{2}(m)}+e^{-\tilde{c}_{2} m}+e^{-\hat{c}_{2} \sqrt{mn}},
\end{equation*}
and with probability at least $1-e^{-c_{bs}^{l-1} \ln ^{2}(m)}$, we have $\left|s_{i}^{l-1}\right| =\Tilde{O}\left(\frac{1}{\sqrt{m}}\right)$. 
This implies, 
\begin{equation}
    \left\|\mathbf{s}^{l-1}\right\|_{\infty}  =\Tilde{O}\left(\frac{1}{\sqrt{m}}\right), 
\end{equation}
with probability at least $1-m e^{-c_{bs}^{l-1} \ln ^{2}(m)}$, 
i.e., by induction we proved \eqref{eqbinf} for any $l\in [L]$.
\end{proof}

\begin{Lemma}
    \label{lm:9}
   The $l_2$-norm of difference between $\mathbf{b}_{s}^{l}$ and $\mathbf{b}_{s,0}^{l}$ is in $\tilde{O}(1/\sqrt{m})$ for any $l \in [L-1]$,
   i.e.,
   \begin{equation}
       \label{d2eq}
       \left\|\mathbf{b}_{s}^{l}-\mathbf{b}_{s,0}^{l}\right\|=\tilde{O}(1/\sqrt{m}) \quad \forall l \in [L-1].
   \end{equation}

\end{Lemma}
\begin{proof}
%Firstly, we define
%$$
%\mathbf{b}_{s}=\left[\begin{array}{c}
%\mathbf{b}_{s}^{1} \\
%\mathbf{b}_{s}^{2} \\
%\vdots \\
%\mathbf{b}_{s}^{L}
%\end{array}\right]
%$$
we prove \eqref{d2eq} by using Induction.
For $l=L$, we have $\left\|\mathbf{b}_{s}^{(L)}-\mathbf{b}_{s, 0}^{(L)}\right\|=0$. 
Let us consider \eqref{d2eq} is valid for any $l\in[L]$. 
Now, we prove that \eqref{d2eq} is also valid for $l-1$. 
\begin{equation*}
    \begin{aligned}
\left\|\mathbf{b}_{s}^{l-1}-\mathbf{b}_{s,0}^{l-1}\right\|&= \frac{1}{\sqrt{m}}\left\|\left(W_{2}^{l}\right)^{T} \Sigma^{\prime l} \mathbf{b}_{s}^{l}-\left(W_{20}^{l}\right)^{T} \Sigma_{0}^{\prime l} \mathbf{b}_{s,0}^{l}\right\|  \\
&= \frac{1}{\sqrt{m}} \|\left(W_{2}^{l}\right)^{T} \Sigma^{\prime l} \mathbf{b}_{s}^{l}-\left(W_{20}^{l}\right)^{T} \Sigma_{0}^{\prime l} \mathbf{b}_{s,0}^{l}+\left(W_{20}^{l}\right)^{T} \Sigma^{\prime l} \mathbf{b}_{s,0}^{l} \\
&+\left(W_{20}^{l}\right)^{T} \Sigma^{\prime l} \mathbf{b}_{s}^{l}-\left(W_{20}^{l}\right)^{T} \Sigma^{\prime l} \mathbf{b}_{s,0}^{l}-\left(W_{20}^{l}\right)^{T} \Sigma^{\prime l} \mathbf{b}_{s}^{l} \| \\
&= \frac{1}{\sqrt{m}} \left\|\left(\left(W_{2}^{l}\right)^{T}-\left(W_{20}^{l}\right)^{T}\right) \Sigma^{\prime l} \mathbf{b}_{s}^{l}+\left(W_{20}^{l}\right)^{T}\left(\Sigma^{\prime l}-\Sigma_{0}^{\prime l}\right) \mathbf{b}_{s,0}^{l} 
+\left(W_{20}^{l}\right)^{T} \Sigma^{\prime l}\left(\mathbf{b}_{s}^{l}-\mathbf{b}_{s,0}^{l}\right) \right\| \\
&\leq  \frac{1}{\sqrt{m}}\left\|\left(\left(W_{2}^{l}\right)^{T}-\left(W_{20}^{l}\right)^{T}\right) \Sigma^{\prime l} \mathbf{b}_{s}^{l}\right\|+\frac{1}{\sqrt{m}}\left\|\left(W_{20}^{l}\right)^{T}\left(\Sigma^{\prime l}-\Sigma_{0}^{\prime l}\right) \mathbf{b}_{s,0}^{l}\right\| 
+\frac{1}{\sqrt{m}}\left\|\left(W_{20}^{l}\right)^{T} \Sigma^{\prime l}\left(\mathbf{b}_{s}^{l}-\mathbf{b}_{s,0}^{l}\right)\right\|\\
&= T_1+T_2+T_3.
\end{aligned}
\end{equation*}
We now provide bounds on $T_1, T_2,$ and $T_3$:
\begin{equation*}
\begin{aligned}
        T_1&=\frac{1}{\sqrt{m}}\left\|\left(\left(W_{2}^{l}\right)^{T}-\left(W_{20}^{l}\right)^{T}\right) \Sigma^{\prime l} \mathbf{b}_{s}^{l}\right\|
         \leq    \frac{1}{\sqrt{m}}\left\|W_{2}^{l}-W_{20}^{l}\right\| \left\| \Sigma^{\prime l} \right\|  \left\| \mathbf{b}_{s}^{l}\right\| \leq \frac{R_2 L_{\sigma}^{L-l+1}\left(c_{20}+R_{2} / \sqrt{m}\right)^{L-l}}{\sqrt{m}} =O(1/\sqrt{m}).
\end{aligned}
\end{equation*}
To obtain bound on $T_2$, we need the following inequality,
\begin{equation*}
    \begin{aligned}
\left\|\tilde{\mathbf{x}}^{l}(\mathbf{W})-\tilde{\mathbf{x}}^{l}\left(\mathbf{W}_{0}\right)\right\| 
=&\left\|\frac{1}{\sqrt{m}} W_{2}^{l} \mathbf{x}^{l-1}(\mathbf{W})-\frac{1}{\sqrt{m}} W_{20}^{l} \mathbf{x}^{l-1}\left(\mathbf{W}_{0}\right)+\frac{1}{\sqrt{n}} W_{1}^{l} \mathbf{y}-\frac{1}{\sqrt{n}} W_{10}^{l} \mathbf{y}\right\| \\
\leq & \frac{1}{\sqrt{m}}\left\|W_{20}^{l}\right\| L_{\sigma}\left\|\tilde{\mathbf{x}}^{l-1}(\mathbf{W})-\tilde{\mathbf{x}}^{l-1}\left(\mathbf{W}_{0}\right)\right\|+\frac{1}{\sqrt{m}}\left\|W_{2}^{l}-W_{20}^{l}\right\|\left\|\mathbf{x}^{l-1}(\mathbf{W})\right\| 
+\frac{1}{\sqrt{n}}\left\|W_{1}^{l}-W_{10}^{l}\right\|\|\mathbf{y}\| \\
\leq & c_{20} L_{\sigma}\left\|\tilde{\mathbf{x}}^{l-1}(\mathbf{W})-\tilde{\mathbf{x}}^{l-1}\left(\mathbf{W}_{0}\right)\right\|+\frac{R_{2}}{\sqrt{m}}\left\|\mathbf{x}^{l-1}(\mathbf{W})\right\|+R_{1} C_y \\
\leq & c_{20} L_{\sigma}\left\|\tilde{\mathbf{x}}^{l-1}(\mathbf{W})-\tilde{\mathbf{x}}^{l-1}\left(\mathbf{W}_{0}\right)\right\|+\frac{R_{2} c_{\mathrm{ISTA} ; \mathbf{x}}^{l-1}}{\sqrt{m}}+R_{1} C_y.
\end{aligned}
\end{equation*}
Since
\begin{equation*}
    \left\|\tilde{\mathbf{x}}^{(1)}(\mathbf{W})-\tilde{\mathbf{x}}^{(1)}\left(\mathbf{W}_{0}\right)\right\| \leq \frac{1}{\sqrt{m}}\left\|W_{2}^{(1)}-W_{20}^{(1)}\right\|\left\|\mathbf{x}^{(0)}\right\|+\frac{1}{\sqrt{n}}\left\|W_{1}^{(1)}-W_{10}^{(1)}\right\|\|\mathbf{y}\| \leq R_{2} C_{\mathbf{x}}+R_{1} C_y=O(1).
\end{equation*}
Recursively applying the previous equation, we get
\begin{equation*}
    \begin{aligned}
\left\|\tilde{\mathbf{x}}^{l}(\mathbf{W})-\tilde{\mathbf{x}}^{l}\left(\mathbf{W}_{0}\right)\right\| & \leq c_{20}^{l-1} L_{\sigma}^{l-1}\left(R_{2} C_{\mathbf{x}}+R_{1} C_y\right)+\left(\frac{R_{2} c_{\mathrm{ISTA} ; \mathbf{x}}^{l-1}}{\sqrt{m}}+R_{1} C_y\right) \sum_{i=1}^{l-2} c_{20}^{i} L_{\sigma}^{i} =O(1).
\end{aligned}
\end{equation*}
Using the above inequality bound and Lemma (\ref{lm:8}), we can write the following with probability $1-m e^{-c_{bs}^{l} \ln ^{2}(m)}$:
% Also, note that  $\left\|\mathbf{b}_{s,0}^{l}\right\|_{\infty} = \tilde{O}(\frac{1}{\sqrt{m}})$ with probability $1-m e^{-c_{b s}^{l} \ln ^{2}(m)}$ for some constant $c_{b s}^{l}>0$, according to Lemma (\ref{lm:8}) , then we have
\begin{equation*}
    \begin{aligned}
\left\|\left[\Sigma^{\prime l}-\Sigma_{0}^{\prime l}\right] \mathbf{b}_{s,0}^{l}\right\| &=\sqrt{\sum_{i=1}^{m}\left(\mathbf{b}_{s,0}^{l}\right)_{i}^{2}\left[\sigma^{\prime}\left(\tilde{\mathbf{x}}^{l}(\mathbf{W})\right)-\sigma^{\prime}\left(\tilde{\mathbf{x}}^{l}\left(\mathbf{W}_{0}\right)\right)\right]^{2}} 
\leq\left\|\mathbf{b}_{s,0}^{l}\right\|_{\infty} \sqrt{\sum_{i=1}^{m}\left[\sigma^{\prime}\left(\tilde{\mathbf{x}}^{l}(\mathbf{W})\right)-\sigma^{\prime}\left(\tilde{\mathbf{x}}^{l}\left(\mathbf{W}_{0}\right)\right)\right]^{2}} \\
& \leq\left\|\mathbf{b}_{s,0}^{l}\right\|_{\infty} \beta_{\sigma}\left\|\tilde{\mathbf{x}}^{l}(\mathbf{W})-\tilde{\mathbf{x}}^{l}\left(\mathbf{W}_{0}\right)\right\| 
 = \tilde{O}\left(\frac{1}{\sqrt{m}}\right).
\end{aligned}
\end{equation*}
This leads to,
\begin{equation*}
    T_2 = \frac{1}{\sqrt{m}}\left\|\left(W_{20}^{l}\right)^{T}\left(\Sigma^{\prime l}-\Sigma_{0}^{\prime l}\right) \mathbf{b}_{s,0}^{l}\right\|\leq \frac{1}{\sqrt{m}}\|W_{20}^{l}\|\left\|\left[\Sigma^{\prime l}-\Sigma_{0}^{\prime l}\right] \mathbf{b}_{s,0}^{l}\right\| = \tilde{O}\left(\frac{1}{\sqrt{m}}\right). 
\end{equation*}
Besides, by using the induction hypothesis on $l$, the term $T_3$ is bounded as
\begin{equation*}
    \begin{aligned}
        T_3=&\frac{1}{\sqrt{m}}\left\|\left(W_{20}^{l}\right)^{T} \Sigma^{\prime l}\left(\mathbf{b}_{s}^{l}-\mathbf{b}_{s,0}^{l}\right)\right\|
        \leq \frac{1}{\sqrt{m}}\left\|W_{20}^{l}\right\| \left\|\Sigma^{\prime l}\right\|\left\|\mathbf{b}_{s}^{l}-\mathbf{b}_{s,0}^{l}\right\|=\tilde{O}(1/\sqrt{m}).
    \end{aligned}
\end{equation*}
Now combining the bounds on the terms $T_1,\ T_2$, and $T_3$, we can write
\begin{equation}
\label{eq:}
\left\|\mathbf{b}_{s}^{l-1}-\mathbf{b}_{s,0}^{l-1}\right\|\leq T_1+T_2+T_3 = \tilde{O}\left( \frac{1}{\sqrt{m}} \right).
\end{equation}
Therefore, \eqref{d2eq} is true for $l-1$. Hence, by induction \eqref{d2eq} is true for all $l \in [L]$. 
\end{proof}
By using Lemma \ref{lm:8} and \ref{lm:9}, in equation \eqref{eq:inf}, we get
\begin{equation}
\left\|\mathbf{b}_{s}^{l}\right\|_{\infty} \leq\left\|\mathbf{b}_{s,0}^{l}\right\|_{\infty}+\left\|\mathbf{b}_{s}^{l}-\mathbf{b}_{s,0}^{l}\right\|  = \tilde{O}\left(\frac{1}{\sqrt{m}}\right).
\end{equation}
This implies,
\begin{equation}
\mathcal{Q}_{\infty}\left(f_s\right)= \max _{1 \leq l \leq L}\left\{\left\|\mathbf{b}_s^l\right\|_{\infty}\right\} = \tilde{O}\left(\frac{1}{\sqrt{m}}\right).
\end{equation}

\subsection{Bound on $Q_{\infty}$ For ADMM-CSNet}
\label{Q_Infinity_For_ADMMCSNet}
Let $\mathbf{b}_s^l = \frac{\partial f_s}{\partial \mathbf{z}^{l}}$, then $
    \mathcal{Q}_{\infty}\left(f_s\right)= \max _{1 \leq l \leq L}\left\{\left\|\mathbf{b}_s^l\right\|_{\infty}\right\}$.
We now compute bound on $\left\|\mathbf{b}_s^l\right\|_{\infty}$ by using \eqref{eq:inf}.
Similar to the previous LISTA network analysis, one can obtain the bound on $\left\|\mathbf{b}_{s}^{l}\right\|_{\infty}$ by computing the bounds on $\left\|\mathbf{b}_{s,0}^{l}\right\|_{\infty}$ and $\left\|\mathbf{b}_{0}^{l}-\mathbf{b}_{s,0}^{l}\right\|$, which are provided in
Lemma \ref{lm:15} and Lemma \ref{lm:16}, respectively.
Moreover, in order to compute the bound on $\left\|\mathbf{b}_{s,0}^{l}\right\|_{\infty}$, we require several lemmas which are stated below. In specific, Lemma \ref{lm:13} and Lemma \ref{lm:14} provide the bound on each component of the hidden layer's output at initialization and the bound on $l_2$-norm of $\mathbf{b}_s^l,\ l\in[L]$, respectively. 
% To prove $\mathcal{Q}_{\infty}(f_s)=O(1/\sqrt{m})$ for all $s \in [m]$, we require several lemmas which are stated below with proofs.
%Like Unfolded ISTA Network, here also we require several lemmas to show  $\mathcal{Q}_{\infty}(f_s)=O(1/\sqrt{m})$ for all $s \in [m]$, which are stated below with proofs.
\begin{Lemma}
    \label{lm:13}
  For any $l\in [L]$ and $i\in[m]$, we have $\left|\mathbf{z}_{i}^{l}\right|\leq \ln (m)+L_{\sigma}\left|\mathbf{u}_{i}^{l-1}\right|+|\sigma(0)|$ at initialization
     with probability at least $1-2 e^{-c_{\mathbf{z}}^{l} l n^{2}(m)}$ for some constant $c_{\mathbf{z}}^{l}>0$ and $\left|\mathbf{u}_{i}^{l}\right| \leq \ln (m)+\left|\mathbf{u}_{i}^{l-1}\right|+\left|\mathbf{z}_{i}^{l}\right|$ at initialization with probability at least $1-2 e^{-c_{\mathbf{u}}^{l} \ln ^{2}(m)}$ for some constant $c_\mathbf{u}^l>0$.
\end{Lemma}
\begin{proof}
From \eqref{eq:6},
\begin{equation*}
   \begin{aligned}
\left|\mathbf{z}_{i}^{l}\right| &=\left|\sigma\left(\mathbf{u}_{i}^{l-1}+ \sum_{k=1}^{m} \frac{1}{\sqrt{m}} \left(W_{2}^{l}\right)_{ik}\left(\mathbf{z}_{k}^{l-1}-\mathbf{u}_{k}^{l-1}\right)+\sum_{k=1}^{n}\frac{1}{\sqrt{n}}\left(W_{1}^{l}\right)_{ik} \mathbf{y}_{k}\right)\right| \\
& \leq\left|\frac{L_{\sigma}}{\sqrt{m}} \sum_{k=1}^{m}\left(W_{2}^{l}\right)_{ik}\left(\mathbf{z}_{k}^{l-1}-\mathbf{u}_{k}^{l-1}\right)+\frac{L_{\sigma}}{\sqrt{n}} \sum_{k=1}^{n}\left(W_{1}^{l}\right)_{ik} \mathbf{y}_{k}\right|+L_{\sigma}\left|\mathbf{u}_{i}^{l-1}\right|+|\sigma(0)|.
\end{aligned} 
\end{equation*}
As $\left(W_{1}^{l}\right)_{ik} \sim$ $\mathcal{N}(0,1)$ and $\left(W_{2}^{l}\right)_{ik} \sim \mathcal{N}(0,1)$, so that $\sum_{k=1}^{n}\left(W_{1}^{l}\right)_{ik} \mathbf{y}_{k}$ $\sim \mathcal{N}\left(0,\|\mathbf{y}\|^{2}\right)$ and $\sum_{k=1}^{m}\left(W_{2}^{l}\right)_{ik}\left(\mathbf{z}_{k}^{l-1}-\mathbf{u}_{k}^{l-1}\right) \sim \mathcal{N}\left(0,\left\|\mathbf{z}^{l-1}-\mathbf{u}^{l-1}\right\|^{2}\right)$. \
In addition, since $\left(W_{1}^{l}\right)_{ik}$ and $\left(W_{2}^{l}\right)_{ik}$ are independent, \\
$\sum_{k=1}^{m}\left(W_{1}^{l}\right)_{ik} \mathbf{y}_{k}+$ $\sum_{k=1}^{m}\left(W_{2}^{l}\right)_{ik}\left(\mathbf{z}_{k}^{l-1}-\mathbf{u}_{k}^{l-1}\right) \sim \mathcal{N}\left(0,\|\mathbf{y}\|^{2}+\left\|\mathbf{z}^{l-1}-\mathbf{u}^{l-1}\right\|^{2}\right)$.
Using the concentration inequality of a Gaussian random variable, we obtain
%where $\sum_{k=1}^{m}\left(W_{1}^{l}\right)_{ik} \mathbf{y}_{k} \sim \mathcal{N}\left(0,\|\mathbf{y}\|^{2}\right)$ and $\sum_{k=1}^{m}\left(W_{2}^{l}\right)_{ik}\left(\mathbf{z}_{k}^{l-1}-\mathbf{u}_{k}^{l-1}\right) \sim \mathcal{N}\left(0,\left\|\mathbf{z}^{l-1}-\mathbf{u}^{l-1}\right\|^{2}\right)$ since $\left(W_{1}^{l}\right)_{ik} \sim \mathcal{N}(0,1)$ and $\left(W_{2}^{l}\right)_{ik} \sim \mathcal{N}(0,1)$. As $\left(W_{1}^{l}\right)_{ik}$ and $\left(W_{2}^{l}\right)_{ik}$ are independent, $\sum_{k=1}^{m}\left(W_{1}^{l}\right)_{ik} \mathbf{y}_{k}+$ $\sum_{k=1}^{m}\left(W_{2}^{l}\right)_{ik}\left(\mathbf{z}_{k}^{l-1}-\mathbf{u}_{k}^{l-1}\right) \sim \mathcal{N}\left(0,\|\mathbf{y}\|^{2}+\left\|\mathbf{z}^{l-1}-\mathbf{u}^{l-1}\right\|^{2}\right)$. By the concentration inequality for Gaussian random variable, we have
\begin{equation*}
    \begin{aligned}
\operatorname{Pr}\left[\left|\mathbf{z}_{i}^{l}\right| \geq \ln (m)+L_{\sigma}\left|\mathbf{u}_{i}^{l-1}\right|+|\sigma(0)|\right] 
&\leq  \operatorname{Pr}\left[\left|\frac{L_{\sigma}}{\sqrt{m}} \sum_{k=1}^{m}\left(W_{2}^{l}\right)_{ik}\left(\mathbf{z}_{k}^{l-1}-\mathbf{u}_{k}^{l-1}\right)+\frac{L_{\sigma}}{\sqrt{n}} \sum_{k=1}^{n}\left(W_{1}\right)_{i k}^{(1)} \mathbf{y}_{k}\right| \geq \ln (m)\right] \\
&\leq  2 e^{-\frac{m \ln^{2}(m)}{2 L_{\sigma}^{2}\left(\|\mathbf{y}\|^{2}+\left\|\mathbf{z}^{l-1}-\mathbf{u}^{l-1}\right\|^{2}\right)}} 
= 2 e^{-c_{\mathbf{z}}^{l} \ln ^{2}(m)},
\end{aligned}
\end{equation*}
where $c_{\mathbf{z}}^{l}=\frac{m}{2 L_{\sigma}^{2}\left(\|\mathbf{y}\|^{2}+\left\|\mathbf{z}^{(l-1)}-\mathbf{u}^{(l-1)}\right\|^{2}\right)}$. 
Therefore,
\begin{equation*}
    \begin{aligned}
\operatorname{Pr}\left[\left|\mathbf{z}_{i}^{l}\right|\right.&\left.\leq \ln (m)+L_{\sigma}\left|\mathbf{u}_{i}^{l-1}\right|+|\sigma(0)|\right] \geq 1-2 e^{-c_{\mathbf{z}}^{(l)} \ln ^{2}(m)}.
\end{aligned}
\end{equation*}
Since the bound on $\left|\mathbf{z}_{i}^{l}\right|$ depends on $\left|\mathbf{u}_{i}^{l-1}\right|$ (mentioned in above equation), we now find the bound of $\left|\mathbf{u}_{i}^{l}\right|$,
\begin{equation*}
    \begin{aligned}
\left|\mathbf{u}_{i}^{l}\right| &=\left|\mathbf{u}_{i}^{l-1}- \mathbf{z}_{i}^{l}+ \sum_{k=1}^{n} \frac{1}{\sqrt{n}}\left(W_{1}^{l}\right)_{ik} \mathbf{y}_{k}+\sum_{k=1}^{m}\frac{1}{\sqrt{m}}\left(W_{2}^{l}\right)_{ik}\left(\mathbf{z}_{k}^{l-1}-\mathbf{u}_{k}^{l-1}\right)\right| \\
& \leq\left|\mathbf{u}_{i}^{l-1}\right|+\left|\mathbf{z}_{i}^{l}\right|+\left|\sum_{k=1}^{n} \frac{1}{\sqrt{n}}\left(W_{1}^{l}\right)_{ik} \mathbf{y}_{k}+\sum_{k=1}^{m}\frac{1}{\sqrt{m}}\left(W_{2}^{l}\right)_{ik}\left(\mathbf{z}_{k}^{l-1}-\mathbf{u}_{k}^{l-1}\right)\right|.
\end{aligned}
\end{equation*}
By the concentration inequality for the Gaussian random variable, we have
\begin{equation*}
    \begin{aligned}
\operatorname{Pr}\left[\left|\mathbf{u}_{i}^{l}\right| \geq \ln (m)+\left|\mathbf{u}_{i}^{l-1}\right|+\left|\mathbf{z}_{i}^{l}\right|\right] 
&\leq\operatorname{Pr}\left[\left|\frac{L_{\sigma}}{\sqrt{m}} \sum_{k=1}^{m}\left(W_{2}^{l}\right)_{ik}\left(\mathbf{z}_{k}^{l-1}-\mathbf{u}_{k}^{l-1}\right)+\frac{L_{\sigma}}{\sqrt{n}} \sum_{k=1}^{n}\left(W_{1}^{l}\right)_{ik} \mathbf{y}_{k}\right| \geq \ln (m)\right] \\
\leq & 2 e^{-\frac{m \ln^{2}(m)}{2 L_{\sigma}^{2}\left(\|\mathbf{y}\|^{2}+\left\|\mathbf{z}^{l-1}-\mathbf{u}^{l-1}\right\|^{2}\right)}}.
\end{aligned}
\end{equation*}
Therefore, we have
\begin{equation*}
\operatorname{Pr}\left[\left|\mathbf{u}_{i}^{l}\right| \leq \ln (m)+\left|\mathbf{u}_{i}^{l-1}\right|+\left|\mathbf{z}_{i}^{l}\right|\right] \geq 1-2 e^{-c_{\mathbf{u}}^{l} \ln ^{2}(m)}.
\end{equation*}
In a recursive manner, we get
\begin{equation*}
    \begin{aligned}
\left|\mathbf{z}_{i}^{l}\right| & \leq \ln (m)+|\sigma(0)|+\sum_{i=0}^{l-2}\left(1+L_{\sigma}\right)^{i} L_{\sigma}(  2 \ln (m)+|\sigma(0)|)+\left(1+L_{\sigma}\right)^{l-1} L_{\sigma} C_{\mathbf{u}},
 \\
\left|\mathbf{u}_{i}^{l}\right| & \leq \sum_{i=0}^{l-1}\left(1+ L_{\sigma}\right)^{i}( 2 \ln (m)+|\sigma(0)|)+\left(1+ L_{\sigma}\right)^{l} C_{\mathbf{u}},
\end{aligned}
\end{equation*}
with possibility $1-2 e^{-\frac{m \ln^{2}(m)}{2 L_{\sigma}^{2}\left(\|\mathbf{y}\|^{2}+\left\|\mathbf{z}^{l-1}-\mathbf{u}^{l-1}\right\|^{2}\right)}}$.
\end{proof}
% \begin{definition}
%  Define $\mathbf{b}_{s}^{l}=\partial f_{s} / \partial \mathbf{z}^{l}$ for $l \in[L]$, and $\mathbf{b}_{s,0}^{l}$ denotes $\mathbf{b}_{s}^{l}$ at initialization. Specifically, $\mathbf{b}_{s}^{l}$ takes the following form in ADMM-Net:
% \begin{equation*}
%     \mathbf{b}_{s}^{l}=\prod_{l^{\prime}=l+1}^{L}\left(\left(\frac{2}{\sqrt{m}} W_{2}^{l}- I\right)^{T} \Sigma^{\prime\left(l^{\prime}\right)}\right) \frac{1}{\sqrt{m}} v_{s}
% \end{equation*}
% where $\Sigma^{\prime l^{\prime}}$ is a diagonal matrix with $\left(\Sigma^{\prime  ^{\prime}}\right)_{ss}=\sigma^{\prime}\left(\tilde{\mathbf{z}}_{s}^{l^{\prime}}\right)$, and $v_{s}$ is a vector with sth element to be 1 and others to be 0.
% \end{definition}
\begin{Lemma}
    \label{lm:14}
    Consider an $L$-layer ADMM-CSNet with $\left(W_{10}^{l}\right)_{i, j} \sim \mathcal{N}(0,1)$ and $\left(W_{20}^{l}\right)_{i, j} \sim \mathcal{N}(0,1)$, $\forall l\in[L]$, then, for any $\mathbf{W}_{1}$ and $\mathbf{W}_{2}$ such that $\left\|\mathbf{W}_{1}-\mathbf{W}_{10}\right\| \leq R_{1}$ and $\left\|\mathbf{W}_{2}-\mathbf{W}_{20}\right\| \leq R_{2}$, we have,
    %If the initial parameters $\mathbf{W}_{10}$ and $\mathbf{W}_{20}$ of the multi-layer neural network satisfy Assumptions, then, for any $\mathbf{W}_{1}$ and $\mathbf{W}_{2}$ such that $\left\|\mathbf{W}_{1}-\mathbf{W}_{10}\right\| \leq R_{1}$ and $\left\|\mathbf{W}_{2}-\mathbf{W}_{20}\right\| \leq R_{2}$, we have, at all hidden layers, i.e., $l=1,2, \cdots, L$,
\begin{equation}
    \label{eq:17}
    \left\|\mathbf{b}_{s}^{l}\right\| \leq L_{\sigma}^{L-l}\left(2\left(c_{20}+R_2 / \sqrt{m}\right)+1\right)^{L-l}.
\end{equation}
From this at initialization, i.e., for $R_2=0$, we get
\begin{equation}
    \label{eq:18}
\left\|\mathbf{b}_{s,0}^{l}\right\| \leq L_{\sigma}^{L-l}\left(2 c_{20}+1\right)^{L-l}.
\end{equation}

%In \eqref{eq:17}, $\mathbf{b}_{s}^{l}=\partial f_{s} / \partial \mathbf{z}^{l}$ for $l=1,2, \cdots, L-1$, and $\mathbf{b}_{s,0}^{l}$ denotes $\mathbf{b}_{s}^{l}$ at initialization. Specifically, $\mathbf{b}_{s}^{l}$ takes the following form in ADMM-Net:

%$$
%\mathbf{b}_{s}^{l}=\prod_{l^{\prime}=l+1}^{L}\left(\left(\frac{2}{\sqrt{m}} W_{2}^{l}- I\right)^{T} \Sigma^{\prime\left(l^{\prime}\right)}\right) \frac{1}{\sqrt{m}} v_{s}
%$$

\end{Lemma}

\begin{proof}
We prove this lemma by using induction on $l$.
Initially, for $l=L$, we have
\begin{equation*}
    \left\|\mathbf{b}_{s}^{L}\right\|=\left\|\frac{\partial f_s}{\partial \mathbf{z}^{L}}\right\|=(1 / \sqrt{m })\left\|\mathbf{v}_{s}\right\|=1 / \sqrt{m }<1.
\end{equation*}
That is the quantity in \eqref{eq:17} is true for $l=L$.
Assume that at $l^{th}$ layer the inequality holds, i.e., $\left\|\mathbf{b}_{s}^{l}\right\| \leq L_{\sigma}^{L-l}\left(2\left(c_{20}+R_2 / \sqrt{m}\right)+1\right)^{L-l}$, then below we prove that \eqref{eq:17} holds true even for the $(l-1)^{th}$  layer:
\begin{equation*}
    \begin{aligned}
\left\|\mathbf{b}_{s}^{l-1}\right\| &=\left\|\frac{\partial f_s}{\partial \mathbf{z}^{l-1}}\right\|=\left\|\frac{\partial \mathbf{z}^l}{\partial \mathbf{z}^{l-1}}\frac{\partial f_s}{\partial \mathbf{z}^{l}}\right\|=\left\|\left(\frac{2}{\sqrt{m}}\left(W_{2}^{l}\right)^{T} \Sigma^{\prime l}-\Sigma^{\prime l}\right) \mathbf{b}_{s}^{l}\right\|
\leq \frac{2}{\sqrt{m}}\left\|\left(W_2^{l}\right)\right\|\left\|\Sigma^{\prime l}\right\|\left\|\mathbf{b}_{s}^{l}\right\|+\left\|\Sigma^{\prime l}\right\|\left\|\mathbf{b}_{s}^{l}\right\| \\
& \leq\left(2\left(c_{20}+R_2 / \sqrt{m}\right)+1\right) L_{\sigma}\left\|\mathbf{b}_{s}^{l}\right\| 
\leq\left(2\left(c_{20}+R_2 / \sqrt{m}\right)+1\right)^{L-l+1} L_{\sigma}^{L-l+1}.
\end{aligned}
\end{equation*}
% i.e. For $l-1$, \eqref{eqlm71} is true.
So, from the above analysis, we claim that the inequality in \eqref{eq:17} holds true for any $l\in [L]$.
Now, at initialization, i.e., substituting $R_2=0$ in \eqref{eq:17} directly leads to \eqref{eq:18}.
\end{proof}

We now use the two lemmas that are mentioned above to provide the bound on $\left\|\mathbf{b}_{s, 0}^{l}\right\|_{\infty}$.
\begin{Lemma}
    \label{lm:15}
At initialization, the $\infty$-norm of $\mathbf{b}_s^l$ is in  $\Tilde{O}(1/\sqrt{m})$ with probability $1-m e^{-c_{b s}^{l} \ln ^{2}(m)}$ for some constant $c_{b s}^{l}>0,$
 %With probability at least $1-m e^{-c_{b s}^{l} \ln ^{2}(m)}$ for some constant $c_{b s}^{l}>0,\left\|\mathbf{b}_{s, 0}^{l}\right\|_{\infty} = $ %$ \frac{\ln (m)}{\sqrt{m}}+\tilde{c}_{1} e^{-\frac{\ln \tilde{c}_{3}(m)}{\sqrt{m}}}+\hat{c}_{1} e^{-\frac{\ln ^{c_{3}(m)}}{\sqrt{m}}}$.
 i.e., 
 \begin{equation}
     \label{eqbinfad}
     \|\mathbf{b}_{s, 0}^{l}\|_{\infty}=\Tilde{O}\left(\frac{1}{\sqrt{m}}\right).
 \end{equation}
\end{Lemma}
\begin{proof}
We prove this lemma by induction. Before proceeding, lets denote $\mathbf{s}^{l}=\mathbf{b}_{s, 0}^{l}$.
Initially, for $l=L$, we have
\begin{equation*}
    \left\|\mathbf{s}^{L}\right\|_{\infty}=1 / \sqrt{m}\left\|\mathbf{v}_{s}\right\|_{\infty}={O}(1 / \sqrt{m}).
\end{equation*} 
Implies that \eqref{eqbinfad} holds true for $l=L$.
Suppose that at $l^{th}$ layer with probability at least $1-m e^{-c_{b s}^{l} \ln ^{2}(m)}$, for some constant $c_{b s}^{l}>0,\left\|\mathbf{s}^{l}\right\|_{\infty} = \Tilde{O}(\frac{1}{\sqrt{m}})$.
We now prove that equation \eqref{eqbinfad} is valid for $(l-1)^{th}$ layer with probability at least $1-m e^{-c_{b s}^{l-1} \ln ^{2}(m)}$ for some constant $c_{b s}^{l-1}>0.$
% We analyze every component of $s^{l-1}$
In particular, the absolute value of $i^{th}$ component of $\mathbf{s}_{i}^{l-1}$ is bounded as

\begin{equation*}
\begin{aligned}
\left|s_{i}^{l-1}\right| 
&=\left|\sum_{k=1}^{m}\left(\frac{2}{\sqrt{m}} W_{2}^{l-1}- I\right)_{k i} \sigma^{\prime}\left(\frac{1}{\sqrt{m}} \sum_{j=1}^{m}\left(W_{2}^{l-1}\right)_{k j}\left(\mathbf{z}^{(l-2)}-\mathbf{u}^{(l-2)}\right)_{j}+\frac{1}{\sqrt{n}} \sum_{j=1}^{n}\left(W_{1}^{l-1}\right)_{k j} \mathbf{y}_{j}+\mathbf{u}_{k}^{(l-2)}\right) \mathbf{s}_{k}^{l}\right| \\
&\leq\left|\sum_{k=1}^{m}\left(\frac{2}{\sqrt{m}} W_{2}^{l-1}- I\right)_{k i} \sigma^{\prime}\left(\frac{1}{\sqrt{m}} \sum_{j \neq i}^{m}\left(W_{2}^{l-1}\right)_{k j}\left(\mathbf{z}^{(l-2)}-\mathbf{u}^{(l-2)}\right)_{j}+\frac{1}{\sqrt{n}} \sum_{j \neq i}^{n}\left(W_{1}^{l-1}\right)_{k j} \mathbf{y}_{j}+\mathbf{u}_{k}^{(l-2)}\right) \mathbf{s}_{k}^{l}\right|  \\
&+\left|\frac{2}{m} \beta_{\sigma}\left(\mathbf{z}_{i}^{(l-2)}-\mathbf{u}_{i}^{(l-2)}\right) \sum_{k=1}^{m}\left(W_{2}^{l-1}\right)_{ki}\left(W_{2}^{l-1}\right)_{ki} \mathbf{s}_{k}^{l}\right|+\left|\frac{2}{\sqrt{mn}} \beta_{\sigma} \mathbf{y}_{i} \sum_{k=1}^{m}\left(W_{2}^{l-1}\right)_{ki}\left(W_{1}^{l-1}\right)_{ki} \mathbf{s}_{k}^{l}\right| \\
&+\left|\frac{1}{\sqrt{m}} \beta_{\sigma}\left(\mathbf{z}_{i}^{(l-2)}-\mathbf{u}_{i}^{(l-2)}\right) \sum_{k=1}^{m}\left(W_{2}^{l-1}\right)_{ki} \mathbf{s}_{k}^{l}\right|+\left|\frac{1}{\sqrt{n}} \beta_{\sigma} \mathbf{y}_{i} \sum_{k=1}^{m}\left(W_{1}^{l-1}\right)_{ki} \mathbf{s}_{k}^{l}\right|+ \left| L_\sigma \sum_{k=1}^{m}\left(\frac{2}{\sqrt{m}} W_{2}^{l-1}- I\right)_{k i} \mathbf{s}_{k}^{l} \right|\\
&=|T_1|+|T_2|+|T_3|+|T_4|+|T_5|+|T_6|.
\end{aligned}
\end{equation*}
Now, we provide bounds on the terms ($T_1, T_2, T_3,T_4,T_5$, and $T_6$) individually:

\begin{equation*}
    \begin{aligned}
|T_1|=&\left|\ \sum_{k=1}^{m}\left(\frac{2}{\sqrt{m}} W_{2}^{l-1}- I\right)_{k i} \sigma^{\prime}\left(\frac{1}{\sqrt{m}} \sum_{j \neq i}^{m}\left(W_{2}^{l-1}\right)_{k j}\left(\mathbf{z}^{(l-2)}-\mathbf{u}^{(l-2)}\right)_{j}+\frac{1}{\sqrt{n}} \sum_{j \neq i}^{n}\left(W_{1}^{l-1}\right)_{k j} \mathbf{y}_{j}+\mathbf{u}_{k}^{(l-2)}\right) \mathbf{s}_{k}^{l} \right|\ \\
&\leq \left|L_{\sigma} \sum_{k=1}^{m}\left(\frac{2}{\sqrt{m}} W_{2}^{l-1}- I\right)_{k i} \mathbf{s}_{k}^{l} \right|\ \leq\left|\ L_{\sigma} \sum_{k=1}^{m} \frac{2}{\sqrt{m}}\left(W_{2}^{l-1}\right)_{ki} \mathbf{s}_{k}^{l} \right|\ + 
 \left|\ L_{\sigma} s_{i}^{l} \right|,
\end{aligned}
\end{equation*}
\begin{equation*}
    \begin{aligned}
|T_2|=&\left|\frac{2}{m} \beta_{\sigma}\left(\mathbf{z}_{i}^{(l-2)}-\mathbf{u}_{i}^{(l-2)}\right) \sum_{k=1}^{m}\left(\left(W_{2}^{l-1}\right)_{ki}\right)^{2} \mathbf{s}_{k}^{l}\right| \leq \frac{2}{m} \beta_{\sigma}\left|\mathbf{z}_{i}^{(l-2)}-\mathbf{u}_{i}^{(l-2)}\right|\left\|\mathbf{s}^{l}\right\|_{\infty} \sum_{k=1}^{m}\left(\left(W_{2}^{l-1}\right)_{ki}\right)^{2}, 
\end{aligned}
\end{equation*}
\begin{equation*}
    \begin{aligned}
|T_3|=&\left|\frac{2}{\sqrt{mn}} \beta_{\sigma} \mathbf{y}_{i} \sum_{k=1}^{m}\left(W_{2}^{l-1}\right)_{ki}\left(W_{1}^{l-1}\right)_{ki} \mathbf{s}_{k}^{l}\right| \leq \frac{2}{\sqrt{mn}} \beta_{\sigma}\left|\mathbf{y}_{i}\right|\left\|\mathbf{s}^{l}\right\|_{\infty} \left|\sum_{k=1}^{m}\left(W_{2}^{l-1}\right)_{ki}\left(W_{1}^{l-1}\right)_{ki}\right|,
\end{aligned}
\end{equation*}
\begin{equation*}
    \begin{aligned}
        |T_4|&=\left|\frac{1}{\sqrt{m}} \beta_{\sigma}\left(\mathbf{z}_{i}^{(l-2)}-\mathbf{u}_{i}^{(l-2)}\right) \sum_{k=1}^{m}\left(W_{2}^{l-1}\right)_{ki} \mathbf{s}_{k}^{l}\right|, \quad
               |T_5|=\left|\frac{1}{\sqrt{n}} \beta_{\sigma} \mathbf{y}_{i} \sum_{k=1}^{m}\left(W_{1}^{l-1}\right)_{ki} \mathbf{s}_{k}^{l}\right|,
            \end{aligned}
\end{equation*}

\begin{equation*}
    \begin{aligned}
        |T_6|&=\left|L_\sigma \sum_{k=1}^{m}\left(\frac{2}{\sqrt{m}} W_{2}^{l-1}- I\right)_{k i} \mathbf{s}_{k}^{l}\right|
         \leq\left|\ L_{\sigma} \sum_{k=1}^{m} \frac{2}{\sqrt{m}}\left(W_{2}^{l-1}\right)_{ki} \mathbf{s}_{k}^{l} \right|\ + 
 \left|\ L_{\sigma} \mathbf{s}_{i}^{l} \right|.
    \end{aligned}
\end{equation*}
By using the concentration inequality on the derived $T_1$ and $T_6$ bounds, we obtain
\begin{equation}
    \label{eq:28}
\operatorname{Pr}\left[\left|\frac{2L_{\sigma}}{\sqrt{m}} \sum_{k=1}^{m}\left(W_{2}^{l-1}\right)_{ki} \mathbf{s}_{k}^{l}\right| \geq \frac{\ln (m)}{\sqrt{m}}\right] \leq 2 e^{-\frac{\ln^{2}(m)}{2 (2)^2 L_{\sigma}^{2}\left\|\mathbf{s}^{l}\right\|^{2}}} \leq 2 e^{-c_{\sigma}^{l} \ln^{2}(m)}.
\end{equation}
Substituting the bound of $\left\|\mathbf{s}^{l}\right\|$, obtained from Lemma (\ref{lm:14}), in the above inequality leads to  $c_{\sigma}^{l}=1 /(8 L_{\sigma}^{2}\left\|\mathbf{s}^{l}\right\|^{2}) \geq 1 /(8L_{\sigma}^{2 L-2 l+2}\left((2 c_{20}+1\right)^{2 L-2 l})$. Also using the induction hypothesis, we get
\begin{equation}
\label{eq:29}
    \left| L_{\sigma} s_{i}^{l}\right| \leq  L_{\sigma}\left\|\mathbf{s}^{l}\right\|_{\infty} =\tilde{O}(1/\sqrt{m}).
\end{equation}
%\leq \sum_{i=1}^{L-l}\left(\frac{3 \ln (m)}{\sqrt{m}}+\tilde{c}_{1} e^{-\frac{l \tilde{c}_{3(m)}}{\sqrt{m}}}+\hat{c}_{1} e^{-\frac{l n^{\hat{c}_{3}(m)}}{\sqrt{m}}}\right) \gamma^{i} L_{\sigma}^{i}+\frac{\gamma^{L-l+1} L_{\sigma}^{L-l+1} 1}{\sqrt{m}}
Therefore, from \eqref{eq:28} and \eqref{eq:29}, we get both $T_1$ and $T_6$ is  $\tilde{O}(1/\sqrt{m})$ with probability at least $1-2 e^{-c_{\sigma}^{l} l n^{2}(m)}$.
%$\frac{\ln (m)}{\sqrt{m}}+\sum_{i=1}^{L-l}\left(\frac{3 \ln (m)}{\sqrt{m}}+\tilde{c}_{1} e^{-\frac{\ln \tilde{c}_{3}(m)}{\sqrt{m}}}+\hat{c}_{1} e^{-\frac{\ln \hat{c}_{3}(m)}{\sqrt{m}}}\right) \gamma^{i} L_{\sigma}^{i}+$ $\frac{\gamma^{L-l+1} L_{\sigma}^{L-l+1} 1}{\sqrt{m}}$ with probability at least $1-2 e^{-c_{\sigma}^{l} l n^{2}(m)}$.
As $\sum_{k=1}^{m}\left(\left(W_{2}^{l-1}\right)_{ki}\right)^{2} \sim \chi^{2}(m)$ and $\sum_{k=1}^{m}\left(W_{1}^{l-1}\right)_{ki}\left(W_{2}^{l-1}\right)_{ki} \sim \chi^{2}(m)$.    
Hence, to derive bounds on $T_2$ and $T_3$, by using Lemma 1 in \citesupp{appendixref}, there exist constants $\tilde{c}_{1}, \tilde{c}_{2}$, and $\tilde{c}_{3}>0$, such that

%With probability at least $1-2 e^{-c_{\mathbf{z}}^{(l-2)} \ln ^{2}(m)}$, we have $\left|\mathbf{z}_{i}^{l-2}\right| \leq c_{\mathrm{ADMM} ; \mathbf{z}, i}^{l}(m)$ and $\left|\mathbf{u}_{i}^{l}\right| \leq c_{\mathrm{ADMM} ; \mathbf{u}, i}^{l}(m)$. Hence, by Lemma 1 in [4], there exist constants $\tilde{c}_{1}, \tilde{c}_{2}$, and $\tilde{c}_{3}>0$, such that
\begin{equation}
    \label{eq:30}
    \operatorname{Pr}\left[\left|\frac{2}{m} \beta_{\sigma}| \mathbf{z}_{i}^{(l-2)}|\left\|\mathbf{s}^{l}\right\|_{\infty} \sum_{k=1}^{m}\left(\left(W_{2}^{l-1}\right)_{ki}\right)^{2}\right| \geq \tilde{c}_{1} e^{-\frac{\ln ^{\tilde{c}_3}(m)}{\sqrt{m}}}\right] \leq e^{-\tilde{c}_{2} m}.
\end{equation}
Here, by using Lemma (\ref{lm:13}), we can write $\left|\mathbf{z}_{i}^{l-2}\right| \leq \ln (m)+L_{\sigma}|\mathbf{u}_{i}^{l-3}|+|\sigma(0)|$ with probability at least $1-2 e^{-c_{\mathbf{z}}^{l-2} \ln ^{2}(m)}$and by induction hypothesis we have $\left\|\mathbf{s}^{l}\right\|_{\infty}=\tilde{O}(1/\sqrt{m})$ with probability $1-m e^{-c_{b s}^{l} \ln^{2}(m)}$.
%with probability $1-m e^{-c_{b}^{l} l n^{2}(m)}$ by the induction hypothesis.
Similarly, there exist constants $\hat{c}_{1}$, $\hat{c}_{2}$, and $\hat{c}_{3}>0$, such that
\begin{equation}
    \label{eq:31}
    \operatorname{Pr}\left[\left|\frac{2}{\sqrt{mn}} \beta_{\sigma}\left| \mathbf{y}_{i}\right|\left\|\mathbf{s}^{l}\right\|_{\infty} \sum_{k=1}^{m}\left(\left(W_{1}^{l-1}\right)_{ki}\right)^{2}\right| \geq \hat{c}_{1} e^{-\frac{\ln ^{\hat{c}_{3}(m)}}{\sqrt{\sqrt{mn}}}}\right] \leq e^{-\hat{c}_{2} \sqrt{mn}}.
\end{equation}
%with probability $1-m e^{-c_{b s}^{l} \ln ^{2}(m)}$.
Again by using concentration inequality, we obtain the bound for  $T_4$ and $T_5$ as follows.
\begin{equation}
\label{eq:32}
\operatorname{Pr}\left[\left|\frac{\beta_{\sigma}}{\sqrt{m}} \left(\mathbf{z}_{i}^{(l-2)}-\mathbf{u}_{i}^{(l-2)}\right) \sum_{k=1}^{m} (W_2)_{k i}^{l-1} \mathbf{s}_{k}^{l}\right| \geq \frac{ \ln (m)}{\sqrt{m}}\right] \leq 2 e^{-\frac{\ln^{2}(m)}{2 \beta_{\sigma}^{2}\left(\mathbf{z}_{i}^{(l-2)}-\mathbf{u}_{i}^{(l-2)}\right)^{2}\left\|\mathbf{s}^{l}\right\|^{2}}} \leq 2 e^{-c_{a \mathbf{z}} \ln ^{2}(m)},
\end{equation}
\begin{equation}
    \label{eq:33}
\operatorname{Pr}\left[\left|\frac{\beta_{\sigma}}{\sqrt{n}}  \mathbf{y}_{i} \sum_{k=1}^{m} (W_1)_{k i}^{l-1} \mathbf{s}_{k}^{l}\right| \geq \frac{ \ln (m)}{\sqrt{m}}\right] \leq 2 e^{-\frac{n \ln^{2}(m)}{2m \beta_{\sigma}^{2}\left(\mathbf{y}_{i}\right)^{2}\left\|\mathbf{s}^{l}\right\|^{2}}}\leq 2 e^{-\frac{\ln^{2}(m)}{2\beta_{\sigma}^{2}\left(\mathbf{y}_{i}\right)^{2}\left\|\mathbf{s}^{l}\right\|^{2}}} \leq 2 e^{-c_{a \mathbf{y}} \ln ^{2}(m)},
\end{equation}
for some constants $c_{a \mathbf{z}}=1 / 2 \beta_{\sigma}^{2}\left(\mathbf{z}_{i}^{(l-2)}-\mathbf{u}_{i}^{(l-2)}\right)^{2}\left\|\mathbf{s}^{l}\right\|^{2} \geq 1 / 2 \beta_{\sigma}^{2}\left(\mathbf{z}_{i}^{(l-2)}-\mathbf{u}_{i}^{(l-2)}\right)^{2} L_{\sigma}^{L-l}\left(2 c_{20}+1\right)^{L-l}$ and $c_{a \mathbf{y}}=1 / 2 \beta_{\sigma}^{2}\left(\mathbf{y}_{i}\right)^{2}\left\|\mathbf{s}^{l}\right\|^{2} \geq 1 / 2 \beta_{\sigma}^{2}\left(\mathbf{z}_{i}^{(l-2)}-\mathbf{u}_{i}^{(l-2)}\right)^{2} L_{\sigma}^{L-l}\left(2 c_{20}+1\right)^{L-l}$.
Combining probabilities in \eqref{eq:28}, \eqref{eq:29}, \eqref{eq:30}, \eqref{eq:31}, \eqref{eq:32} and \eqref{eq:33}, there exists a constant $c_{b s}^{l-1}$ such that
\begin{equation*}
    e^{-c_{b s}^{l-1} \ln ^{2}(m)} \leq 2m e^{-c_{b s}^{l} \ln ^{2}(m)}+4 e^{-c_{\sigma}^{l} \ln ^{2}(m)}+2 e^{-c_{\mathbf{z}}^{l} l n^{2}(m)}+e^{-\tilde{c}_{2} m}+e^{-\hat{c}_{2} \sqrt{mn}}+2e^{-c_{a \mathbf{z}} \ln ^{2}(m)}+2e^{-c_{a \mathbf{y}} \ln ^{2}(m)}
\end{equation*}
and with probability at least $1-e^{-c_{b s}^{l-1} \ln ^{2}(m)}$, we have$ \left|s_{i}^{l-1}\right| =\tilde{O}(1/\sqrt{m}) $. This implies
\begin{equation}
    \left\|s^{l-1}\right\|_{\infty} =\tilde{O}(1/\sqrt{m}),
\end{equation}
with probability at least $1-me^{-c_{b s}^{l-1} \ln ^{2}(m)},$ i.e. by induction we prove \eqref{eqbinfad} for any $l \in [L].$
\end{proof}
\begin{Lemma}
    \label{lm:16}
    The $l_2$-norm of difference between $\mathbf{b}_{s}^{l}$ and $\mathbf{b}_{s,0}^{l}$ is in $\tilde{O}(1/\sqrt{m})$ for any $l \in [L-1]$,
   i.e.,
   \begin{equation}
       \label{d2eqad}
       \left\|\mathbf{b}_{s}^{l}-\mathbf{b}_{s,0}^{l}\right\|=\tilde{O}(1/\sqrt{m}) \quad \forall l \in [L-1].
   \end{equation}
\end{Lemma}
\begin{proof}
we prove \eqref{d2eqad} by using induction.
For $l=L$, we have $\left\|\mathbf{b}_{s}^{(L)}-\mathbf{b}_{s, 0}^{(L)}\right\|=0$. 
Let us consider \eqref{d2eqad} is valid for any $l\in[L]$. 
Now, we prove that \eqref{d2eqad} is also valid for $l-1$. 
\begin{equation*}
    \begin{aligned}
\left\|\mathbf{b}_{s}^{l-1}-\mathbf{b}_{s,0}^{l-1}\right\|&=\left\|\left(\frac{2}{\sqrt{m}}\left(W_{2}^{l}\right)^{T} \Sigma^{\prime l}- \Sigma^{\prime l}\right) \mathbf{b}_{s}^{l}-\left(\frac{2}{\sqrt{m}}\left(W_{20}^{l}\right)^{T} \Sigma_{0}^{\prime l}- \Sigma_{0}^{\prime l}\right) \mathbf{b}_{s,0}^{l}\right\| \\
&= \|\left(\frac{2}{\sqrt{m}}\left(W_{2}^{l}\right)^{T} \Sigma^{\prime l}- \Sigma^{\prime l}\right) \mathbf{b}_{s}^{l}-\left(\frac{2}{\sqrt{m}}\left(W_{20}^{l}\right)^{T} \Sigma_{0}^{l l}- \Sigma_{0}^{\prime l}\right) \mathbf{b}_{s,0}^{l} \\
& \quad +\left(\frac{2}{\sqrt{m}}\left(W_{20}^{l}\right)^{T} \Sigma^{\prime l}- \Sigma^{\prime l}\right) \mathbf{b}_{s,0}^{l}+\left(\frac{2}{\sqrt{m}}\left(W_{20}^{l}\right)^{T} \Sigma^{\prime l}- \Sigma^{\prime l}\right) \mathbf{b}_{s}^{l} \\
& \quad -\left(\frac{2}{\sqrt{m}}\left(W_{20}^{l}\right)^{T} \Sigma^{\prime l}- \Sigma^{\prime l}\right) \mathbf{b}_{s,0}^{l}-\left(\frac{2}{\sqrt{m}}\left(W_{20}^{l}\right)^{T} \Sigma^{\prime l}- \Sigma^{\prime l}\right) \mathbf{b}_{s}^{l} \| \\
&= \| \frac{2}{\sqrt{m}}\left(\left(W_{2}^{l}\right)^{T}-\left(W_{20}^{l}\right)^{T}\right) \Sigma^{\prime l} \mathbf{b}_{s}^{l} 
+\left(\frac{2}{\sqrt{m}}\left(W_{20}^{l}\right)^{T}\left(\Sigma^{\prime l}-\Sigma_{0}^{\prime l}\right)-\left(\Sigma^{\prime l}-\Sigma_{0}^{\prime l}\right)\right) \mathbf{b}_{s,0}^{l} \\
& \quad +\left(\frac{2}{\sqrt{m}}\left(W_{20}^{l}\right)^{T} \Sigma^{\prime l}- \Sigma^{\prime l}\right)\left(\mathbf{b}_{s}^{l}-\mathbf{b}_{s,0}^{l}\right) \| \\
& \leq  \| \frac{2}{\sqrt{m}}\left(\left(W_{2}^{l}\right)^{T}-\left(W_{20}^{l}\right)^{T}\right) \Sigma^{\prime l} \mathbf{b}_{s}^{l} \| 
+\frac{1}{\sqrt{m}}\left\|\left(\left(2\left(W_{20}^{l}\right)^{T}-\sqrt{m} I\right)\left(\Sigma^{\prime l}-\Sigma_{0}^{\prime l}\right)\right) \mathbf{b}_{s,0}^{l}\right\| \\
&+ \frac{1}{\sqrt{m}}\left\|\left(\left(2\left(W_{20}^{l}\right)^{T}-\sqrt{m} I\right) \Sigma^{\prime l}\right)\left(\mathbf{b}_{s}^{l}-\mathbf{b}_{s,0}^{l}\right)\right\|\\
&=T_1+T_2+T_3.
\end{aligned}
\end{equation*}
We now provide bounds on $T_1, T_2,$ and $T_3$:
\begin{equation*}
    \begin{aligned}
        T_1&= \left\| \frac{2}{\sqrt{m}}\left(\left(W_{2}^{l}\right)^{T}-\left(W_{20}^{l}\right)^{T}\right) \Sigma^{\prime l} \mathbf{b}_{s}^{l} \right\|
        \leq \frac{2}{\sqrt{m}} \left\|W_{2}^{l}-W_{20}^{l}\right\| \left\| \Sigma^{\prime l} \right\|  \left\| \mathbf{b}_{s}^{l}\right\| \leq \frac{2R_2 L_{\sigma}^{L-l+1}\left(2\left(c_{20}+R_2 / \sqrt{m}\right)+1\right)^{L-l}}{\sqrt{m}}\\
        &=O(1/\sqrt{m}).
    \end{aligned}
\end{equation*}
To obtain bound on $T_2$, we need the following inequality,
\begin{equation*}
    \begin{aligned}
\left\|\tilde{\mathbf{z}}^{l}(\mathbf{W})-\tilde{\mathbf{z}}^{l}\left(\mathbf{W}_{0}\right)\right\| 
&= \| \frac{1}{\sqrt{m}} W_{2}^{l} \mathbf{z}^{l-1}(\mathbf{W})-\frac{1}{\sqrt{m}} W_{20}^{l} \mathbf{z}^{l-1}\left(\mathbf{W}_{0}\right)-\frac{1}{\sqrt{m}} W_{2}^{l} \mathbf{u}^{l-1}(\mathbf{W})+\frac{1}{\sqrt{m}} W_{20}^{l} \mathbf{u}^{l-1}\left(\mathbf{W}_{0}\right) \\
& \quad +\frac{1}{\sqrt{n}} W_{1}^{l} \mathbf{y}-\frac{1}{\sqrt{n}} W_{10}^{l} \mathbf{y} \| \\
& \leq  \frac{1}{\sqrt{m}}\left\|W_{20}^{l}\right\| L_{\sigma}\left\|\tilde{\mathbf{z}}^{l-1}(\mathbf{W})-\tilde{\mathbf{z}}^{l-1}\left(\mathbf{W}_{0}\right)\right\|+\frac{1}{\sqrt{m}}\left\|W_{2}^{l}-W_{20}^{l}\right\|\left\|\mathbf{z}^{l-1}(\mathbf{W})\right\| \\
& \quad +\frac{1}{\sqrt{m}}\left\|W_{20}^{l}\right\|\left\|\mathbf{u}^{l-1}(\mathbf{W})-\mathbf{u}^{l-1}\left(\mathbf{W}_{0}\right)\right\|+\frac{1}{\sqrt{m}}\left\|W_{2}^{l}-W_{20}^{l}\right\|\left\|\mathbf{u}^{l-1}(\mathbf{W})\right\| 
+\frac{1}{\sqrt{n}}\left\|W_{1}^{l}-W_{10}^{l}\right\|\|\mathbf{y}\| \\
&\leq  c_{20} L_{\sigma}\left\|\tilde{\mathbf{z}}^{l-1}(\mathbf{W})-\tilde{\mathbf{z}}^{l-1}\left(\mathbf{W}_{0}\right)\right\|+\frac{1}{\sqrt{m}}\left\|W_{2}^{l}-W_{20}^{l}\right\|\left\|\mathbf{z}^{l-1}(\mathbf{W})\right\| \\
& \quad +c_{20}\left\|\mathbf{u}^{l-1}(\mathbf{W})-\mathbf{u}^{l-1}\left(\mathbf{W}_{0}\right)\right\|+\frac{1}{\sqrt{m}}\left\|W_{2}^{l}-W_{20}^{l}\right\|\left\|\mathbf{u}^{l-1}(\mathbf{W})\right\| 
+\frac{1}{\sqrt{n}}\left\|W_{1}^{l}-W_{10}^{l}\right\|\|\mathbf{y}\| \\
& \leq  c_{20} L_{\sigma}\left\|\tilde{\mathbf{z}}^{l-1}(\mathbf{W})-\tilde{\mathbf{z}}^{l-1}\left(\mathbf{W}_{0}\right)\right\|+c_{20}\left\|\mathbf{u}^{l-1}(\mathbf{W})-\mathbf{u}^{l-1}\left(\mathbf{W}{ }_{0}\right)\right\|\\
& \quad +\frac{R_{2}}{\sqrt{m}}\left(c_{\mathrm{ADMM} ; \mathbf{z}}^{l-1}(m)+c_{\mathrm{ADMM} ; \mathbf{u}}^{l-1}(m)\right)+R_{1} C_y.
\end{aligned}
\end{equation*}
Since 
\begin{equation*}
    \left\|\mathbf{u}^{l}(\mathbf{W})-\mathbf{u}^{l}\left(\mathbf{W}_{0}\right)\right\|\leq\left(L_{\sigma}+1\right)\left\|\tilde{\mathbf{z}}^{l}(\mathbf{W})-\tilde{\mathbf{z}}^{l}\left(\mathbf{W}_{0}\right)\right\|,
\end{equation*}
we have
\begin{equation*}
    \left\|\tilde{\mathbf{z}}^{l}(\mathbf{W})-\tilde{\mathbf{z}}^{l}\left(\mathbf{W}_{0}\right)\right\| \leq c_{20}\left(2L_{\sigma}+1\right)\left\|\tilde{\mathbf{z}}^{l-1}(\mathbf{W})-\tilde{\mathbf{z}}^{l-1}\left(\mathbf{W}_{0}\right)\right\|+\frac{R_{2}}{\sqrt{m}}\left(c_{\mathrm{ADMM} ; \mathbf{z}}^{l-1}(m)+c_{\mathrm{ADMM} ; \mathbf{u}}^{l-1}(m)\right)+R_{1} C_y.
\end{equation*}
Since
\begin{equation*}
    \begin{aligned}
\left\|\tilde{\mathbf{z}}^{(1)}(\mathbf{W})-\tilde{\mathbf{z}}^{(1)}\left(\mathbf{W}_{0}\right)\right\| & \leq \frac{1}{\sqrt{m}}\left\|W_{2}^{(1)}-W_{20}^{(1)}\right\|\left\|\mathbf{z}^{(0)}-\mathbf{u}^{(0)}\right\|+\frac{1}{\sqrt{n}}\left\|W_{1}^{(1)}-W_{10}^{(1)}\right\|\|\mathbf{y}\| \\
& \leq R_{2}\left(C_{\mathbf{z}}+C_{\mathbf{u}}\right)+R_{1} C_y.
\end{aligned}
\end{equation*}
Recursively applying the previous equation, we get
\begin{equation*}
    \begin{aligned}
&\left\|\tilde{\mathbf{z}}^{l}(\mathbf{W})-\tilde{\mathbf{z}}^{l}\left(\mathbf{W}_{0}\right)\right\| \\
&\leq \left(\frac{R_{2}}{\sqrt{m}}\left(c_{\mathrm{ADMM} ; \mathbf{z}}^{l-1}(m)+c_{\mathrm{ADMM} ; \mathbf{u}}^{l-1}(m)\right)+R_{1} C_y\right) \sum_{i=0}^{l-2} c_{20}^{i}\left(L_{\sigma}+1\right)^{i}+c_{20}^{l-1}\left(L_{\sigma}+1\right)^{l-1}\left(R_{2}\left(C_{\mathbf{z}}+C_{\mathbf{u}}\right)+R_{1} C_y\right) \\
& = O(1).
%c_{\mathrm{ADMM} ; \Delta \tilde{\mathbf{z}}}^{l}
\end{aligned}
\end{equation*}
Using the above inequality bound and Lemma (\ref{lm:15}), we can write the following with probability $1-m e^{-c_{bs}^{l} \ln ^{2}(m)}$:
\begin{equation*}
    \begin{aligned}
\left\|\left[\Sigma^{l}-\Sigma_{0}^{\prime l}\right] \mathbf{b}_{s,0}^{l}\right\| &=\sqrt{\sum_{i=1}^{m}\left(\mathbf{b}_{s,0}^{l}\right)_{i}^{2}\left[\sigma^{\prime}\left(\tilde{\mathbf{z}}^{l}(\mathbf{W})\right)-\sigma^{\prime}\left(\tilde{\mathbf{z}}^{l}\left(\mathbf{W}_{0}\right)\right)\right]^{2}} 
 \leq\left\|\mathbf{b}_{s,0}^{l}\right\|_{\infty} \sqrt{\sum_{i=1}^{m}\left[\sigma^{\prime}\left(\tilde{\mathbf{z}}^{l}(\mathbf{W})\right)-\sigma^{\prime}\left(\tilde{\mathbf{z}}^{l}\left(\mathbf{W}_{0}\right)\right)\right]^{2}} \\
& \leq\left\|\mathbf{b}_{s,0}^{l}\right\|_{\infty} \beta_{\sigma}\left\|\tilde{\mathbf{z}}^{l}(\mathbf{W})-\tilde{\mathbf{z}}^{l}\left(\mathbf{W}_{0}\right)\right\| =\tilde{O}(1/\sqrt{m}).
%\leq \beta_{\sigma} c_{\mathrm{ADMM} ; b s}^{l} c_{\mathrm{ADMM} ; \Delta \tilde{\mathbf{z}}}^{l}
\end{aligned}
\end{equation*}
This leads to,
\begin{equation*}
\begin{aligned}
    T_2 = \frac{1}{\sqrt{m}}\left\|\left(\left(2\left(W_{20}^{l}\right)^{T}-\sqrt{m} I\right)\left(\Sigma^{\prime l}-\Sigma_{0}^{\prime l}\right)\right) \mathbf{b}_{s,0}^{l}\right\| 
    \leq \frac{1}{\sqrt{m}}\|2\left(W_{20}^{l}\right)^{T}-\sqrt{m} I \| \left\|\left[\Sigma^{l}-\Sigma_{0}^{\prime l}\right] \mathbf{b}_{s,0}^{l}\right\|=\tilde{O}(1/\sqrt{m}). 
    \end{aligned}
\end{equation*}
Besides, by using the induction hypothesis on $l$, the term $T_3$ is bounded as
\begin{equation*}
    \begin{aligned}
        T_3&=\frac{1}{\sqrt{m}}\left\|\left(\left(2\left(W_{20}^{l}\right)^{T}-\sqrt{m} I\right) \Sigma^{\prime l}\right)\left(\mathbf{b}_{s}^{l}-\mathbf{b}_{s,0}^{l}\right)\right\| 
        \leq \frac{1}{\sqrt{m}} \|2\left(W_{20}^{l}\right)^{T}-\sqrt{m} I \| \| \Sigma^{\prime l}\| \| \mathbf{b}_{s}^{l}-\mathbf{b}_{s,0}^{l}\|= \tilde{O}(1/\sqrt{m}).
    \end{aligned}
\end{equation*}
Now combining the bounds on the terms $T_1,\ T_2$ and $T_3$, we can write
\begin{equation}
\label{eqad:}
\left\|\mathbf{b}_{s}^{l-1}-\mathbf{b}_{s,0}^{l-1}\right\|\leq T_1+T_2+T_3 = \tilde{O}\left( \frac{1}{\sqrt{m}} \right).
\end{equation}
Therefore, \eqref{d2eqad} is true for $l-1$. Hence, by induction \eqref{d2eqad} is true for all $l \in [L]$.
\end{proof}
By using Lemma \ref{lm:15} and \ref{lm:16}, in equation \eqref{eq:inf}, we get
\begin{equation}
\left\|\mathbf{b}_{s}^{l}\right\|_{\infty} \leq\left\|\mathbf{b}_{s,0}^{l}\right\|_{\infty}+\left\|\mathbf{b}_{s}^{l}-\mathbf{b}_{s,0}^{l}\right\|  = \tilde{O}\left(\frac{1}{\sqrt{m}}\right).
\end{equation}
This implies,
\begin{equation}
\mathcal{Q}_{\infty}\left(f_s\right)= \max _{1 \leq l \leq L}\left\{\left\|\mathbf{b}_s^l\right\|_{\infty}\right\} = \tilde{O}\left(\frac{1}{\sqrt{m}}\right).
\end{equation}

%\leq \max _{1 \leq l \leq L}\left\{c_{\mathrm{ADMM} ; b s}^{l}+c_{\mathrm{ADMM} ; b}^{l} \sum_{i=0}^{L-l+1} c_{20}^{i} L_{\sigma}^{i}\right\}=\mathcal{C}_{\mathrm{ADMM} ; \infty}$
\bibliographystylesupp{ieeetr}
\bibliographysupp{bibs_2}
% \bibliography{bibs_2}

\end{document}